\documentclass[journal]{IEEEtran}
\usepackage{tikz}
\usepackage{tikz}

\usepackage{pifont}

\usepackage{amssymb}
\usepackage{supertabular}

\usepackage{bbm}
\usepackage{rotating}
\usepackage{dsfont}

\usepackage{bm}
\usepackage{makecell}
\usepackage{array}
\usepackage{eucal}
\usepackage{graphicx}
\usepackage{epstopdf}
\usepackage{amsfonts,amsthm}
\usepackage{setspace}
\usepackage{cite}
\usepackage[colorlinks,linkcolor=blue,anchorcolor=blue,citecolor=blue]{hyperref}
\usepackage{subfig}
\usepackage{subfloat}
\usepackage{multirow,booktabs}

\usepackage{amsmath,mathrsfs}
\usepackage[misc]{ifsym}

\usepackage{amsmath,graphicx}
\usepackage{epstopdf}
\usepackage{amsfonts,amsthm}
\usepackage{algorithmic}
\usepackage[ruled,linesnumbered,boxed]{algorithm2e}
\usepackage{setspace}
\usepackage{cite}
\usepackage[colorlinks,linkcolor=blue,anchorcolor=blue,citecolor=blue]{hyperref}
\usepackage{subfig}
\usepackage{subfloat}
\usepackage{multirow,booktabs}

\usepackage{amsmath}
\usepackage{amssymb}
\usepackage[misc]{ifsym}
\usepackage{url}

\usepackage{color,soul}
\definecolor{myred}{RGB}{255,0,0}
\definecolor{myblue}{RGB}{0,0,255}
\usepackage{threeparttable}

\usepackage{hyperref}
\hypersetup{hypertex=true,
            colorlinks=true,
            linkcolor=red,
            anchorcolor=green,
            citecolor=blue}

\newtheorem{Definition}{Definition}[section]
\newtheorem{Theorem}{Theorem}[section]
\newtheorem{Remark}{Remark}[section]
\newtheorem{Lemma}{Lemma}[section]

\newcommand{\tabincell}[2]{\begin{tabular}{@{}#1@{}}#2\end{tabular}}

\ifCLASSINFOpdf
\else
\fi

\begin{document}

\title{Multi-Dimensional 
Visual  Data
 Recovery:
Scale-Aware
Tensor Modeling and 
Accelerated Randomized
 Computation
}

\author{
Wenjin~Qin,~Hailin~Wang,~Jiangjun~Peng,~Jianjun~Wang,~\IEEEmembership{Member,~IEEE,}~and~Tingwen~Huang,~\IEEEmembership{Fellow,~IEEE}
\thanks{
%
This work was supported in part by the National Key Research and Development Program of China under Grant 2023YFA1008502;
in part by the National Natural Science Foundation of China under Grant U24A2001,
in part by the Natural Science Foundation of Chongqing, China, under Grant CSTB2023NSCQ-LZX0044; 
and in part by Chongqing Talent Project, China, under Grant
cstc2021ycjh-bgzxm0015.  (Corresponding author: Jianjun Wang.)  
%
}
\vspace{-0.9500cm}
\thanks{Wenjin Qin and Jianjun Wang are with the School of Mathematics and Statistics, Southwest University, Chongqing 400715, China (e-mail:
qinwenjin2021@163.com, wjj@swu.edu.cn).}
\thanks{Hailin Wang 
is with the  Department of Statistics and Data Science,   The Chinese University of Hong Kong, Shatin, Hong Kong
(e-mail: wanghailin97@163.com).}
\thanks{Jiangjun Peng is with the School of Mathematics
and Statistics, Northwestern Polytechnical University, Xi'an 710021, China
(e-mail: pengjj@nwpu.edu.cn).}
\thanks{Tingwen Huang is with the Faculty of Computer Science and Control Engineering, Shenzhen University of Advanced Technology, Shenzhen 518055,
China (e-mail: huangtingwen2013@gmail.com).}
}
\maketitle
\begin{abstract}
The recently proposed \textit{fully-connected tensor network} (FCTN) decomposition  has demonstrated  significant advantages in
correlation characterization and transpositional invariance, and has  achieved notable achievements  in 
multi-dimensional data
processing and analysis.
However, existing  multi-dimensional data recovery 
methods leveraging FCTN decomposition  still have room for further enhancement,
particularly in   computational efficiency and modeling capability. 
 To address these issues,  we first propose a FCTN-based generalized nonconvex regularization paradigm  from the  perspective of gradient mapping. Then, reliable and scalable  multi-dimensional data 
 recovery models are investigated,
 where the model formulation is shifted from unquantized observations to coarse-grained quantized observations.
 Based on  the \textit{alternating direction method of multipliers} (ADMM) framework,
 we  derive  efficient optimization   algorithms with convergence guarantees to solve the formulated  models.
 To alleviate the computational bottleneck encountered when processing large-scale 
  multi-dimensional data,
fast and efficient randomized compression  algorithms are    devised in virtue of sketching techniques in numerical linear algebra.
 These dimensionality-reduction techniques serve as the computational acceleration core of our proposed algorithm framework.
   Theoretical results on approximation error upper bounds and convergence analysis
   for the proposed method are derived.
 %
Extensive numerical experiments 
illustrate the effectiveness and superiority of the proposed algorithm  over other state-of-the-art methods in terms of quantitative  metrics, visual quality, and running time.

\end{abstract}

\IEEEpeerreviewmaketitle

\section{\textbf{Introduction}}

\IEEEPARstart{W}ith the tremendous advancement of sensorial and information technology,
 multi-dimensional visual data  have emerged widely  
 in numerous modern 
 application fields, including
 statistics \cite{han2022optimal,cai2023generalized},
 biomedical imaging \cite{ mahyari2016tensor, lll10494874},  remote sensing \cite{wang2025hypersigma, qin2024tensor, qin2025untrfr},
 image and video processing \cite{qin2022low, qin2023nonconvex},
   and
  computer vision \cite{ wang2023guaranteed , luo2024low}.
These data   typically exhibit  high dimensionality, multiple modalities, 
 large volume, high veracity, 
 and complex structure, 
 thereby posing substantial challenges to subsequent relevant processing and analysis tasks
 \cite{cichocki2016tensor,park2016bigtensor, song2019tensor}.
 For instance, large volume  demands scalable strategies that can adapt to increasing data scales;
 high veracity calls for robust and reliable  methods  for noisy, incomplete and/or inconsistent data;
  and
 high dimensionality requires efficient compression 
 techniques to mitigate the curse of dimensionality.
As such, developing efficient  and  scalable multi-dimensional data
processing/analysis 
approaches
that can jointly address these challenges has become increasingly urgent  and crucial.

Tensors (also known as  multi-dimensional arrays),  as the higher-order generalization of vectors and matrices,  possess  powerful capabilities  in
capturing the intrinsic structural characteristics   embedded within a wide range of   high-dimensional data.
This advantage  has driven   the development of  extensive 
 tensor-based  methodologies and theoretical frameworks, which are tailored
  for high-dimensional data processing and analysis \cite{liu2021tensors, liu2022tensor}.
Among them, multi-dimensional data recovery is the most fundamental and 
 crucial  task,
 which addresses the challenge of reconstructing the clean data 
from its imperfect counterpart degraded by 
noise pollution, information missing, outliers interference, and other factors.
Based on the solutions and techniques employed, existing tensor  approaches for high-dimensional data recovery
can be subdivided into two branches.
The first branch is tensor factorization-based methods, which alternatively update the low-dimensional factors with the predefined initial tensor rank
\cite{tong2022scaling, qin2024guaranteed, li2023robust88,  zheng2021fully, liu2024fully,
wu2022tensor, liu2024low11}.
While demonstrating computational efficiency in certain scenarios, their performance heavily relies on accurate rank estimation.
The other important branch is tensor regularization-based methods,
whose idea is to characterize prior features (e.g., low-rankness, smoothness, sparsity)
underlying multi-dimensional  data finely via proper   regularization items
 \cite{liu2012tensor, goldfarb2014robust,huang2015provable,mu2014square,  zheng2020tensor44, wang2020robust11, feng2023multiplex, liu2024revisiting, zhang2023tensor, 
bengua2017efficient, yu2019tensor,huang2020robust, yuan2019tensor777 
}.

The \textit{Tensor Nuclear Norm} (TNN) serves as a  benchmark regularization paradigm, effectively inducing low-rank structure in 
high-dimensional  data modeling.
Within diverse tensor decomposition frameworks, researchers have extensively explored  TNN-based regularization approaches for high-dimensional data recovery,
accompanied by rigorous theoretical guarantees on exact recovery.
Some typical examples  include
   SNN    \cite{ liu2012tensor}  under Tucker framework,
   TTNN \cite{bengua2017efficient} under \textit{tensor train} (TT) framework,
   TRNN \cite{yu2019tensor,huang2020robust} under \textit{tensor ring} (TR) framework,
   FCTN nuclear norm \cite{ liu2024fully} under \textit{Fully-Connected Tensor Network} (FCTN) framework,
   and 
   HTNN \cite{ qin2022low}, WSTNN \cite{zheng2020tensor44}, OITNN \cite{wang2020robust11}, METNN \cite{liu2024revisiting}
    under \textit{Tensor Singular Value Decomposition} (T-SVD) framework.
   To further enhance the recovery performance, a series of nonconvex variants with stronger low-rankness promoting capabilities have been successively proposed
   \cite{zhang2018nonconvex1,  zhang2019robust,
   wang2021generalized, yang2022355,  wang2024low2222,  zhang2023tensor,
 chen2020robust,qiu2021nonlocal,zhao2022robust,qiu2024robust, 
  qin2023nonconvex , gao2023tensor, yu2024generalized,qin2024tensor
  }.
  Another  influential   regularization paradigm, 
  \emph{Tensor Correlated Total Variation} (TCTV), is essentially 
   a tensor nuclear norm imposed on the gradient domain \cite{wang2023guaranteed}.
This method  has been widely recognized for its ability 
   to simultaneously constrain 
   the low-rankness and smoothness of multi-dimensional data, see \cite{wang2023guaranteed} and its extensions \cite{hou2024tensor 
 ,zhang2024fusion,feng2024poisson
  }.

Although the above 
tensor regularization-based methods
  yield commendable  performance 
 in practical applications, 
 they demonstrate inefficiency in handling   large-scale multi-dimensional   data.
In response to the demands arising from the ``V"  features of the big data era, namely large Volume, high Velocity and high Veracity,
it is of crucial importance to
address the issue of large-scale multi-dimensional   data recovery
 in a fast and  accurate 
  manner.
Speed  is 
 manifested in the computation module, whereas accuracy is 
 embodied in the modeling module.
However,
 achieving high accuracy and fast speed
 simultaneously remains fraught with some   challenges, 
 because there is a trade-off between accuracy and speed.

\subsection{\textbf{Our Motivations}}


Compared with the TT and TR decompositions, the  FCTN decomposition   adequately characterizes the correlations between any modes of the tensor,
and avoids the influence of the permutations and the ordering of the tensor dimensions,
thus showing outstanding performance on 
 multi-dimensional data  modeling  
\cite{zheng2021fully, zheng2022tensor, lyu2022multi, liu2024fully,  zheng2023spatial22,tu2024fully, yu2025low,  han2024nested , zheng2024provable}.
 However,  existing  multi-dimensional data 
 recovery methods
 \cite{zheng2021fully, zheng2022tensor, lyu2022multi,liu2024fully, zheng2023spatial22,tu2024fully, yu2025low,  han2024nested}
leveraging FCTN decomposition still have room for further enhancement, particularly in  modeling capability 
and  computational efficiency.

\subsubsection{\textbf{Modeling perspective}}
In many application scenarios,  the  multi-dimensional visual  
data tend to exhibit
strong global low-rankness (\textbf{L})   due to redundancy,
 and great local smoothness (\textbf{S}) due to fine-grained continuity.
 However, existing FCTN-based tensor  regularization methods often failed to simultaneously capture these hybrid \textbf{L}+\textbf{S}
characteristics in a concise yet effective manner.
Besides,  previous related models  are limited to unquantized observations and are blind to the fact that real-world applications typically involve quantization of the data
 \cite{zheng2021fully, zheng2022tensor, lyu2022multi,liu2024fully, zheng2023spatial22,tu2024fully, yu2025low,  han2024nested}.
%
For  some  machine learning or signal/image  processing systems  involving large-scale  tensors,
it would be impractical to work with original tensor  with high precision
due to the 
prohibitive storage requirements, communication bandwidth limitations 
or excessive power consumption \cite{chen2023high, 
hou2021robust, hou2025robust}.
In this situation, a coarse-grained quantization method using a small number of bit-widths is more appropriate.
%
%
\textit{Therefore, we consider   investigating  concise yet effective FCTN-based tensor regularization strategies to reveal  the
 insightful  prior features  of large-scale multi-dimensional data. 
 This scheme takes into account the impact of data scale variations on its structural characteristics.
  Furthermore,   reliable and scalable models for robust multi-dimensional data recovery
are  developed, where the model formulation is shifted from unquantized  to coarsely quantized observations.
This is the first research motivation   of this paper.}

\subsubsection{\textbf{Computational perspective}}
From a computational perspective,  connecting any two factors in FCTN decomposition  is a double-edged sword:
the upside is that FCTN factorization  provides a more flexible and superior representation compared to other decompositions;
 the downside is that FCTN decomposition  encounters a high computational burden   when dealing with large-scale 
 multi-dimensional
 data,
 thereby incurring significant expenses for the associated 
  tensor
 modeling methods
 \cite{zheng2021fully, zheng2022tensor, lyu2022multi, liu2024fully,zheng2023spatial22,tu2024fully, yu2025low,   han2024nested  }.
 This is because they   generally experience the calculation of multiple SVDs of large-sized unfolding matrices or multiple large-sized \textit{alternate least squares} (ALS) problems.
 Recently,  randomized sketching techniques  have been extensively investigated and have achieved great success
 in large-scale data analysis, especially  in matrix/tensor approximation \cite{minster2020randomized,che2025efficient333, zheng2024provable,
 wang2024randomized}.
 %
The strength of randomized algorithms lies in their excellent computational  efficiency, coupled with low memory storage requirements. Furthermore, their core operations  can be optimized for maximum efficiency on modern computational platforms.
\textit{Therefore,  we consider developing  
novel  fast and accurate 
 compression algorithms  for large-scale 
 high-order tensors 
in virtue of randomized sketching techniques,
which  maintain a low computational complexity while achieving an acceptable   accuracy.
These strategies aimed at dimensionality-reduction serve as the
fundamental computational acceleration component in the design of our multi-dimensional data  
recovery algorithms.
 This is the second research motivation   of this paper.}

\subsection{\textbf{Our Contributions}}

  1)
%
 In virtue of randomized sketching techniques in numerical linear algebra,  we first devise  fast  and accurate randomized  algorithms for large-scale high-order tensor compression. Additionally,  theoretical results on error upper bounds for the proposed randomized  algorithms are derived.
Compared to existing  compression approaches, the proposed method exhibits notable advantages in   computational efficiency.



  2)
  We propose an innovative nonconvex gradient-domain  regularizer   under FCTN decomposition framework,
which simultaneously encodes   both global low-rankness and local continuity of large-scale tensors.
%
By combining it  with another enhanced  nonconvex  noise/outlier regularization,
novel  robust  multi-dimensional data 
recovery models are established, which cover  a range from non-quantized to quantized observations.

  3)
  To efficiently solve the proposed multiple recovery models, we develop  efficient 
  ADMM-based  algorithms  and theoretically demonstrate their convergence.
Besides,  to alleviate the   computational burden  faced by the proposed algorithms when processing large-scale  multi-dimensional data, 
we technically  embed the proposed randomized  compression modules  
into  its 
time-intensive  optimization  procedure.

  4) Extensive 
  experiments conducted on color images/videos,   
  hyperspectral images/videos, magnetic
resonance images, 
   and  face
databases
      provide empirical support  for the theoretical analysis,
      and demonstrate that the proposed method  outperforms other state-of-the-art  approaches quantitatively and qualitatively.
Strikingly, in cases of slight 
or no accuracy loss, our  method  incorporated randomized  techniques decreases the CPU time by  approximately
$88\%$  in comparison to the deterministic version (\textit{please refer to the experimental section for details}).



\section{\textbf{notations and preliminaries}}\label{nota}

In this section, we summarize the critical notations and preliminaries  utilized in this paper.


\subsubsection{\textbf{Notations}}
We use $x$, $\textbf{x}$, $\textbf{X}$, and $\bm{\mathcal{X}}$ to denote scalars, vectors, matrices, and tensors, respectively.
 For simplicity, we use $\textbf{x}_{1:q}$,  $\textbf{X}_{1:q}$, and $\bm{\mathcal{X}} _{1:q}$ to
represent ordered sets $\{\textbf{x}_{1}, \cdots, \textbf{x}_{q}   \}$, $\{\textbf{X}_{1},  \cdots, \textbf{X}_{q}\}$, and
$\{\bm{\mathcal{X}} _{1}, \cdots, \bm{\mathcal{X}} _{q}\}$,
respectively.
For tensor $\boldsymbol{\mathcal{X}} \in \mathbb{R}^{I_1 \times I_2 \times \cdots \times I_N}$,
we denote 
$\boldsymbol{\mathcal{X}}_ {i_1,i_2,\cdots,i_N}$
as its $(i_1,i_2,\cdots,i_N)$-th element.
The \textbf{inner product} of two tensors $\boldsymbol{\mathcal{X}}$ and $\boldsymbol{\mathcal{Y}}$ with the same size is defined as the sum of the products of their entries, i.e., $\langle \boldsymbol{\mathcal{X}}, \boldsymbol{\mathcal{Y}} \rangle = \sum_{i_1,i_2,\cdots,i_N}
\boldsymbol{\mathcal{X}} _{i_1,i_2,\cdots,i_N}    \cdot \boldsymbol{\mathcal{Y}} _{i_1,i_2,\cdots,i_N}$.
The \textbf{$\mathit{l}_1$-norm}, \textbf{Frobenius norm}
 and \textbf{$\mathit{l}_{\mathnormal{F}, 1} $-norm}
of $\boldsymbol{\mathcal{X}}$ are defined as $\|\boldsymbol{\mathcal{X}}\|_1=\sum_{i_1,i_2,\cdots,i_N}|\boldsymbol{\mathcal{X}}_  {i_1,i_2,\cdots,i_N} |$, $\|\boldsymbol{\mathcal{X}}\|_F=\sqrt{\sum_{i_1,i_2,\cdots,i_N}| \boldsymbol{\mathcal{X}} _{ i_1,i_2,\cdots,i_N} |^2}$
 and
 $\| {\boldsymbol{\mathcal{X}}}\|_{\mathnormal{F}, 1}  =  
\sum_{i_1, i_2}
\|{\boldsymbol{\mathcal{X}}}_{i_1 i_2 : \cdots : }
\|_{ \mathnormal{F}  }$, respectively.
Supposing that
the vector  $\textbf{n}$ is a  reordering 
of the vector $[1,2,\cdots,N]$.
The vector  $\textbf{n}$-based  \textbf{generalized tensor transposition} of $\bm{\mathcal{X}}$
is denoted as $\vec{\bm{\mathcal{X}}}^{\textbf{n}} \in \mathbb{R}^{I_{n_1} \times I_{n_2} \times \cdots \times I_{n_N}}$,
which is generated by
rearranging the modes of $\bm{\mathcal{X}}$ in the order  specified by the vector $\textbf{n}$.
The corresponding operation and its inverse operation are denoted as $\vec{\bm{\mathcal{X}}}^{\textbf{n}} =\rm{permute}(\bm{\mathcal{X}},\textbf{n})$ and $\bm{\mathcal{X}}=\rm{ipermute}$$(\vec{\bm{\mathcal{X}}}^{\textbf{n}},\textbf{n})$, respectively.
%
%
The \textbf{generalized  tensor unfolding} of $\bm{\mathcal{X}}$ is a matrix defined as $\textbf{X}_{[\textbf{n}_{1:d};\textbf{n}_{d+1:N}]}=\rm{reshape}$ $(\vec{\bm{\mathcal{X}}}^{\textbf{n}},
\prod_{i=1}^d I_{\textbf{n}_i}, \prod_{i=d+1}^N I_{\textbf{n}_i})$. The corresponding inverse operation is defined as $\bm{\mathcal{X}}=\Upsilon (\textbf{X}_{[\textbf{n}_{1:d};\textbf{n}_{d+1:N}]})$.



\subsubsection{\textbf{FCTN framework}}
Below, we introduce  some key algebraic foundations  under the FCTN decomposition  framework, which will be used extensively throughout this paper.
Please refer to the literatures \cite{zheng2021fully, zheng2022tensor,liu2024fully} for more details.

\begin{Definition}\label{tensor-FCTN} 
\textbf{(FCTN decomposition \cite{zheng2021fully})
}
The FCTN decomposition aims to decompose
an order-$N$  tensor $\bm{\mathcal{X}} \in \mathbb{R}^{I_1 \times I_2 \times \cdots \times I_N}$ 
into a series of
order-$N$
factor tensors $\bm{\mathcal{G}}_k \in \mathbb{R}^{R_{1,k}\times R_{2,k}\times \cdots \times R_{k-1,k}\times I_k \times R_{k,k+1} \times \cdots \times R_{k,N}}, (k=1,2,\cdots,N)$. 
The element-wise form of the FCTN decomposition can be expressed as
 \begin{equation}\label{fctndec}
 \begin{aligned}
&\bm{\mathcal{X}} (i_1,i_2,\cdots,i_N)
= \\
&  \sum_{r_{1,2}=1}^{R_{1,2}}\sum_{r_{1,3}=1}^{R_{1,3}}\cdots \sum_{r_{1,N}=1}^{R_{1,N}}\sum_{r_{2,3}=1}^{R_{2,3}}\cdots\sum_{r_{2,N }=1}^{R_{2,N}} \cdots \sum_{r_{N-1,N}=1}^{R_{N-1,N}}
\\
&
\{ \bm{\mathcal{G}}_1 (i_1,r_{1,2},r_{1,3},\cdots,r_{1,N})
 \bm{\mathcal{G}}_2 (r_{1,2},i_2,r_{2,3},\cdots,r_{2,N}) \cdots
 \\
 &
 \bm{\mathcal{G}}_k (r_{1,k},r_{2,k},\cdots,r_{k-1,k},i_k,r_{k,k+1},\cdots,r_{k,N})
 \cdots
 \\
 &
 \bm{\mathcal{G}}_N (r_{1,N},r_{2,N},\cdots,r_{N-1,N},i_N)\}.
 \end{aligned}
 \end{equation}
The factors $\bm{\mathcal{G}}_1$, $\bm{\mathcal{G}}_2$, $\cdots$,  $\bm{\mathcal{G}}_N$ are the core tensors of $\bm{\mathcal{X}}$ and can be abbreviated as $\{  \bm{\mathcal{G}} \}  _{1:N}$, then ${ \bm{\mathcal{X}} }  ={\rm{FCTN}}   (\{\bm{\mathcal{G}}\}_{1:\mathit{N}})$. The vector $(R_{1,2}, R_{1,3},\cdots, R_{1,N},$ $R_{2,3},R_{2,4},\cdots, R_{2,N}, \cdots, R_{N-1,N})$ is defined as the FCTN-rank of the original tensor $\bm{\mathcal{X}}$.
\end{Definition}

\begin{Lemma}(\textbf{Transpositional Invariance  \cite{zheng2021fully}})
\label{reshssss}
Supposing that an order-${N}$ 
tensor  $\bm{\mathcal{X}} \in \mathbb{R}^{I_1 \times I_2 \times \cdots \times I_N}$
has the following FCTN decomposition:
${ \bm{\mathcal{X}} }  ={\rm{FCTN}}   (\bm{\mathcal{G}}_{1}, \bm{\mathcal{G}}_{2}, \cdots, \bm{\mathcal{G}}_{\mathit{N}})$.
Then its vector $\textbf{n}$-based generalized tensor transposition $\vec{\bm{\mathcal{X}}}^{\textbf{n}}$
can be expressed as
${\vec{\bm{\mathcal{X}}}} ^ {\textbf{n}}   ={\rm{FCTN}} ({\vec{\bm{\mathcal{G}}}} ^ {\textbf{n}} _{n_1},
{\vec{\bm{\mathcal{G}}}} ^ {\textbf{n}} _{n_2},\cdots, {\vec{\bm{\mathcal{G}}}} ^ {\textbf{n}} _{n_N}    )$,
where $\textbf{n}= ({n}_1,{n}_2, \cdots,{n}_N)$ is a reordering of the vector  $(1,2, \cdots,N)$.
\end{Lemma}

\subsubsection{\textbf{Uniformly Dithered Quantization}} 
Below, we introduce the uniformly dithered quantization procedure for high-order tensor.
It mainly involves the following two steps:

(I) \textbf{Dithering step:}
This step is applied to tensor's entries 
that we plan to quantize.
For instance,
let $y_k$ ($\forall k \in [m]$)
 be the $k$-th noisy observation from the  underlying  tensor
${\boldsymbol{\mathcal{L}}} ^{*}  \in\mathbb{R}^{I_1\times \cdots\times I_N}$ via
$
y_k  =  \langle{\boldsymbol{\mathcal{P}}}_{k},{\boldsymbol{\mathcal{L}}} ^{*}  \rangle + \epsilon _k
$,
where ${\boldsymbol{\mathcal{P}}}_{k}$ is the sampler that uniformly  and randomly extracts one entry of ${\boldsymbol{\mathcal{L}}} ^{*}$, and $\epsilon _k$ denotes additive noise.  
Then, we can obtain its dithered counterpart via $y_k + \xi_k$,
where   $\xi_k$ refers to the dither generated from a uniform distribution  on  $[-\frac{\delta}{2}, \frac{\delta}{2}]  $ with some $\delta>0$.

(II) \textbf{Quantization step:} In this step,  we  apply quantizer
$\operatorname{Q}_{\delta} (\cdot)$  to all dithers  $y_k + \xi_k$. 
Then, 
we attain
the resulting 
 dithered quantized measurements,
i.e., $q_k=\operatorname{Q}_{\delta}  (y_k + \xi_k)$.
As presented in existing 
literature \cite{chen2023quantized,  hou2025robust, he2025robust, hou2025tensor},
the quantizer $\operatorname{Q}_{\delta} (\cdot)$ utilized in this paper
is set to be uniform scalar quantizer
$\operatorname{Q}_{\delta} (\cdot):=  \delta ( \lfloor \cdot / \delta \rfloor + \frac{1}{2}) $ with resolution $\delta>0$.

The following
Theorem  \ref{dither-unfo} demonstrates that the uniformly dithered quantized measurements exhibit statistical equivalence to the original data.
\begin{Theorem}  \label{dither-unfo} \cite{hou2025tensor} 
For quantizer
$\operatorname{Q}_{\delta} (\cdot):=  \delta ( \lfloor \cdot / \delta \rfloor + \frac{1}{2}) $ with resolution $\delta>0$, and  dither  $ \xi \thicksim \operatorname{Unif} ([-\frac{\delta}{2}, \frac{\delta}{2}] )$, we have
\begin{align}
\mathbb{E} \big[ \operatorname{Q}_{\delta} (x+\xi) \big]
= \mathbb{E}
\Big [ \delta \big ( \lfloor \frac{x+\xi} { \delta}  \rfloor + \frac{1}{2} \big) \Big ]=x,
\end{align}
for any $x \in \mathbb{R}$, where the expectation is taken over $\xi$.
\end{Theorem}

\section{\textbf{Fast Randomized Compression Algorithms for  Large-Scale High-Order Tensor}} \label{randomized-lrtc} \label{fast-fctn-dec}

In this section, we devise fast and efficient randomized compression algorithms for large-scale high-order  tensor.
 These compression techniques serve as the computational acceleration core of our proposed recovery algorithm framework.


\begin{algorithm}[!htbp]
\setstretch{0.5}
     \caption{Fixed-Rank    Randomized Compression. 
     }
     \label{tucker-utv}
      \KwIn{$\bm{\mathcal{X}}      \in\mathbb{R}^{I_1\times   \cdots \times I_N}$,
      target rank:        $\bm{r}=(r_1, \cdots, r_N)$, sketch size:  $\bm{l}=(l_1, \cdots, l_N)$,
       processing order:  $\bm{\rho}$,      oversampling parameter: $p$, power iteration: $q$.
       }
      {\color{black}\KwOut{$  \hat{\bm{\mathcal{X}}}= [\bm{\mathcal{C}};  \bm{F}_{1}, \bm{F}_{2},\cdots, \bm{F}_{N}] $.
      }}

Set $\bm{\mathcal{C}} ^{(0)}  \leftarrow  \bm{\mathcal{X}} $\;

      \For{$v=1,2,\cdots, \operatorname{\textit{length}} (\bm{\rho})$}
      {
Set   $ \bm{A} \leftarrow  \bm{C}_{ (\bm{\rho} _{v}) } ^{({v} -1) }$,
 $[m, n] =\operatorname{\textit{size}} (\bm{A}  )$\;

Initialize $\bm{l}_{\bm{\rho} _{v}}$: $\bm{r}_{\bm{\rho} _{v}}+p  \leq \bm{l}_{\bm{\rho} _{v}} \leq  \min\{m,n \}$, and
draw      a  random sketching  matrix  ${\boldsymbol{{G}}}   ^{(v)}   \in\mathbb{R}^{ z_v  \times \bm{l}_{\bm{\rho} _{v}}   }$\;

     {  Let $\bm{T}_{2}=\bm{G} ^{(v)}$ \;

       \For{$ j=1: q+1$}
    {
     $ \boldsymbol{{T}}_{1}=
     {\boldsymbol{{A}}}
     \cdot \boldsymbol{{T}}_{2}
    $,
   $
    \boldsymbol{{T}}_{2}=
  \boldsymbol{{A}} ^{\mit{T}}  \cdot \boldsymbol{{T}}_{1}  $\;
   }

Do \textit{QR factorizations}:  $ \boldsymbol{{T}}_{1}=    {\boldsymbol{{Q}}}_{1} \boldsymbol{{R}}_{1}
   $,
   $    \boldsymbol{{T}}_{2}=    {\boldsymbol{{Q}}}_{2}      \boldsymbol{{R}}_{2}    $; \\

    Compute $\boldsymbol{{D}}=     { {\boldsymbol{{Q}}}_{1} ^{\mit{T}}  }    \boldsymbol{{A}}     \boldsymbol{{Q}}_{2}$,  or
     $\boldsymbol{{D}}=  { {\boldsymbol{{Q}}}_{1} ^{\mit{T}} }  \boldsymbol{{T}}_{1}    ( { {\boldsymbol{{Q}}}_{2} ^{\mit{T}}  } 
     \boldsymbol{{T}}_{2}) ^ {\dagger}$ \;

     Compute a QRCP:  $\boldsymbol{{D}}= \widetilde{\boldsymbol{{Q}} }   \widetilde{\boldsymbol{{R}}} \widetilde{ \boldsymbol{{P}}} ^{\mit{T}} $ \;

Form the UTV-based low-rank approximation:
$
\boldsymbol{{U}} ^{(v)} \longleftarrow  \boldsymbol{{Q}}_{1}  \cdot \widetilde{\boldsymbol{{Q}} },
\boldsymbol{{T}} ^{(v)} \longleftarrow  \widetilde{\boldsymbol{{R}} },
\boldsymbol{{V}} ^{(v)} \longleftarrow \boldsymbol{{Q}}_{2}  \cdot \widetilde{\boldsymbol{{P}} }
$ ;\\ }

Update
  $\bm{F}_{\bm{\rho} _{v}}   \leftarrow      \boldsymbol{U} ^{(v)} (:, 1:\bm{r}_{\bm{\rho} _{v}})$,
  $\bm{C}_{(\bm{\rho} _{v})} ^{({v} ) }   \leftarrow  (\bm{F}_{\bm{\rho} _{v}}) ^{\mit{T}}  \bm{C}_{(\bm{\rho} _{v})}   ^{({v} -1) }$\;
  $  \bm{\mathcal{C}}^{(v)} \leftarrow \bm{C}_{(\bm{\rho} _{v})}   ^{({v} ) }$, in tensor format.
}

 $\hat {\bm{\mathcal{X}} } =  {\boldsymbol{\mathcal{C}}}   {{\times}}_{1}  {{\bm{F}}_{1}}
 {{\times}}_{2}  {{\bm{F}}_{2}}  \cdots  {{\times}}_{N}  {{\bm{F}}_{N}}  $;
    \end{algorithm}

\subsection{\textbf{Proposed Fixed-Rank Randomized  Algorithm}}
In this subsection, based on  full-mode random projection, UTV factorization,  and subspace  iteration techniques,   we put forward an efficient   randomized algorithm for representation and compression of large-scale high-order tensors.

\textbf{Algorithm Description:}
At each stage of Algorithm  \ref{tucker-utv},  the core tensor $\bm{\mathcal{C}}$ is unfolded, and the factor matrix $\bm{F}_{\bm{\rho} _{v}}$ is formed by taking the first $r_{\bm{\rho} _{v}}$ left singular vectors. The new core tensor  $\bm{\mathcal{C}}^{(v)}$
is obtained by projecting the previous core tensor onto the subspace spanned by the columns of $\bm{F}_{\bm{\rho} _{v}}$.
We then have a $v$-th partial  approximation, defined as $ \hat{\bm{\mathcal{X}}} ^{(v)}   =   \bm{\mathcal{C}}^{(v)}    {\times}_{i=1}^{v} \bm{F}_{{\bm{\rho} _{i}}} $. For convenience, given a processing order  $\bm{\rho}:=[1,2,3,\cdots, N]$,
 we define
 \begin{equation*}
  z_1=I_2  \cdots I_d,    z_j=
 \bigg ( \prod_{ i= \bm{\rho}_{1} }      ^  { \bm{\rho}_{j-1}  }  r_i \bigg )
\bigg (   \prod_{i=\bm{\rho}_{j+1} } ^  {\bm{\rho}_{N}}  I_i \bigg), \; j=2, \cdots, N.
 \end{equation*}
For tensors whose  singular values of unfolded matrices exhibit a certain degree of decay,
the proposed Algorithm  \ref{tucker-utv}   may be sufficiently accurate  when $q=0$.
However,  in applications where the unfolding  matrices of the target tensor  display a slowly decaying singular values,
it may 
obtain the approximation result that deviate significantly from the exact ones (computed by the THOSVD \cite{kolda2009tensor} and
  STHOSVD \cite{vannieuwenhoven2012new}). Thus, we incorporate $q$ steps of a power iteration  to improve
the accuracy of the proposed randomized compression algorithm in these circumstances.

Below, we mainly clarify the differences between our proposed algorithm and existing related methods  \cite{minster2020randomized, dong2023practical}.
\begin{itemize}
  \item
When performing compressed representation for each mode of the target tensor, the proposed method
leverages a sketch of the core tensor's  unfolding matrix   in order to project it onto its row space, i.e.,
\textit{Line $7$ in  Algorithm  \ref{tucker-utv}}.
    This i) significantly improves the quality of the approximate basis $   \boldsymbol{{Q}}_{2}$ and,
    as a result, the quality of the approximate right singular subspace of $\bm{A}$ compared to that
    of two-sided sketching based STHOSVD  method \cite{dong2023practical}, which utilizes      a random matrix for the projection,
    and ii) allows
    $\boldsymbol{{D}}_{\operatorname{approx}}=    { {\boldsymbol{{Q}}}_{1} ^{\mit{T}} }    \boldsymbol{{T}}_{1}
    ( { {\boldsymbol{{Q}}}_{2} ^{\mit{T}}  }       \boldsymbol{{T}}_{2}) ^ {\dagger}$
     to provide a highly accurate approximation to
    $\boldsymbol{{D}}=  { {\boldsymbol{{Q}}}_{1} ^{\mit{T}}  }      \boldsymbol{{A}}    \boldsymbol{{Q}}_{2}$.

     \item
     The proposed method applies a column-pivoted QR decomposition to small-size compressed matrix $\bm{D}$, i.e.,
     \textit{Line $11$ in  Algorithm  \ref{tucker-utv}}.
     Whereas, the R-STHOSVD \cite{minster2020randomized} utilizes a truncated SVD       to decompose the compressed matrix.
     Compared to the latter, the former  can boost the  computational efficiency.

     \end{itemize}

\begin{algorithm}[!htbp]
\setstretch{0.3}
\caption{Fixed-Accuracy Randomized Compression. 
}
     \label{bb-rsthosvdWWWWWW} 
      \KwIn{$\bm{\mathcal{X}}\in\mathbb{R}^{I_1\times \cdots\times I_N}$, processing order: $\bm{\rho}$,
      error tolerance:   $\epsilon$,   block size: $b$,  
      power parameter: $q$.
       }
 {\color{black}\KwOut{ \textcolor[rgb]{0.00,0.00,0.00}{$  \hat{\bm{\mathcal{X}}}= [\bm{\mathcal{C}};  \bm{F}_{1}, \bm{F}_{2},\cdots, \bm{F}_{d}]
 $.}}}

Set $\bm{\mathcal{C}} ^{(0)} \leftarrow  \bm{\mathcal{X}} $\;

      \For{$v=1,2,\cdots, \operatorname{\textit{length}} (\bm{\rho})$}
      {

Set   $ \bm{A} \leftarrow \bm{C}_{(\bm{\rho} _{v})} ^{(v-1)}$,  $n=\operatorname{\textit{size}} (\bm{A} ,2)$\;

  ${\boldsymbol{{Y}} } ^{\{v\}}=[\;\;];\;   {\boldsymbol{{W}}}^{\{v\}}=[\;\;]$\;

$
E=   \| {{\boldsymbol{{A}}}} \|_{\mathnormal{F}} ^{2}, \;\; \operatorname{tol} \longleftarrow \epsilon^2$\;

\For{$i=1,2,3, \cdots$}
              {
               Generate  a  random sketching  matrix $\bm{{\Omega}}_{i} \in\mathbb{R}^{z_v \times  b }$, $ \alpha  \longleftarrow 0  $\;

               \For{$j=1,2, \cdots, q$}
               {
               $
{ {\boldsymbol{{W}}} }^{(i)} \longleftarrow {{\boldsymbol{{A}}}}  ^{\mit{T}}  {{\boldsymbol{{A}}}}  \bm{{\Omega}}_{i}
-  { {\boldsymbol{{W}}}}^{\{v\}}   {{\boldsymbol{{Z}}}}  ^{-1}   ( {{ {\boldsymbol{{W}}}}^{\{v\}}
 }  )^{\mit{T}} \bm{{\Omega}}_{i} - \alpha \bm{{\Omega}}_{i} $\;

$[  \bm{{\Omega}}_{i}, \hat{{\boldsymbol{{S}}}} , \sim]   =\operatorname{\textit{eigSVD}}(    { {\boldsymbol{{W}}} }^{(i)})$ \;
    \If{
$ j>1  $ and $\alpha < \hat{{\boldsymbol{{S}}}}(b,b)$ }
{
$\alpha  \longleftarrow  (\hat{{\boldsymbol{{S}}}}(b,b) +\alpha) /2$;
}

               }

             $  { {\boldsymbol{{Y}}} }^{(i)}= {{\boldsymbol{{A}}}}  \bm{{\Omega}}_{i}$,
            $ { {\boldsymbol{{W}}} }^{(i)} = {{\boldsymbol{{A}}}}  ^{\mit{T}}  { {\boldsymbol{{Y}}} }^{(i)} $
             \;

             $\boldsymbol{{Y}} ^{\{v\}}=[ \boldsymbol{{Y}}^{\{v\}}, {{\boldsymbol{{Y}}}^{(i)}}]$,
             $\boldsymbol{{W}} ^{\{v\}}=[ \boldsymbol{{W}}^{\{v\}}, {{\boldsymbol{{W}}}^{(i)}}]$\;

              $ { {\boldsymbol{{Z}}} } =
({{\boldsymbol{{Y}}}^{\{v\}}} ) ^{\mit{T}}  { {\boldsymbol{{Y}}} ^{\{v\}}}  $,
 $ { {\boldsymbol{{T}}} } =
({{\boldsymbol{{W}}}^{\{v\}}}  ) ^{\mit{T}}  { {\boldsymbol{{W}}}^{\{v\}} }  $
             \;

\If{
$ 
E - \operatorname{tr} ({\boldsymbol{{T}}}  { {\boldsymbol{{Z}}} } ^{-1} ) <    \epsilon ^2  $
}
{
break;
}

}

   $ [ \hat{  {\boldsymbol{{V}}}},
      \hat{  {\boldsymbol{{D}}}}      ]
     =\operatorname{\textit{eig}}(
     {\boldsymbol{{Z}}})
     $\;




$\bm{U} ^{\{v\}}=  {\boldsymbol{{Y}}}^{\{v\}}  \hat{  {\boldsymbol{{V}}}}  \Big (\sqrt{\hat{  {\boldsymbol{{D}}}} } \Big)^{-1} $\;

  Update
  $\bm{F}_{\bm{\rho} _{v}}   \leftarrow   \bm{U} ^{\{v\}} $,
  $\bm{C}_{(\bm{\rho} _{v})} ^{(v)}  \leftarrow  (\bm{F}_{\bm{\rho} _{v}}) ^{\mit{T}}  \bm{C}_{(\bm{\rho} _{v})} ^{(v-1)}$\;
  $  \bm{\mathcal{C}} ^{(v)}   \leftarrow \bm{C}_{(\bm{\rho} _{v})}  ^{(v)} $, in tensor format.
}
        \vspace{-0.0mm}
    \end{algorithm}

\subsection{\textbf{Proposed Fixed-Precision Randomized  Algorithm}}

The previous  algorithm   assume prior knowledge of the multi-linear rank.
However, in practice, estimating the multi-linear  rank of the target tensor can be challenging.
Thus, we introduce an   adaptive randomized algorithm  for large-scale tensor compression, where
the multi-linear rank is unknown.

\textbf{Algorithm Description:}
For a given large-scale tensor $\bm{\mathcal{X}}$, 
it is often desirable to effectively and accurately find a approximate decomposition $\hat{\bm{\mathcal{X}}}$ such that $\| \bm{\mathcal{X}}- \hat{\bm{\mathcal{X}}} \|_{\mathnormal{F}} \leq \epsilon \|\bm{\mathcal{X}} \|_{\mathnormal{F}}  $, where $0< \epsilon<1$ is a user-defined parameter.
To solve the aforementioned problem of fixed-precision approximation, several adaptive randomized algorithms
for large-scale tensors have been proposed in the literature \cite{minster2020randomized,che2025efficient333}.
The core idea of these existing methods is to first develop effective and  adaptive randomized  algorithm  in the   matrix format,
and then apply it  either separately or sequentially
to the   mode-unfolding matrices of large tensor.
Finally, they will return a compressed small-scale core tensor and factor matrices.


Our proposed   fixed-accuracy randomized  method is presented in 
Algorithm \ref{bb-rsthosvdWWWWWW}, 
which is developed by the  novel incremental QB approximation for enhanced parallel efficiency
and the shifted power iteration  for better accuracy.


\begin{itemize}
  \item \textbf{Incremental QB approximation imposed on mode-$v$ unfolding $\bm{C}^{(v-1)}_{(v)}$:}
For each $v$, $v \in \{1,2, \cdots, N\}$,  assume that  $\bm{\Omega}  \in \mathbb{R} ^{z_v \times k}$
is a random sketching matrix,
let  $ \bm{A}:=  \bm{C}^{(v-1)}_{(v)} \in \mathbb{R} ^{I_v \times z_v}  $,
$\bm{Y}= \bm{A} \bm{\Omega}$, $\bm{W}= \bm{A}^{\mit{T}}   \bm{Y}$, and the economic  SVD of $\bm{Y}$ be
$\bm{Y}=\hat{  {\boldsymbol{{U}}}}   \hat{  {\boldsymbol{{\Sigma}}}}     \hat{  {\boldsymbol{{V}}}}  ^{\mit{T}}$.
Then, setting
\begin{align}\label{newqbqbqbqbqbqb}
{\boldsymbol{{Q}}} = {\boldsymbol{{Y}}}\hat{  {\boldsymbol{{V}}}} \hat{  {\boldsymbol{{\Sigma}}}}^{-1},
{\boldsymbol{{B}}} = ({\boldsymbol{{W}}}\hat{  {\boldsymbol{{V}}}} \hat{  {\boldsymbol{{\Sigma}}}}^{-1}) ^{\mit{T}}.
\end{align}
And, if $\operatorname{tr}(\cdot)$ denotes the trace of a matrix,
\begin{align} \label{qrqrerror}
\| \bm{A} - {\boldsymbol{{Q}}} {\boldsymbol{{B}}} \|_{\mathnormal{F}} ^{2}
= \|\bm{A} \|_{\mathnormal{F}} ^{2} -\operatorname{tr}({\boldsymbol{{W}}}  ^{\mit{T}}  {\boldsymbol{{W}}}   ({\boldsymbol{{Y}}}  ^{\mit{T}}  {\boldsymbol{{Y}}})^{-1}   ).
\end{align}
The novel  QB approximation  mentioned above  
 enables us to calculate $\bm{Y}= \bm{A} \bm{\Omega}$, $\bm{W}= \bm{A}^{\mit{T}}   \bm{Y}$ incrementally to generate
the matrices ${\boldsymbol{{Q}}}$ and ${\boldsymbol{{B}}}$ at the end of process.
This strategy differs from the previous $\textit{{randQB}-{EI}}$ approach 
proposed in \cite{yu2018efficient1}, 
   which computes ${\boldsymbol{{Q}}}$ and ${\boldsymbol{{B}}}$ incrementally with multiple QR decompositions.
  Besides, a new approach can be derived   for evaluating the approximation error in the Frobenius norm according to (\ref{qrqrerror}).


  \item
\textbf{Shifted power iteration scheme:}
Under the above QB approximation framework, a more effective power iteration strategy can be developed.
Specifically, 
${\boldsymbol{{W}}} _i=\bm{H}^{\mit{T}} \bm{H} \bm{\Omega}_i $ is computed in each step of the power iteration,
where  $\bm{\Omega}_i$  denotes small-block random matrix,
 ${\boldsymbol{{H}}} = {\boldsymbol{{A}}} -  {\boldsymbol{{Q}}} {\boldsymbol{{B}}}$.
In virtue of Equation (\ref{qrqrerror}) and the formulation of $\bm{W}=\bm{A}^{\mit{T}}   \bm{Y}, \bm{Y}=\bm{A} \bm{\Omega}, \bm{Z}= \bm{Y}^{\mit{T}}   \bm{Y}$,
$\bm{A}-  {\boldsymbol{{Q}}} {\boldsymbol{{B}}}$ can be denoted by
 $\bm{A}-   \bm{Y}   {\boldsymbol{{Z}}} ^{-1}  \bm{W}^{\mit{T}} $. Thus, combining
 ${\boldsymbol{{T}}} _i \longleftarrow
\bm{A} \bm{\Omega}_i - \bm{Y}   {\boldsymbol{{Z}}} ^{-1}  \bm{W}^{\mit{T}}  \bm{\Omega}_i $,
${\boldsymbol{{W}}} _i \longleftarrow
\bm{A}^{\mit{T}}  \bm{T}_i -\bm{W} {\boldsymbol{{Z}}} ^{-1}    \bm{Y}     ^{\mit{T}}  \bm{T}_i $, and
${\boldsymbol{{W}}} =\bm{A}^{\mit{T}} \bm{Y}  $, we can derive that
  $$
{ {\boldsymbol{{W}}} }_{i} 
\longleftarrow {{\boldsymbol{{A}}}}  ^{\mit{T}}  {{\boldsymbol{{A}}}}  \bm{{\Omega}}_{i}
-  { {\boldsymbol{{W}}}}   {{\boldsymbol{{Z}}}}  ^{-1}   ( {{ {\boldsymbol{{W}}}}
 }  )^{\mit{T}} \bm{{\Omega}}_{i}.
 $$
 To further improve the approximation accuracy, the shift module   is dynamically introduced, thus leading to the shifted power iteration scheme
 \cite{feng2023fast}, \cite{feng2024algorithm}, i.e.,
  $$
{ {\boldsymbol{{W}}} }_{i} 
\longleftarrow {{\boldsymbol{{A}}}}  ^{\mit{T}}  {{\boldsymbol{{A}}}}  \bm{{\Omega}}_{i}
-  { {\boldsymbol{{W}}}}   {{\boldsymbol{{Z}}}}  ^{-1}   ( {{ {\boldsymbol{{W}}}}
 }  )^{\mit{T}} \bm{{\Omega}}_{i}- \alpha \bm{{\Omega}}_{i}.
 $$
\end{itemize}
%
The two modules involved in the proposed algorithm essentially replace the QR decomposition in the $\textit{{randQB}-{EI}}$ 
method \cite{yu2018efficient1}
with some matrix skills 
to enhance parallel efficiency.
When the power parameter is set to be  $q=0,1,2$,
the proposed algorithm is  actually  equivalent to the adaptive R-STHOSVD algorithms \cite{minster2020randomized,che2025efficient333} using $\textit{{randQB}-{EI}}$ technique.
%
For $q>2$, the shifted parameter $\alpha$ is dynamically 
updated
to enhanced accuracy.


\subsection{\textbf{Theoretical Analysis on  
Error Upper Bounds}}
For the sake of conciseness in the proof, the processing order is set to be $\bm{\rho}:=[1,2,\cdots, N]$.
Without loss of generality, let $I^{'}_v= \min\{ I_v,  r_v \cdots  r_{v-1} I_{v+1}\cdots  I_{N} \} , v=1,2, \cdots, N  $.
In order to facilitate the representation of the two dimensions of intermediate matrix $\bm{C}_{ ({v}) } ^{({v} -1)}$,  we define
 \begin{equation*}
  z_1=I_2 I_3 \cdots I_d, \;\;
 z_j=  \bigg ( \prod_{ i=  {1} }      ^  {  {j-1}  }  r_i \bigg ) \bigg (   \prod_{i= {j+1} } ^  { {N}}  I_i \bigg), \;\; j=2, \cdots, N.
 \end{equation*}

For each $v$,  the \textit{Singular Value Decomposition}  (SVD) of $\bm{C}_{ ({v}) } ^{({v} -1)}
=\bm{U}^  {(v)}  \bm{\Sigma}^ {(v)} (\bm{V}^ {(v)} ) ^{\mit{T}} $  is further denoted  as follows:
 \begin{equation} \label{svdmidd}
 \bm{C}_{ ({v}) } ^{({v} -1)}=
 {\begin{bmatrix} \bm{U}_{1} ^{(v)}   \bm{U}_{2} ^{(v)}
 \\
\end{bmatrix} }
 {\begin{bmatrix}
\bm{\Sigma}_{1} ^{(v)} & \\
  &\bm{\Sigma}_{2} ^{(v)}\\
\end{bmatrix} }
{\begin{bmatrix}
\bm{V}_{1} ^{(v)}
 \\
  \bm{V}_{2} ^{(v)}
 \\
\end{bmatrix}^{\mit{T}} },
 \end{equation}
 $\bm{U}_{1} ^{(v)}=  \bm{U} ^{(v)} (:, 1:\bm{l}_{{v}}-p)$,
$\bm{U}_{2} ^{(v)}=  \bm{U} ^{(v)} (:, \bm{l}_{{v}}-p+1:I^{'}_v)$,
 $\bm{V}_{1} ^{(v)}=  \bm{V} ^{(v)}(:, 1:\bm{l}_{ {v}}-p)$,
$\bm{V}_{2} ^{(v)}=  \bm{V}^{(v)} (:, \bm{l}_{ {v}}-p+1:I^{'}_v  )$,
 $\bm{\Sigma}_{1} ^{(v)}=  \bm{\Sigma}^{(v)} ( 1: \bm{r}_{{v}}  , 1: \bm{r}_{{v}})$,
 and  $\bm{\Sigma}_{2} ^{(v)}=  \bm{\Sigma}^{(v)} (  \bm{r}_{{v}}  +1 : I^{'}_v, \bm{r}_{{v}} +1:I^{'}_v  )$.

 Furthermore,  we define $ \bm{\Psi}_{1}^{(v)} \in \mathbb{R} ^{(\bm{l}_{ {v}}-p) \times \bm{l}_{ {v}}}  $ and
  $ \bm{\Psi}_{2}^{(v)} \in \mathbb{R} ^{(I^{'}_v- \bm{l}_{ {v}}+p) \times \bm{l}_{ {v}}}  $   as follows:
  \begin{equation} \label{omega12}
  \bm{\Psi}_{1}^{(v)}=  (\bm{V}_{1} ^{(v)}) ^{\mit{T}} \cdot \bm{G} ^ { ({v}) }, \;\;
  \bm{\Psi}_{2}^{(v)}=  (\bm{V}_{2} ^{(v)}) ^{\mit{T}} \cdot \bm{G} ^ { ({v}) }.
  \end{equation}
  We assume that $\bm{\Psi}_{1}^{(v)}$ is full row rank and its pseudo-inverse satisfies
  $ 
  \bm{\Psi}_{1}^{(v)} \cdot (\bm{\Psi}_{1}^{(v)})^ {\dagger} =\bm{I}, \;\; v=1,2, \cdots, N 
  $.

 Based on the aforementioned preliminaries,  we now primarily focus on providing an error upper bound analysis for Algorithm \ref{tucker-utv}.
Please see Theorem \ref{laaaa} for more details.

 \begin{Theorem}\label{laaaa}
 Let  $  \hat{\bm{\mathcal{X}}}  =  \boldsymbol{\mathcal{C}} {\times}_{1} \bm{F}_{1}  {\times}_{2} \bm{F}_{2}   \cdots
 {\times}_{N} \bm{F}_{N}$ be the low multilinear rank-$(r_1, r_2, \cdots, r_N)$ approximation of
 an order-$N$  tensor   ${\boldsymbol{\mathcal{X}}} \in \mathbb{R}^{{I_1 \times I_2  \times \cdots \times I_N}}$
 by Algorithm  \ref{tucker-utv} with processing order   $\bm{\rho}:=[1,2,\cdots, N]$,   sketch size:  $\bm{l}=(l_1,l_2, \cdots, l_N)$,
  and oversampling parameter $p\geq 2$ satisfying
  $$
  r_j +p \leq  l_j \leq \min  \{I_j, z_j
   \}, \;  \; j=1,2,\cdots, N.
  $$
  For each $v$, assume that ${ \bm{\Psi} }_{1}^{(v)}$ is of full row rank.   Let
  ${ \hat{\boldsymbol{\mathcal{X}}}  }_{\operatorname{opt}}$  denote the optimal rank-$\bm{r}$ approximation,
   $ \tau_1  = \sigma _{{r}_v} ( \bm{C}_{ ({v}) } ^{({v} -1)} ) $,
$ \tau_2  = \sigma _{{l}_v-p+1} ( \bm{C}_{ ({v}) } ^{({v} -1)}) $,
$\tau_3  =  \sigma _{1} (\bm{C}_{ ({v}) } ^{({v} -1)}  ) $.
  Then, we have
   \begin{align}
   \label{utv-error}
\big\| {{\boldsymbol{\mathcal{X}}}}    - \hat{\bm{\mathcal{X}}} \big \|_{\mathnormal{F}}^2 &
\leq
\sum_{v=1}^{N} \bigg\{ \bigg( \sqrt{
  \frac{
   {(\alpha ^{(v)})}^{2}  \omega^{2} }
 {1+{(\beta ^{(v)})}^{2}    \omega^{2}   }  }
 +\sqrt{   \frac{   {(\eta ^{(v)})}^{2}   \omega^{2}}
 {1+{(\xi ^{(v)})}^{2}   \omega^{2}  }}
\notag \\
 & + \Big \| {{\boldsymbol{\mathcal{X}}}}-  { \hat{\boldsymbol{\mathcal{X}}}  }_{\operatorname{opt}}  \Big \|_{\mathnormal{F}} \bigg)^2+
 \frac{
   {(\alpha ^{(v)})}^{2}   \omega^{2}
 }
 {1+{(\beta ^{(v)})}^{2}   \omega^{2}  }
 \bigg \},
\end{align}
where $\alpha ^{(v)} = \sqrt{{r}_v} \frac{ \tau_2^2 }{ \tau_1}  ( \frac{ \tau_2 }{ \tau_1})^{2q}$ ($q$ denotes the power iteration parameter),
$\beta^{(v)} =   \frac{ \tau_2^2 }{ \tau_1 \tau_3} (\frac{ \tau_2 }{ \tau_1})^{2q}$, $\eta^{(v)} =   \frac{ \tau_1 } { \tau_2} \alpha ^{(v)}$,
$\xi^{(v)} =  \frac{1 } { \tau_2} \beta ^{(v)}$, and $ \omega_1  = \| { \bm{\Psi} }_{2}^{(v)} \| _{2}$,  $\omega_2 =  \|( { \bm{\Psi} }_{1}^{(v)} )  ^ {\dagger} \| _{2} $, $\omega=\omega_1 \omega_2$.
\end{Theorem}

Next,   we provide the  explicit error-upper-bounds  for the  proposed Algorithm  \ref{tucker-utv},
which, in contrast to the argument in Theorem \ref{laaaa}, depends on distributional assumptions on the random matrix
$\{ {\bm{G}^ {(v)} } \}_{v=1} ^{N}$.
Below, we mainly conduct theoretical analysis from different sketching  matrices $\{ {\bm{G}^ {(v)} } \}_{v=1} ^{N}$, such as
the standard Gaussian  matrices,   the uniform random  matrices, the Kronecker products of \textit{Subsampled Randomized Fourier Transform} (SRFT) matrices,  and  the \textit{Khatri-Rao} (KR)  products of the standard Gaussian matrices (or the uniform random matrices).

\begin{Theorem}(\textbf{Gaussian Sketching})
For each $v$, we assume that $ {\bm{G}^ {(v)} },  v=1, 2, \cdots, N$  is a standard Gaussian sketching  matrix.
With the notation of  Theorem \ref{laaaa}, we obtain the average error bound for Algorithm  \ref{tucker-utv}, i.e.,
    \begin{align}
  &  \mathbb{E} _{\{ {\bm{G}^ {(v) } } \} _{v=1}^{N}  } \big\| {{\boldsymbol{\mathcal{X}}}}    - \hat{\bm{\mathcal{X}}} \big \|_{\mathnormal{F}}
\leq  \bigg( \sum_{v=1}^{N}  \Big( \Big( \| {{\boldsymbol{\mathcal{X}}}}- { \hat{\boldsymbol{\mathcal{X}}}  }_{\operatorname{opt}}  \|_{\mathnormal{F}}+ \Big\{  \big(1+ \frac  {\tau_2} {\tau_1}  \big )
\notag \\
 &  \sqrt{ {r}_{v}}  \phi^{(v)}   \tau_2 {\big( \frac {\tau_2} {\tau_1} \big)}^{2q} \Big\} \Big )^2 +
{
  \frac{   {(\alpha ^{(v)})}^{2}    (\phi  ^{(v)})^{2}
 }
 {1+{(\beta ^{(v)})}^{2}   (\phi ^{(v)})^{2} } } \Big) \bigg) ^{\frac{1}{2}},
\end{align}
where
$\phi^{(v)} =  \frac {4e \sqrt{{l}_{v}} }{p+1}  (\sqrt{\min  \{I_j, z_j
   \}
- {l}_{v} +p }  +\sqrt{{l}_{v}} +7) $.
\end{Theorem}

\begin{Theorem} \label{kr-gauuniform}  (\textbf{KR-Gaussian Sketching})
For each $v$, $v\in \{1,  \cdots, N\}$,   we assume that $ {\bm{G}^ {(v)} }$ is the Khatri-Rao product of the  standard Gaussian matrices. By setting
$ \omega =\frac{\sqrt{\bm{l}_v} +  t \sqrt{ \prod_{m=1}^{v-1}(\sqrt{I_m} +t )    \prod_{m=v+1}^{N}(\sqrt{\bm{r}_m} +t )  } } { \sqrt{\bm{l}_v} -   t \sqrt{ \prod_{m=1}^{v-1}(\sqrt{\bm{r}_m} +t )    \prod_{m=v+1}^{N}(\sqrt{I_m} +t )  }}$
 in   (\ref{utv-error}),
 we obtain the  error bound for Algorithm  \ref{tucker-utv}
with probability at least
\begin{align*}
& 1- 2 \sum_{v=1}^{N} \Big \{ (\bm{l}_v -p ) \cdot \exp (-c_v t^2)   +
( {I'}_{v} -\bm{l}_v+p) \cdot \exp  \\ &  (-c'_v t^2 ) \Big \} -
2 \sum_{v=1}^{N} \sum_{k=1}^{\bm{l}_v }  \sum_{h \neq v} \exp(-c_{v,k, h} t^2 / K_{v,k,h}^2),
\end{align*}
for every $t \geq 0$.
Here, for each $v$, $k$, $h$ ($v \neq h$), $c_v$, $c'_v$, and $c_{v,k, h}$ are absolute constants,
$K_{v,k,h}$ satisfies the following equation:
\begin{eqnarray*}
K_{v,k, h}=
\begin{cases}
{\max_{i_v= 1,2, \cdots, \bm{r}_v }}   \| \mathbf{\Omega}_{v,h} (i_v, k)  \|_{\psi_2},  \;\;\; \operatorname{if} h < v;  \\
{ \max_{i_v= 1,2, \cdots, I_v }}   \| \mathbf{\Omega}_{v,h} (i_v, k)  \|_{\psi_2},  \;\;\; \operatorname{if} h > v.
\end{cases}
\end{eqnarray*}
\end{Theorem}

\begin{Theorem} (\textbf{Uniform Sketching})
For each $v$, we assume that $ {\bm{G}^ {(v)} },  v=1, 2, \cdots, N$  is a uniform  sketching  matrix.
 By setting    $\omega= \omega_1  \omega_2$, $ \omega_1= a_{v,1} \sqrt{\max\{  {I'}_{v} -\bm{l}_v+p, \bm{l}_v  \}}$,
$ \omega_2=  c_{v,1} \sqrt{\bm{l}_v}$, $\bm{l}_v > \big(1+  1/ \ln(\bm{l}_v-p)\big ) (\bm{l}_v-p)$ in   (\ref{utv-error}),
we obtain the  error upper bound  for Algorithm  \ref{tucker-utv}
with probability at least
$$
1-  \sum _{v=1}^{N}  \Big \{  \exp (-a_{v,2} \cdot   {\max\{  {I'}_{v} -\bm{l}_v+p, \bm{l}_v  \}}) + \exp (-c_{v,2}   \cdot \bm{l}_v ) \Big \}.
$$
Here, for each $v$,  we define $a_{v,1}$, $a_{v,2}$, $c_{v,1}$, $c_{v,2}$ as  positive constants with $a_{v,1} = 6 \alpha_v \sqrt{a_{v,2} +4 } $ and $\alpha_v >1$.
\end{Theorem}

\begin{Theorem} \label{kr-uniform}  (\textbf{KR-Uniform Sketching})
For each $v$, $v\in \{1, 2, \cdots, N\}$,   we assume that $ {\bm{G}^ {(v)} }$
is the Khatri-Rao product of the uniform random matrices.
By setting   $\omega= \omega_1  \omega_2$, $ \omega_1=  \sqrt{\bm{l}_v} + C_v  \sqrt{ {I'}_{v} -\bm{l}_v+p } +t$,
$ \omega_2=  \frac {1} { \sqrt{\bm{l}_v} -    C'_{v} \sqrt{\bm{l}_v-p}-t  }$  in   (\ref{utv-error}),
we obtain the  error bound for Algorithm  \ref{tucker-utv}
with probability at least
$$
1- 2 \sum_{v=1}^{N} \Big \{ \exp (-c_v t^2) + \exp (-c'_v t^2 ) \Big \},
$$
for every $t \geq 0$. Here, for each $v$,  $C_{v}=C_{K_{v}}$ and $c_{v}=c_{K_{v}}\geq0$ depend only on the  sub-Gaussian  norm
 $$
 K_{v}=\max_{j=1,2, \cdots, \bm{l}_v}  \|(\bm{\Psi}_{2}^{(v)} ) ^{\mit{T}}    (j, :)\|_{\psi_2}=
 \max_{j=1,2, \cdots, \bm{l}_v}  \| \bm{\Psi}_{2}^{(v)}     (:, j)\|_{\psi_2},
 $$
  and
  $C_{v}'=C_{K_{v}'}$ and $c_{v}'=c_{K_{v}'}\geq0$ depend only on the  sub-Gaussian  norm
  $$
  K_{v}'=\max_{j=1,2, \cdots, \bm{l}_v}  \|(\bm{\Psi}_{1}^{(v)} ) ^{\mit{T}}    (j, :)\|_{\psi_2}=
 \max_{j=1,2, \cdots, \bm{l}_v}  \| \bm{\Psi}_{1}^{(v)}     (:, j)\|_{\psi_2}.
  $$
\end{Theorem}

\begin{Theorem} \label{kr-gauuniform}  (\textbf{Kronecker-SRFT Sketching})
For each $v$, $v\in \{1, 2, \cdots, N\}$,   we assume that $ {\bm{G}^ {(v)} }$ is
the  Kronecker products of SRFT matrices.
By setting   $\omega=  \sqrt{ {\kappa_1^{(v)} z_v}  / {l_v} }$ in   (\ref{utv-error}),  we obtain the  error bound for Algorithm  \ref{tucker-utv}
with probability at least
$
1- 2 \sum_{v=1}^{N}  {1}/ {\kappa_2^{(v)}}
$.
Here,  the  sequences ${   \kappa_1^{(v)} } >1 $ and  $ { \kappa_2^{(v)} } >1 $ ($v \in \{1,2,\cdots, N\}$) satisfy
\begin{equation*}
\min_{v,h} \{S_{v,h}\} \geq \frac{ (\kappa_1^{(v)})^2  \kappa_2^{(v)}
}{(\kappa_1^{(v)}-1)^2}[(\bm{l}_{ {v}}-p)^2+(\bm{l}_{ {v}}-p)],
\end{equation*} where
$z_v = r_1 r_2 \cdots r_{v-1} I_{v+1} \cdots I_N $,  $\bm{l}_v = \prod_{h=1}^N S_{v,h}$.
\end{Theorem}

\section{\textcolor[rgb]{0.00,0.00,0.00}{\textbf{Generalized Nonconvex  Approach for 
Multi-Dimensional Data 
Modeling}}}  \label{model}

In this section, under the framework of FCTN decomposition,  we develop a novel generalized nonconvex gradient-domain regularization approach
tailored for 
high-order tensors. 
This regularization strategy lays the foundation for the model establishment of
subsequent 
multi-dimensional data recovery.

\subsection{\textbf{Generalized Nonconvex Regularizers}} \label{nonconvex-regulari}

\subsubsection{\textbf{\textcolor[rgb]{0.00,0.00,0.00}{Novel  regularizer encoding  prior structures}}}
In many application scenarios, the large-scale multi-dimensional data 
 to be estimated are not only with low-rankness but also possess
significant smooth structure.
As exemplified in the supplementary materials, 
the 
color images/videos, hyperspectral images/videos, MRI datasets and face datasets
represented by 
high-order tensors fromat
exhibit strong global low-rankness ($\textbf{L}$)
due to the strong sparsity of singular values obtained through generalized tensor unfolding,
as well as strong local smoothness ($\textbf{S}$) resulting from the strong sparsity of their gradient tensors.
 Therefore, a natural idea is  how to simultaneously promote these two types of equivalent sparsity through a concise yet effective
 regularization method, so as to deeply mine insightful $\textbf{L}$+$\textbf{S}$ prior features latent in 
 multi-dimensional
 data.
{In response to the aforementioned, 
  we consider   investigating  novel  generalized  nonconvex gradient-domain regularizer   within the   FCTN decomposition,
  which  can   simultaneously encode  essential
  $\textbf{L}$+$\textbf{S}$ prior features underlying high-order tensors.  
  }
  %

We begin by stating a few definitions and lemmas that are utilized later on.

\begin{Definition}\label{def10diff} \textbf{(Gradient tensor)}
For ${\boldsymbol{\mathcal{A}}} \in \mathbb{R}^{ I_1\times \cdots \times I_N} $, its gradient tensor along the $t$-th mode is defined as
\begin{align}\label{gradient-tensor}
 {\boldsymbol{\mathcal{G}}}_{ t } &:= \nabla_{t} ({\boldsymbol{\mathcal{A}}})=
\boldsymbol{\mathcal{A}} \;{\times}_{t} \;\bm{D}_{I_t}, \;\; t=1,2,\cdots, N,
\end{align}
where $\bm{D}_{I_t}$ is a row circulant matrix of $(-1,1, 0, \cdots,0)$,
\textcolor[rgb]{0.00,0.00,0.00}{$\nabla_{t} $ is defined as the corresponding difference operator along the $t$-th mode of tensor ${\boldsymbol{\mathcal{A}}}$, and $\boldsymbol{\mathcal{A}}{\times}_{t} \bm{D}_{I_t}$ denotes the mode-$t$  product of tensor $\boldsymbol{\mathcal{A}}$ with matrix  $\bm{D}_{I_t}$.}
\end{Definition}

\begin{Definition} 
\cite{zheng2021fully}
\label{resh}
Supposing that $\bm{\mathcal{X}} \in \mathbb{R}^{I_1 \times I_2 \times \cdots \times I_N}$ is an order-${N}$ 
tensor and the vector  $\textbf{n}$ is a  specified rearrangement of the vector $[1,2,\cdots,N]$.
Then,  the generalized tensor unfolding of $\bm{\mathcal{X}}$ is
defined as a matrix
$\textbf{X}_{[\textbf{n}_{1:d};\textbf{n}_{d+1:N}]}=\rm{reshape} $$(\vec{\bm{\mathcal{X}}}^{\textbf{n}},
\prod_{i=1}^d I_{\textbf{n}_i}, \prod_{i=d+1}^N I_{\textbf{n}_i})$.
The corresponding inverse operation is defined as $\bm{\mathcal{X}}=\Upsilon (\textbf{X}_{[\textbf{n}_{1:d};\textbf{n}_{d+1:N}]})$.
\end{Definition}

The following Lemma indicates that the FCTN-rank can bound the rank of all generalized tensor unfolding.
\begin{Lemma}
\cite{zheng2021fully}
Supposing that an order-$N$  tensor $\bm{\mathcal{X} } \in \mathbb{R}^{I_1 \times I_2 \times \cdots \times I_N}$ can be represented by
equation (\ref{fctndec}), then the following inequality holds: 
\begin{equation}
{\rm{Rank}} (\textbf{X}_{[\textbf{n}_{1:d};\textbf{n}_{d+1:N}]}) \leq \prod_{i=1}^d \prod_{j=d+1}^N R_{\textbf{n}_i,\textbf{n}_j},
\end{equation}
where
$R_{\textbf{n}_i,\textbf{n}_j}=R_{\textbf{n}_j,\textbf{n}_i}$  $ {\text{if}} ~ \textbf{n}_i > \textbf{n}_j$,
and $(\textbf{n}_1, \textbf{n}_2, \cdots, \textbf{n}_N)$ is a reordering of the vector $(1,2, \cdots, N)$.
\end{Lemma}


Below, we formally introduce the proposed 
 gradient-domain regularizer named  \textit{FCTN-based  Generalized  Nonconvex  Tensor Correlated Total Variation} (FCTN-GNTCTV).

{For any 
 ${\boldsymbol{\mathcal{X}}} \in \mathbb{R}^{ I_1\times I_2 \times \cdots \times I_N} $,
let $\Gamma$ represent a priori set consisting of directions along which ${\boldsymbol{\mathcal{X}}} $ equips \textbf{L}+\textbf{S} priors, and ${\boldsymbol{\mathcal{G}}}_{t}, t \in  \Gamma $  denote the gradient tensor along the $t$-th mode of ${\boldsymbol{\mathcal{X}}}$.
Then, the FCTN-GNTCTV regularizer  is defined as follows:}
\begin{align} \label{unified-regulare22}
\| {\boldsymbol{\mathcal{X}}}\|_{ \mathfrak{FN} }   &:=  \frac{1}{\gamma} \sum_{t \in   \Gamma} \Big \{
\sum_{k=1}^{\bar{N}} \alpha_k \cdot \big \| {\textbf{G}_{t}} _{[\textbf{n}^k_{\textbf{1}};\textbf{n}^k_{\textbf{2}}]} \big\|_\Phi \Big\},
\notag \\
& =\frac{1}{\gamma} \sum_{t \in   \Gamma}  \bigg  \{ \sum_{k=1} ^{ \bar{N}} \alpha_k \cdot  \sum_{i}
\Phi \Big(  \sigma_{i}  \big( {\textbf{G}_{t}} _{[\textbf{n}^k_{\textbf{1}};\textbf{n}^k_{\textbf{2}}]} \big ) \Big)
\bigg\},
\end{align}
\textcolor[rgb]{0.00,0.00,0.00}{where $ \{\alpha_{ k} \} _{k=1} ^{\bar{N} }$  are the non-negative weights with $  \sum_{k}  \alpha_{k}  =1$,
 $\gamma := \sharp\{\Gamma\}$ equals to the cardinality of $\Gamma$,
where $\textbf{n}^k$ is the $k$-th permutation   of the vector $[1,2,\cdots,N]$, $\textbf{n}^k_{\textbf{1}}=\textbf{n}^k_{1:\lfloor N/2 \rfloor}$, $\textbf{n}^k_{\textbf{2}}=\textbf{n}^k_{\lfloor N/2 \rfloor+1:N}$,
$$
 \bar{N}=   \begin{cases}
 C _N^{\lfloor N/2 \rfloor},  \;\; \;\;\; \;  \textit{if}  \;  $N$  \;  \textit{is} \;  \textit{odd},  \\
 C _N^{\lfloor N/2 \rfloor}/2,    \;\; \;  \textit{if}  \;  $N$  \;  \textit{is}  \;  \textit{even},
\end{cases}$$
and $\lfloor \cdot \rfloor$ denotes the floor function.}
The generalized nonconvex function   {$\Phi (\cdot): \mathbb{R}  \rightarrow \mathbb{R} $  satisfies the following assumptions:}

\textbf{(I)}: \textcolor[rgb]{0.00,0.00,0.00}{$\Phi (\cdot)$: $\mathbb{R} \rightarrow \mathbb{R}$ is proper, lower semi-continuous and symmetric with respect to y-axis;}

\textbf{(II)}:  \textcolor[rgb]{0.00,0.00,0.00}{$\Phi (\cdot)$ is concave and monotonically increasing  on $[0,\infty)$  with  $\Phi(0)=0 $.}

\begin{Remark}
Many popular nonconvex penalty functions $\Phi (\cdot)$ satisfy  the above assumptions, such as firm function \cite{gao1997waveshrink}, logarithmic (Log) function  \cite{gong2013general},  $\ell_{q}$ function \cite{marjanovic2012l_q},
capped-$\ell_{q}$ function \cite{li2020matrix, pan2021group},
smoothly clipped absolute deviation (SCAD) function \cite{fan2001variable1 },
and  minimax concave penalty  (MCP) function \cite{zhang2010nearly}.  \textcolor[rgb]{0.00,0.00,0.00}{Thus,  in the above  gradient-domain regularizer (\ref{unified-regulare22}), we  effectively employ a family of nonconvex functions on the singular values of all
 generalized unfolding matrices derived from  each  gradient tensor.}
\end{Remark}

Below, we 
provide a mathematical explanation to elucidate how the proposed regularization method can simultaneously encode global low-rankness and local smoothness priors.
\begin{Remark}
For 
 ${\boldsymbol{\mathcal{X}}} \in \mathbb{R}^{ I_1\times I_2 \times \cdots \times I_N} $ with
 FCTN-rank $(R_{1,2}, R_{1,3},\cdots, R_{1,N},R_{2,3},R_{2,4},\cdots, R_{2,N}, \cdots, R_{N-1,N})$,
it can be verified that
\begin{align}
R-1 
\leq
\operatorname{Rank} {({\textbf{G}_{t}} _{[\textbf{n}^k_{\textbf{1}};\textbf{n}^k_{\textbf{2}}]})} \leq
R, 
\end{align}
where
$R=\prod_{i=1}^ {\lfloor N/2 \rfloor} \prod_{j=\lfloor N/2 \rfloor+1}^N R_{\textbf{n}^k_i,\textbf{n}^k_j}$.
This reveals 
that the low-rankness between the original and resulting gradient tensors are consistent,
demonstrating that the FCTN-GNTCTV regularizer can indirectly induce the expected \textbf{L}-prior feature of the
 original  tensor like a low-rank regularizer.

 From the perspective of \textbf{S}-prior encoding, we can also verified that
 \begin{align} \label{s-encoding-ex}
 \| {\boldsymbol{\mathcal{X}}}\|_{ \operatorname{TV}, \Phi }  \lesssim
 \| {\boldsymbol{\mathcal{X}}}\|_{ \mathfrak{FN} }
  \lesssim
  \sqrt{R}
  \| {\boldsymbol{\mathcal{X}}}\|_{ \operatorname{TV}, \Phi }.
 \end{align}
 Here, $\| \cdot \|_{ \operatorname{TV}, \Phi }$ can be interpreted as a nonconvex extension of the TV-norm,
which is  defined as
$$ \| {\boldsymbol{\mathcal{X}}}\|_{ \operatorname{TV}, \Phi } :=
 \sum_{t \in   \Gamma} \| \nabla_{t} ({\boldsymbol{\mathcal{A}}}) \| _{ \ell_1, \Phi } =
 \sum_{t \in   \Gamma} \| \Phi   \odot \nabla_{t} ({\boldsymbol{\mathcal{A}}}) \| _{ 1  },
 $$ where $ \odot $ denotes element-wise operation.
 Besides, $\lesssim$ means that $ a \lesssim b  \Leftrightarrow a \leq C b $,
 $C$ is a fixed absolute constant.
 The inequality (\ref{s-encoding-ex}) means that FCTN-GNTCTV and nonconvex TV are compatible in sense of norm,
 indicating that the proposed regularizer can also indirectly extract the expected \textbf{S}-prior of the targeted tensors
 like a TV regularizer.
\end{Remark}

\subsubsection{\textbf{Novel noise/outliers
regularizer}}
\textcolor[rgb]{0.00,0.00,0.00}{To enhance the robustness against noise/outliers existed in  high-order format, we  define the following  nonconvex regularization penalty, i.e.,}
\begin{equation} \label{unified-regulare11}
\Upsilon( {\boldsymbol{\mathcal{E}}}) := \psi \big( h(  {\boldsymbol{\mathcal{E}}}  ) \big),
\end{equation}
where  $\psi (\cdot):\mathbb{R} \rightarrow \mathbb{R}$  is   a generalized nonconvex function,  which  has the same properties as $\Phi(\cdot)$ in (\ref{unified-regulare22}).
\textcolor[rgb]{0.00,0.00,0.00}{Here,  two types of corrupted  noise/outliers   are taken into account.}
When the  noise/outlier  tensor ${\boldsymbol{\mathcal{E}}}$   has tube-wise structure, and $h(\cdot)=\|\cdot\|_{{\mathnormal{F}},1}$ is defined as an $\ell_{{\mathnormal{F}},1}$-norm. Then, we have
\begin{equation} \label{unified-regular22222}  
\Upsilon( {\boldsymbol{\mathcal{E}}}) := \psi (\| {\boldsymbol{\mathcal{E}}}\|_{\mathnormal{F},1} ) =
 \sum_{{i_1}=1}^{I_1}  \sum_{{i_2}=1}^{I_2}
\psi
\big(
\|{\boldsymbol{\mathcal{E}}}_{i_1 i_2 :  \cdots: }
\|_{\mathnormal{F}} \big
).
\end{equation}
When the tensor ${\boldsymbol{\mathcal{E}}}$ is an entry-wise  noise/outlier tensor, $h(\cdot)=\|\cdot\|_{1}$ is defined as an $\ell_{1}$-norm. Then, we have
\begin{equation} \label{unified-regular22}
\Upsilon( {\boldsymbol{\mathcal{E}}})
:= \psi (\| {\boldsymbol{\mathcal{E}}}\|_{1} ) =
 \sum_{{i_1}=1}^{I_1}  \sum_{{i_2}=1}^{I_2}  \cdots \sum_{{i_N}=1}^{I_N}
\psi \big(
|{\boldsymbol{\mathcal{E}}}_{i_1 i_2 \cdots i_N}
|\big).
\end{equation}


%
\textcolor[rgb]{0.00,0.00,0.00}{Next,  we mainly revisit the commonly-used nonconvex sparsity-inducing penalties and the corresponding  proximal  mappings,
which play a central role in developing subsequent tensor recovery algorithms.}
 \textcolor[rgb]{0.00,0.00,0.00}{For a nonconvex penalty function $\psi(\cdot)$, 
  its proximity operator is defined as}
\begin{equation}\label{equ_gst1}
\operatorname{\textit{Prox}}_ {\psi, \mu} (v)= \arg \min_{x} \Big \{    \mu \cdot \psi(x)  + \frac{1}{2} {(x-v)}^2 \Big \},
\end{equation}
where $\mu>0$ is a penalty parameter.
\textcolor[rgb]{0.00,0.00,0.00}{In the supplementary materials, 
 we have summarized the proximity operators for  several popular nonconvex regularization penalties,
including firm-thresholding, $\ell_{q}$-thresholding,   MCP, Log,  and SCAD  penalties.}
On this basis, we can address the following two types of crucial minimization problems. That is,
\begin{align*}
  \operatorname{\textit{Prox}}_ {\psi, \lambda} (\boldsymbol{\mathcal{E}}) &=
\arg\min_{ \boldsymbol{\mathcal{E}}}
{\lambda} \cdot \Upsilon \big({\boldsymbol{\mathcal{E}}} \big)
 +
\frac
{1}{2}
\|
{\boldsymbol{\mathcal{E}}}-  {\boldsymbol{\mathcal{A}}}
\|^2_{\mathnormal{F}},
\\
 \bm{\mathcal{D}}_{\Phi, \tau } (\bm{Y})
 &=
\arg \min_{\bm{X}} \tau \cdot \| \bm{X}  \| _ {\Phi} + \frac{1}{2} \|\bm{X}-\bm{Y}\|_F^2.
\end{align*}


\section{\textbf{Related Applications}} \label{application}

In this section, we primarily investigate the large-scale  high-order  tensor recovery,
where the model formulation is shifted from unquantized observation to quantized observation.
The 
gradient-domain regularization method in Section \ref{model}  functions as a modeling capability enhancement module, whereas the randomized compression technique in Section \ref{fast-fctn-dec}  serves as a computational acceleration module for algorithm design.


\subsection{\textbf{Nonquantized High-Order  Tensor Recovery}}
\subsubsection{\textbf{Proposed Generalized Nonconvex  Model}} \label{Model_Formulation} 
\textcolor[rgb]{0.00,0.00,0.00}{On the basis of Subsection \ref{nonconvex-regulari}, this subsection proposes  a generalized  nonconvex robust tensor completion   model under the FCTN framework (\textit{abbreviated as FCTN-GNRTC}),  i.e.,}
\begin{align}
& \min_{{\boldsymbol{\mathcal{L}}},{\boldsymbol{\mathcal{E}}}}
\frac{1}{\gamma} \sum_{t \in   \Gamma} \Big \{
\sum_{k=1}^{\bar{N}} \alpha_k \cdot \big \| {\textbf{G}_{t}} _{[\textbf{n}^k_{\textbf{1}};\textbf{n}^k_{\textbf{2}}]} \big\|_\Phi \Big\}
 + {\lambda} \cdot  \psi \big(h( {\boldsymbol{\mathcal{E}}} )\big),
\notag \\ \label{orin_nonconvex}
&\text{s.t.}  \;\; {\boldsymbol{\mathcal{G}}}_{t}=\nabla_{t}({\boldsymbol{\mathcal{L}}}), \;\;
\boldsymbol{\bm{P}}_{{{\Omega}}}( {\boldsymbol{\mathcal{L}}} +{\boldsymbol{\mathcal{E}}})= \boldsymbol{\bm{P}}_{{{\Omega}}}({\boldsymbol{\mathcal{M}}}),
\end{align}
where ${\boldsymbol{\mathcal{L}}}$  represents the underlying tensor with both low-rank and smooth structures,
 ${\boldsymbol{\mathcal{E}}}$ denotes the noise/outlier tensor,
 ${\boldsymbol{\mathcal{M}}}$  signifies the observed tensor,
 $\Phi(\cdot)$ and  $\psi(\cdot)$  are the generalized nonconvex functions,
  and ${\lambda}$ is a trade-off parameter.
  The meanings of other symbols remain the same as those defined in Equations
  (\ref{unified-regulare22}) and (\ref{unified-regulare11}).
  Note that when  the noise/outlier tensor  
  ${\boldsymbol{\mathcal{E}}}$
  is set to be zero, the proposed FCTN-GNRTC model (\ref{orin_nonconvex})
   degenerates into the following \textit{FCTN-based Generalized  Nonconvex  Tensor Completion} (FCTN-GNTC) model:
   \begin{align}
& \min_{{\boldsymbol{\mathcal{L}}},{\boldsymbol{\mathcal{E}}}}
\frac{1}{\gamma} \sum_{t \in   \Gamma} \Big \{
\sum_{k=1}^{\bar{N}} \alpha_k \cdot \big \| {\textbf{G}_{t}} _{[\textbf{n}^k_{\textbf{1}};\textbf{n}^k_{\textbf{2}}]} \big\|_\Phi \Big\},
\notag \\ \label{orin_nonconvex2233333}
&\text{s.t.}  \;\; {\boldsymbol{\mathcal{G}}}_{t}=\nabla_{t}({\boldsymbol{\mathcal{L}}}), \;\;
\boldsymbol{\bm{P}}_{{{\Omega}}}( {\boldsymbol{\mathcal{L}}} )= \boldsymbol{\bm{P}}_{{{\Omega}}}({\boldsymbol{\mathcal{M}}}).
\end{align}

\subsubsection{\textbf{ADMM-Based Optimization Algorithm}}\label{fctn-gnrtc-admm}
Considering the limitation imposed by the paper's length, we have moved the entire derivation of the ADMM optimization algorithm, along with its associated time complexity analysis and convergence analysis, to the supplementary materials section.
\subsection{\textbf{Quantized High-Order Tensor Recovery}} 

 \subsubsection{\textbf{Generalized Nonconvex Quantized   Model}}
 In this subsection, we address the problem of robust  tensor completion from uniform scalar quantization using the proposed FCTN-GNTCTV regularization method.
 Let ${u}_k$ ($\forall k \in [m]$) be the $k$-th corrupted observations from the low-rank plus smooth tensor   ${\boldsymbol{\mathcal{L}}} $ and the noise/outlier tensor ${\boldsymbol{\mathcal{S}}} $  via $u_k= \langle{\boldsymbol{\mathcal{P}}}_{k},{\boldsymbol{\mathcal{L}}} + {\boldsymbol{\mathcal{E}}} \rangle + \epsilon _k$,  where ${\boldsymbol{\mathcal{P}}}_{k}$ is a sampler that uniformly and randomly selects one entry from corrupted tensor ${\boldsymbol{\mathcal{L}}} + {\boldsymbol{\mathcal{E}}}$,  $ \epsilon _k$ refers to the  additive  Gaussian noise.
 Motivated by the fact that the uniformly quantized dithered observations are equal to their original counterparts in terms of expectation,
 we propose the following quantized observation model:
 $$
{y}_k=\operatorname{Q}_{\delta} ( u_k  + \xi_k),
$$
in which $\operatorname{Q}_{\delta} (\cdot):=  \delta ( \lfloor \cdot / \delta \rfloor + \frac{1}{2}) $ represents a uniform scalar quantizer  with   resolution parameter $\delta >0$,   and dithers $\xi_k$ are generated from a uniform distribution $\operatorname{U} ([-\frac{\delta}{2}, \frac{\delta}{2}] )$.

Furthermore, to handle the incomplete   tensor damaged by both Gaussian noise and sparse noise simultaneously, we   propose the following  generalized nonconvex  robust  tensor completion  model in the  quantized  scenario. Specifically,
\begin{align}
& 
\min _    {   \|  \boldsymbol{\mathcal{L}} \|_{\infty} \leq \alpha }
\frac  {1} {2m}  \sum_{k=1} ^{m} \big(  \langle{\boldsymbol{\mathcal{P}}}_{k},{\boldsymbol{\mathcal{L}}} + {\boldsymbol{\mathcal{E}}}  \rangle -  {y}_k \big)^2
+ {\lambda_2} \cdot  \psi \big(h( {\boldsymbol{\mathcal{E}}} )\big)
\notag \\   & +
\frac{\lambda_1}{\gamma} \sum_{t \in   \Gamma} \Big \{ \sum_{k=1}^{\bar{N}} \alpha_k \cdot \big \| {\textbf{G}_{t}} _{[\textbf{n}^k_{\textbf{1}};\textbf{n}^k_{\textbf{2}}]} \big\|_\Phi \Big\},
 \notag \\   \label{1bittensor-gu-spEEEEEE}  &
 \text{s.t.} \;\;  
  {\boldsymbol{\mathcal{G}}}_{t}=\nabla_{t}({\boldsymbol{\mathcal{L}}}),
 t \in \Gamma,
\end{align}
where $\lambda_1 $ and $\lambda_2 $ are  the regularization parameter,   $y_k$ denotes the quantized measurements.
 The meanings of other symbols remain the same as those defined in Equations
  (\ref{unified-regulare22}) and (\ref{unified-regulare11}).
Note that when we disregard the sparse noise scenario (i.e., $ {\boldsymbol{\mathcal{E}}}$ is set to be zero),
the proposed model (\ref{1bittensor-gu-spEEEEEE})  reduces to the following model:
\begin{align}
& 
\min _    {   \|  \boldsymbol{\mathcal{L}} \|_{\infty} \leq \alpha }
\frac  {1} {2m}  \sum_{k=1} ^{m} \Big(  \langle{\boldsymbol{\mathcal{P}}}_{k},{\boldsymbol{\mathcal{L}}}   \rangle -  
\operatorname{Q}_{\delta} \big (
\langle{\boldsymbol{\mathcal{P}}}_{k},{\boldsymbol{\mathcal{L}}}  \rangle + \epsilon _k +
\xi_k
\big )
\Big)^2
%
\notag \\   &
+
\frac{\lambda_1}{\gamma} \sum_{t \in   \Gamma} \Big \{ \sum_{k=1}^{\bar{N}} \alpha_k \cdot \big \| {\textbf{G}_{t}} _{[\textbf{n}^k_{\textbf{1}};\textbf{n}^k_{\textbf{2}}]} \big\|_\Phi \Big\},
 \notag \\   \label{1bittensor-gu-spEEEEEE2222}  &
 \text{s.t.} \;\;  
  {\boldsymbol{\mathcal{G}}}_{t}=\nabla_{t}({\boldsymbol{\mathcal{L}}}),
 t \in \Gamma,
\end{align}
where the 
 symbols $\operatorname{Q}_{\delta} (\cdot) $, ${\boldsymbol{\mathcal{P}}}_{k}$,
$\epsilon _k $,  $\xi_k$, and $\lambda_1$  are  the same as those defined in model (\ref{1bittensor-gu-spEEEEEE}).
For brevity,   the quantized  tensor models (\ref{1bittensor-gu-spEEEEEE})  and  (\ref{1bittensor-gu-spEEEEEE2222})
 are  abbreviated as FCTN-GNQRTC and FCTN-GNQTC, respectively.
It is noteworthy that when the
quantizer $\operatorname{Q}_{\delta} (\cdot)= \operatorname{sign} (\cdot)$ is set to be sign function,
the aforementioned methods are  simplified to the versions corresponding to one-bit quantization scenario.

\subsubsection{\textbf{ADMM-Based Optimization Algorithm}} 
 In this subsection, the  ADMM framework  \cite{boyd2011distributed} is adopted to solve the proposed
   quantized  models (\ref{1bittensor-gu-spEEEEEE})  and  (\ref{1bittensor-gu-spEEEEEE2222}).
\textit{Due to space constraints of this paper,
we have placed the detailed ADMM optimization algorithms
for solving Model (\ref{1bittensor-gu-spEEEEEE})  and  Model (\ref{1bittensor-gu-spEEEEEE2222}),
along with their corresponding time complexity analysis and convergence analysis 
in the supplementary materials.}

\section{\textbf{EXPERIMENTAL RESULTS}}\label{experiments}

In this section, we  perform  extensive experiments    on
synthetic and real-world tensor data
to substantiate the superiority and effectiveness of the proposed tensor recovery 
approach.
\textcolor[rgb]{0.00,0.00,0.00}{All the experiments  are run  on the following
platforms:
\textbf{1)} 
Windows 11 and Matlab (R2020a) with an Intel(R) Core(TM) i9-14900KF  CPU and 64GB memory;
\textbf{2)}
 %
 Windows 10 and Matlab (R2022b) with an Intel(R) Xeon(R) Gold-6230 2.10GHz CPU and 128GB memory.
 }

\subsection
{\textcolor[rgb]{0.00,0.00,0.00}{\textbf{Experimental Settings}}} %
\subsubsection{\textbf{Experimental Datasets}}
In our experiments,
\textit{Color Images} (CIs),  \textit{Color Videos} (CVs), \textit{Multi-Temporal Remote Sensing Images} (MRSIs),
 \textit{Hyperspectral Videos} (HVs), \textit{Face Datasets}, and
 Cardiac \textit{Magnetic Resonance Images}(MRIs) are utilized as the tested tensor datasets.
 Due to the limitation of paper length, we present the detailed description of the experimental datasets in the supplementary materials.


\textbf{Type 1: Color Images (CIs):}
This dataset
includes four large-scale high-dimensional 
   CIs, which are called as
   Maastricht-Bassin ($2177  \times  3113 \times 3$), Wildpark ($2749  \times  4116 \times 3$), WestLotto ($2612  \times  4650 \times 3$), and Sunrise ($3040  \times  4056 \times 3$),
respectively.
We   download these large-scale  CIs from  the Google art project website \footnote{\url{{https://commons.wikimedia.org/wiki/Google_Art_Project}}}.
%

\textbf{Type 2: Color Videos (CVs):}
This dataset 
includes four four-order  CVs called Rush-hour, Johnny, Stockholm, Intotree, respectively.
We   download these large-scale CVs from  the derf website \footnote{\url{https://media.xiph.org/video/derf/}}.
Only the first $50$ frames of each video sequence are selected as the tested CVs owing  to the computational limitation,
in which  each frame has the  size $720 \times 1280 \times 3$.

\textbf{Type 3: Multi-Temporal Remote Sensing Images (MRSIs):}
This dataset
mainly involves  four 
fourth-order MRSIs,
which are named  SPOT-5 \footnote{\url{{https://take5.theia.cnes.fr/atdistrib/take5/client/\#/home}}} ($2000 \times 2000 \times   4 \times 13$),
Landsat-7
($4500 \times 4500 \times   6 \times 11$), T29RMM ($5001\times 5001 \times   4 \times 6$),
 and T22LGN \footnote{\url{{https://theia.cnes.fr/atdistrib/rocket/\#/home}}} ($5001\times 5001 \times   4 \times 7$), respectively.
To speed up the calculation process, the spatial size of  these MRSIs  is downsampled (resized) to
  $1000\times 1000$.

\textbf{Type 4: Hyperspectral Videos (HVs):}
This dataset includes 
three  HVs. 
The first HV,
 named Main \footnote{\url{{https://openremotesensing.net/knowledgebase/hyperspectral-video/}}},
   has dimensions of
  $480 \times  752 \times 33 \times 31$,
which contains $31$ frames, and
each frame has $33$ bands from $400$nm to $720$nm wavelength with a $10$nm step \cite{Mian:121}.
We downsample
each brand size to $240 \times 240 $ 
 and then reformat the $4$D HSV into a  $ 240 \times 240 \times 33 \times31$
smaller tensor
as a result of
computational limitation. %
The other two HVs \footnote{\url{{https://www.hsitracking.com/contest/}}},
named Oranges1 and Pool5 respectively, both have dimensions of $200 \times 200 \times 16 \times 200$.

  \textbf{Type 5: Face Datasets:}
This first face dataset (\textbf{Extended YaleFace Dataset B}
\footnote{\url{http://vision.ucsd.edu/~iskwak/ExtYaleDatabase/ExtYaleB.html}})  includes $38$ subjects with $9$ poses under $64$ illumination conditions  \cite{georghiades2001few,lee2005acquiring}.
Each face image has the size of $192 \times 168$. 
The image subsets of $38$ subjects under $64$ illumination with $1$ pose are considered as  the   testing tensor data
by formatting the data into a fourth-order tensor
with the size of  ${192 \times 168 \times 64 \times 38}$.

The second face dataset (\textbf{UWA Hyperspectral Face Database} \footnote{\url{https://openremotesensing.net/kb/data/ }})
contains hyperspectral image cubes of $78$ subjects imaged in multiple sessions.
Each cube includes $33$ spectral bands covering the spectral range
of $400 $-$ 720$nm with a $10$nm step \cite{uzair2015hyperspectral}. 
Each image is cropped 
(resized) to the size of $256 \times 256$.
We only select the faces of $30$ target objects photographed in $33$ bands as our test data,
which can be represented by a fourth-order tensor with a size of $256 \times  256 \times  30 \times  33$.

\textbf{Type 6: Cardiac \textit{Magnetic Resonance Images}
(MRIs): 
}
These  datasets are acquired from 33 subjects, each of which
consists of 20 frames and $8$-$15$ slices along the long axis \cite{andreopoulos2008efficient}.
%
%
We choose four subjects as the testing cardiac MRIs (i.e., sol-yxzt-part 1/2/8/30
\footnote{\url{{https://jtl.lassonde.yorku.ca/software/datasets/}}}).
Each selected subject 
is a $256 \times 256 \times 14 \times 20$ or $256 \times 256 \times 15 \times 20$
tensor whose elements are   magnetic resonance imaging  measurements
indexed by $(x,y,z,t)$, where  $(x,y,z)$ is a point in space and $t$ corresponds to time.

In our experiments, each  raw tested tensor 
is  conducted with
 normalization operation.
%

\subsubsection{\textbf{Evaluation Metrics}}
We employ the mean of 
peak signal-to-noise ratio 
(MPSNR), the mean of relative squared error (MRSE),
the mean of structural similarity (MSSIM),
 and the mean of CPU time (MTime)
 as 
 the quantitative evaluation metric.
Meanwhile, the best
and the second-best results are highlighted 
in boldface and blue,
respectively.
Generally speaking,  higher MPSNR/MSSIM values and lower MTime/MRSE indicate better restoration performance.

\subsubsection{\textbf{Relevant Configurations}}
In this experiment, we defined the
\textit{sampling ratio} (SR) as $SR=\frac{|\Omega|}{I_1 I_2 \cdots I_N}$
for an $I_1 \times I_2 \times  \cdots \times I_N$ tensor, where the observed index set
$\Omega$ is generated uniformly at random and
$|\Omega|$ represents the cardinality of $\Omega$.
For robust tensor completion task,
the observed tensor is constructed as follows:
firstly, sparse pepper-and-salt impulsive noise 
is added to the underlying tensor with corruption rate $NR$
\footnote{The Matlab function ``$\operatorname{imnoise}$  (I, `salt $\&$ pepper', $NR$)" is used for
noise generation, where $NR$ denotes the normalized noise intensity corresponding to SNR = $\frac{1}{NR}$ dB.},
and then
incorporate
mean-zeros Gaussian noise with a standard deviation of $\sigma$.
Finally, the  observed observations are formed by the given  \textit{sampling ratio} SR.
The parameters in compared methods are manually adjusted to the optimal
performance, which refers to the discussion in their corresponding articles.
The parameter settings of the proposed method are detailed in the supplementary material.

\begin{table*} 
\renewcommand{\arraystretch}{0.85}
\setlength\tabcolsep{1.7pt}
  \centering
  \caption{
  Quantitative evaluation PSNR, SSIM, RSE and  CPU Time (Second) of 
  various LRTC  methods  on face datasets.}
  \vspace{-0.15cm}
  \label{table-facedata1}
  \scriptsize
  \begin{threeparttable}
     \begin{tabular}{cccc cccc cccc  c ccc c}
    \Xhline{1pt}
   \tabincell{c}   {SR}&
   \tabincell{c} {Evaluation\\ Metric}&

\tabincell{c}{R1-FCTN \\ -GNTC} &
 \tabincell{c}{R2-FCTN \\ -GNTC} & \tabincell{c}{FCTNFR \\ \cite{zheng2022tensor}}& \tabincell{c}{ FCTN-TC\\  \cite{zheng2021fully}}  &
\tabincell{c}{ FCTN \\ -NNM \cite{liu2024fully} }& %
\tabincell{c}{ TRNN \\  \cite{yu2019tensor} }& %
\tabincell{c}{TCTV-TC  \\ \cite{wang2023guaranteed}}&
\tabincell{c}{MTTD   \\ \cite{feng2023multiplex} }& %
\tabincell{c}{EMLCP \\ -LRTC\cite{zhang2023tensor}}&
 \tabincell{c}{WSTNN \\ \cite{zheng2020tensor44} }&
 \tabincell{c}{METNN \\  \cite{liu2024revisiting} }& %
 \tabincell{c} {HTNN \\ \cite{qin2022low} }& %
  \tabincell{c} {OTNN \\ \cite{qiuyn2025}}& %
 \tabincell{c} {GTNN \\ -HOC  \cite{wang2024low2222}}&
 \tabincell{c} {t-$\epsilon$-LogDet \\ \cite{ yang2022355}}



    \cr

    \Xhline{1pt}
\hline    \hline
   \multirow{4}{*}{    \tabincell{c}   {$0.1\%$ }  } & 
   MPSNR& \textcolor[rgb]{1.00,0.00,0.00}{\textbf{24.26}} & \textcolor[rgb]{0.00,0.00,1.00}{\textbf{24.24}}&14.58&15.04&18.17&14.33&22.48&16.62&20.24&18.61&14.00&17.92&18.51&18.46&14.18\cr

   \qquad	&MSSIM	& \textcolor[rgb]{0.00,0.00,1.00}{\textbf{0.725}}   & \textcolor[rgb]{1.00,0.00,0.00}{\textbf{0.732}}&0.180&0.188&0.612&0.495&0.646&0.592&0.531&0.639&0.440&0.476&0.609&0.355&0.184\cr

   \qquad	&MRSE	& \textcolor[rgb]{1.00,0.00,0.00}{\textbf{0.257}}  & \textcolor[rgb]{1.00,0.00,0.00}{\textbf{0.257}}&0.700&0.668&0.523&0.746&\textcolor[rgb]{0.00,0.00,1.00}{\textbf{0.315}}&0.583&0.427&0.495&0.763&0.509&0.482&0.472&0.741\cr
    \qquad	&MTime	&4578  &  4872&11244&5009&34357&32969&10198&13091&17061&16782&8484&\textcolor[rgb]{1.00,0.00,0.00}{\textbf{3489}}&6905&\textcolor[rgb]{0.00,0.00,1.00}{\textbf{4523}}&4796  \cr

   \hline
   \multirow{4}{*}{    \tabincell{c}   {$0.3\%$ }  } &MPSNR  
  &   \textcolor[rgb]{0.00,0.00,1.00}{\textbf{26.95}}   & \textcolor[rgb]{1.00,0.00,0.00}{\textbf{26.96}}&19.36&20.28&22.64&20.26&25.21&22.92&23.01&24.21&17.57&22.59&19.96&21.42&16.31\cr

   \qquad	&MSSIM	 &\textcolor[rgb]{0.00,0.00,1.00}{\textbf{0.768}}  & \textcolor[rgb]{1.00,0.00,0.00}{\textbf{0.773}}&0.367&0.368&0.708&0.667&0.693&0.710&0.500&0.750&0.616&0.599&0.659&0.427&0.244\cr
   \qquad	&MRSE&   \textcolor[rgb]{0.00,0.00,1.00}{\textbf{0.199}} &\textcolor[rgb]{1.00,0.00,0.00}{\textbf{0.198}}&0.429&0.388&0.310&0.403&0.239&0.292&0.311&0.274&0.540&0.294&0.409&0.321&0.586\cr
    \qquad	&MTime	& \textcolor[rgb]{0.00,0.00,1.00}{\textbf{4491}}  & 5227&11293&7773&35327&35131&9987&13096&16982&16699&8464&\textcolor[rgb]{1.00,0.00,0.00}{\textbf{3483}}&6969&4550&4795 \cr

   \hline
  \multirow{4}{*}{    \tabincell{c}   {$0.5\%$ }  } &MPSNR  
 & \textcolor[rgb]{0.00,0.00,1.00}{\textbf{28.51}} &\textcolor[rgb]{1.00,0.00,0.00}{\textbf{28.58}}&23.94&23.98&24.82&22.44&26.76&25.28&23.66&26.87&20.76&24.70&21.63&22.75&18.00\cr

   \qquad	&MSSIM	&  \textcolor[rgb]{0.00,0.00,0.00}{0.788} & \textcolor[rgb]{0.00,0.00,1.00}{\textbf{0.795}}&0.567&0.527&0.751&0.718&0.713&0.758&0.509&\textcolor[rgb]{1.00,0.00,0.00}{\textbf{0.797}}&0.682&0.647&0.700&0.451&0.282\cr
   \qquad	&MRSE	 &  \textcolor[rgb]{0.00,0.00,1.00}{\textbf{0.172}} & \textcolor[rgb]{1.00,0.00,0.00}{\textbf{0.170}}&0.263&0.264&0.247&0.313&0.205&0.229&0.303&0.208&0.406&0.244&0.342&0.277&0.485\cr
    \qquad	&MTime	&  5295   & 5718&11373&9463&34380&35072&10051&13069&16947&16666&8471&\textcolor[rgb]{1.00,0.00,0.00}{\textbf{3491}}&7076&\textcolor[rgb]{0.00,0.00,1.00}{\textbf{4560}}&4809 \cr

\hline
  \multirow{4}{*}{    \tabincell{c}   {$1\%$ }  } &MPSNR 
 & \textcolor[rgb]{0.00,0.00,1.00}{\textbf{31.56}} &\textcolor[rgb]{1.00,0.00,0.00}{\textbf{31.60}}&30.92&28.84&28.21&25.00&29.29&28.45&28.38&29.56&26.49&26.72&24.08&24.63&21.54 \cr
   \qquad	&MSSIM	&  \textcolor[rgb]{0.00,0.00,1.00}{\textbf{{0.836}}}  & \textcolor[rgb]{1.00,0.00,0.00}{\textbf{0.840}}&0.788&0.696&0.811&0.767&0.755&0.814&0.653&\textcolor[rgb]{0.00,0.00,1.00}{\textbf{0.836}}&0.779&0.694&0.751&0.519&0.402 \cr
   \qquad	&MRSE	 &  \textcolor[rgb]{1.00,0.00,0.00}{\textbf{0.123}} &\textcolor[rgb]{1.00,0.00,0.00}{\textbf{0.123}}&\textcolor[rgb]{1.00,0.00,0.00}{\textbf{0.123}}&\textcolor[rgb]{0.00,0.00,1.00}{\textbf{0.157}}&0.176&0.239&0.158&0.171&0.200&\textcolor[rgb]{0.00,0.00,1.00}{\textbf{0.157}}&0.232&0.202&0.263&0.232&0.334 \cr
    \qquad	&MTime	& 6114  & 6526&11514&11655&34545&33124&10016&12930&17020&16857&8510&\textcolor[rgb]{1.00,0.00,0.00}{\textbf{3493}}&7253&\textcolor[rgb]{0.00,0.00,1.00}{\textbf{4589}}&4827 \cr

 \hline
  \multirow{4}{*}{    \tabincell{c}   {$3\%$ }  } &MPSNR
 &\textcolor[rgb]{1.00,0.00,0.00}{\textbf{36.49}} & \textcolor[rgb]{0.00,0.00,1.00}{\textbf{36.47}}&33.72&33.61&34.04&29.99&33.66&32.57&33.68&34.05&34.90&29.82&28.39&29.33&28.01 \cr
   \qquad	&MSSIM	&  \textcolor[rgb]{0.00,0.00,1.00}{\textbf{0.915}} & \textcolor[rgb]{1.00,0.00,0.00}{\textbf{0.916}}&0.855&0.836&0.889&0.842&0.825&0.871&0.783&0.889&0.900&0.752&0.828&0.660&0.611 \cr
   \qquad	&MRSE	 &  \textcolor[rgb]{1.00,0.00,0.00}{\textbf{0.066}} & \textcolor[rgb]{1.00,0.00,0.00}{\textbf{0.066}}&\textcolor[rgb]{0.00,0.00,1.00}{\textbf{0.090}}&\textcolor[rgb]{0.00,0.00,1.00}{\textbf{0.090}}&0.099&0.143&0.103&0.113&0.100&0.099&0.091&0.150&0.168&0.152&0.182 \cr
    \qquad	&MTime	&  8086 & 8144&8293&13512&34371&31004&10089&11664&17059&16859&8556&\textcolor[rgb]{1.00,0.00,0.00}{\textbf{3502}}&7396&\textcolor[rgb]{0.00,0.00,1.00}{\textbf{4638}}&4847  \cr

   \hline
    \hline

      \Xhline{1pt}
    \end{tabular}
    \end{threeparttable}
\end{table*}

\begin{figure*}[!htbp]
\renewcommand{\arraystretch}{0.5}
\setlength\tabcolsep{0.43pt}
\centering
\begin{tabular}{ccc  ccc cc }
\\
 \toprule

(a)   & (b)  & (c) & (d) & (e) &(f)& (g) & (h)

\\
\includegraphics[ width=0.875in, height=1.28in]{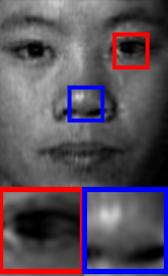} %
&
\includegraphics[ width=0.875in, height=1.28in]{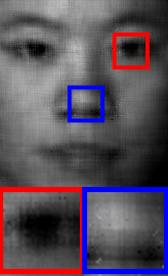}
&
\includegraphics[ width=0.875in, height=1.28in]{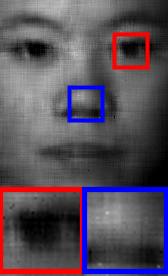}
&
\includegraphics[ width=0.875in, height=1.28in]{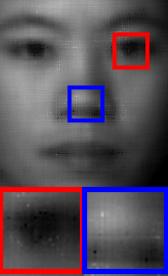}
&
\includegraphics[ width=0.875in, height=1.28in]{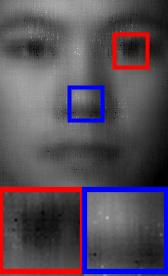}
&
\includegraphics[ width=0.875in, height=1.28in]{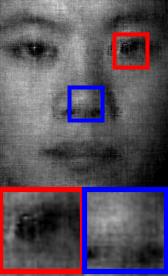}
&
\includegraphics[ width=0.875in, height=1.28in]{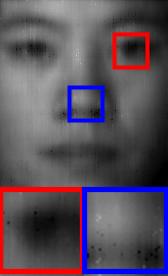}
&
\includegraphics[ width=0.875in, height=1.28in]{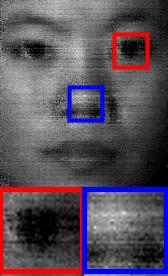}
\\

 (i) &  (j) & (k) &(l)&(m) & (n) & (o) & (p)\\

\includegraphics[ width=0.875in, height=1.28in]{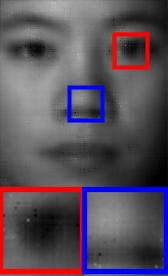}
&
\includegraphics[ width=0.875in, height=1.28in]{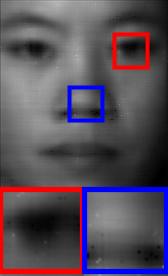}
&
\includegraphics[ width=0.875in, height=1.28in]{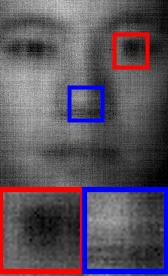}
&
\includegraphics[ width=0.875in, height=1.28in]{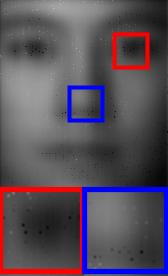}
&
\includegraphics[ width=0.875in, height=1.28in]{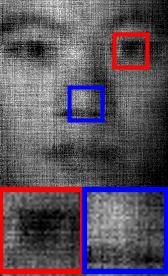}
&
\includegraphics[ width=0.875in, height=1.28in]{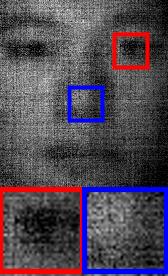}
&
\includegraphics[ width=0.875in, height=1.28in]{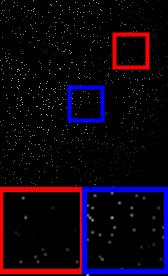}
&
\includegraphics[ width=0.875in, height=1.28in]{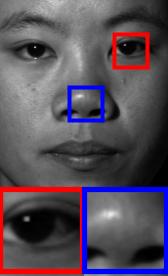}
\\
 \toprule
\end{tabular}
\caption{
Visual comparison of various LRTC methods for  Face datasets 
inpainting under
 $SR=1\%$.
From left to right:
(a) Ours,   (b) FCTNFR, (c) FCTNTC,  (d)  FCTN-NNM,  (e) TRNNM, (f) TCTV, (g) MTTD, (h) EMLCP,
(i) WSTNN,  (j) METNN, (k)  HTNN, (l) OTNN, (m) GTNN-HOC,  (n) t-$\epsilon$-LogDet, (o) Observed, (p) Ground-truth.}
\label{fig_facehv} 
\end{figure*}

\begin{figure*}[!htbp]
\renewcommand{\arraystretch}{0.5}
\setlength\tabcolsep{0.43pt}
\centering
\begin{tabular}{ccc  ccc cc }
\\
 \toprule

(a)   & (b)  & (c) & (d) & (e) &(f)& (g) & (h)
\\
\includegraphics[ width=0.875in, height=1.28in]{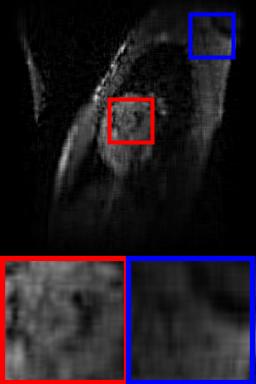} 
&
\includegraphics[ width=0.875in, height=1.28in]{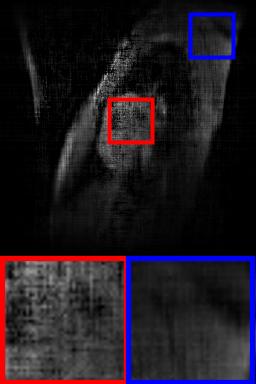}
&
\includegraphics[ width=0.875in, height=1.28in]{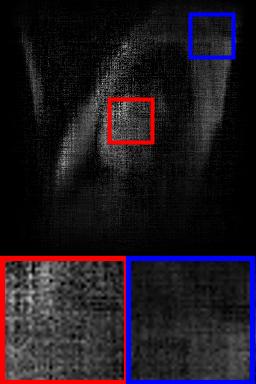}
&
\includegraphics[ width=0.875in, height=1.28in]{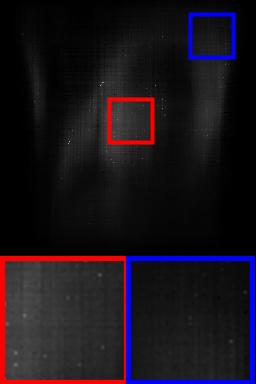}
&
\includegraphics[ width=0.875in, height=1.28in]{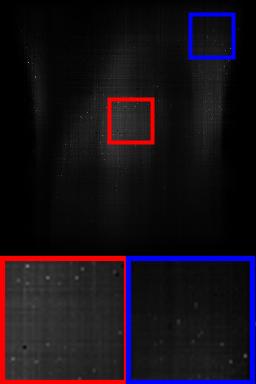}
&
\includegraphics[ width=0.875in, height=1.28in]{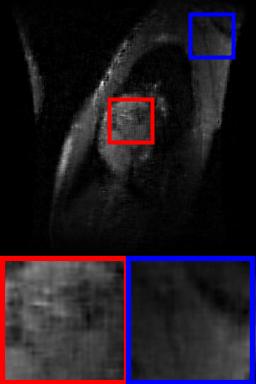}
&
\includegraphics[ width=0.875in, height=1.28in]{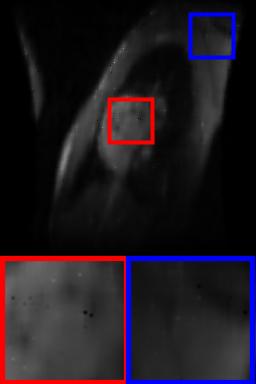}
&
\includegraphics[ width=0.875in, height=1.28in]{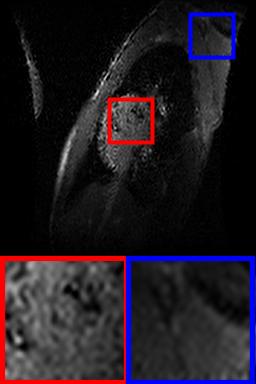} 
\\

 (i) &  (j) & (k) &(l)&(m) & (n) & (o) & (p)\\ %

\includegraphics[ width=0.875in, height=1.28in]{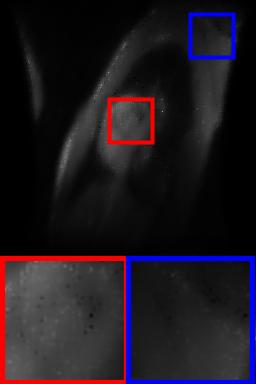}
&
\includegraphics[ width=0.875in, height=1.28in]{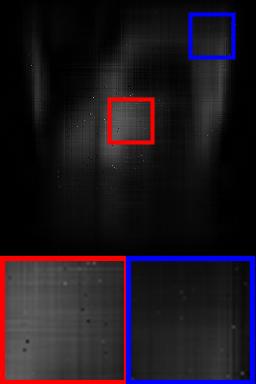}
&
\includegraphics[ width=0.875in, height=1.28in]{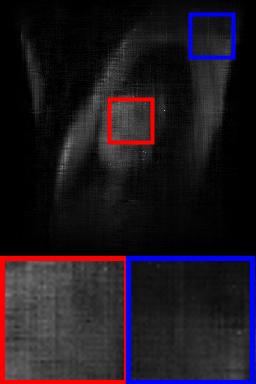}
&
\includegraphics[ width=0.875in, height=1.28in]{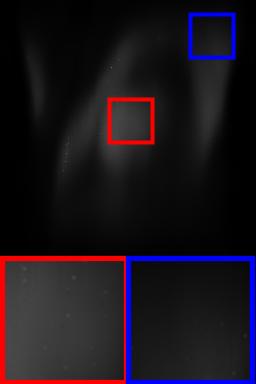}
&
\includegraphics[ width=0.875in, height=1.28in]{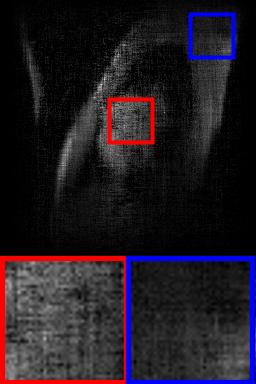}
&
\includegraphics[ width=0.875in, height=1.28in]{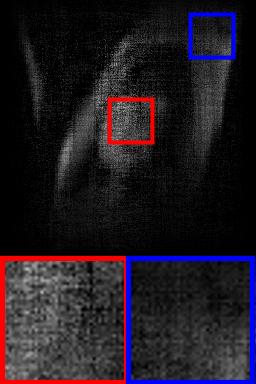}
&
\includegraphics[ width=0.875in, height=1.28in]{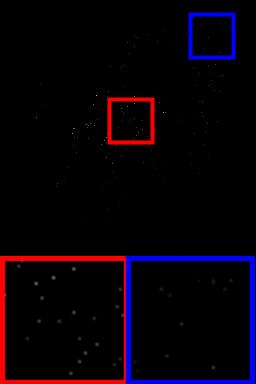}
&
\includegraphics[ width=0.875in, height=1.28in]{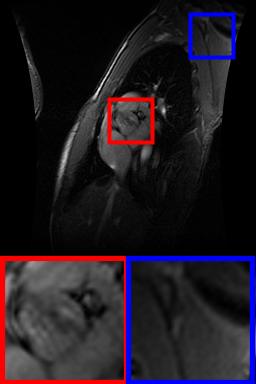}
\\

 \toprule
\end{tabular}
\caption{
Visual comparison of various LRTC methods for  MRI datasets 
inpainting under $SR=1\%$.
From left to right:
(a) Ours,   (b) FCTNFR, (c) FCTNTC,  (d)  FCTN-NNM,  (e) TRNNM, (f) TCTV, (g) MTTD, (h) EMLCP,
(i) WSTNN,  (j) METNN, (k)  HTNN, (l) OTNN, (m) GTNN-HOC,  (n) t-$\epsilon$-LogDet, (o) Observed, (p) Ground-truth.}
\label{fig_mri}
\end{figure*}

\begin{table*} 
\renewcommand{\arraystretch}{0.8}
\setlength\tabcolsep{1.7pt}
  \centering
  \caption{
  Quantitative evaluation PSNR, SSIM, RSE and  CPU Time (Second) of 
  various LRTC  methods  on MRIs.}
  \vspace{-0.15cm}
  \label{table-MSIMSI}
  \scriptsize
  \begin{threeparttable}
     \begin{tabular}{cccc cccc cccc  c ccc c}
    \Xhline{1pt}
   \tabincell{c}   {SR}&
   \tabincell{c} {Evaluation\\ Metric}&

\tabincell{c}{R1-FCTN \\ -GNTC} &
 \tabincell{c}{R2-FCTN \\ -GNTC} & \tabincell{c}{FCTNFR \\ \cite{zheng2022tensor}}& \tabincell{c}{ FCTN-TC\\  \cite{zheng2021fully}}  &
\tabincell{c}{ FCTN \\ -NNM \cite{liu2024fully} }& %
\tabincell{c}{ TRNN \\  \cite{yu2019tensor} }& %
\tabincell{c}{TCTV-TC  \\ \cite{wang2023guaranteed}}&
\tabincell{c}{MTTD   \\ \cite{feng2023multiplex} }& %
\tabincell{c}{EMLCP \\ -LRTC\cite{zhang2023tensor}}&
 \tabincell{c}{WSTNN \\ \cite{zheng2020tensor44} }&
 \tabincell{c}{METNN \\  \cite{liu2024revisiting} }& %
 \tabincell{c} {HTNN \\ \cite{qin2022low} }& %
  \tabincell{c} {OTNN \\ \cite{qiuyn2025}}& %
 \tabincell{c} {GTNN \\ -HOC  \cite{wang2024low2222}}&
 \tabincell{c} {t-$\epsilon$-LogDet \\ \cite{ yang2022355}}


    \cr

    \Xhline{1pt}
\hline    \hline
   \multirow{4}{*}{    \tabincell{c}   {$0.1\%$ }  } &MPSNR&
   \textcolor[rgb]{0,0,1}{\textbf{26.22}}  &
 \textcolor[rgb]{1,0,0}{\textbf{27.38}}&21.14&20.48&19.62&19.19&24.72&21.47&26.21&21.11&19.18&20.89&19.18&21.10&21.31
 \cr

   \qquad	&MSSIM	&
     0.728    &
   \textcolor[rgb]{1,0,0}{\textbf{0.780}}&0.552&0.563&0.577&0.380&\textcolor[rgb]{0,0,1}{\textbf{0.751}}&0.678&0.742&0.652&0.335&0.654&0.378&0.469&0.483 \cr

   \qquad	&MRSE	&
   \textcolor[rgb]{0,0,1}{\textbf{0.452}} &
   \textcolor[rgb]{1,0,0}{\textbf{0.398}}&0.796&0.861&0.952&0.999&0.535&0.776&\textcolor[rgb]{0,0,1}{\textbf{0.452}}&0.807&1.000&0.827&0.999&0.791&0.779 \cr
    \qquad	&MTime	&
      2008&
     2264&7238&3974&7087&7974&3948 
    &3987&6150&6462& 3198 
    &\textcolor[rgb]{1,0,0}{\textbf{910}}&3671&\textcolor[rgb]{0,0,1}{\textbf{1640}}&1737\cr

   \hline
   \multirow{4}{*}{    \tabincell{c}   {$0.3\%$ }  } &MPSNR
   & \textcolor[rgb]{0,0,1}{\textbf{28.92}}
  &  \textcolor[rgb]{1,0,0}{\textbf{29.91}}&23.29&21.89&21.81&20.36&27.59&25.79&28.85&23.77&20.85&24.47&22.03&23.88&23.23\cr

   \qquad	&MSSIM	 & 0.777   &
   \textcolor[rgb]{1,0,0}{\textbf{0.829}}&0.614&0.574&0.712&0.663&0.789&0.753&\textcolor[rgb]{0,0,1}{\textbf{0.790}}&0.732&0.491&0.685&0.675&0.523&0.515\cr
   \qquad	&MRSE& \textcolor[rgb]{0,0,1}{\textbf{0.333}}   &
   \textcolor[rgb]{1,0,0}{\textbf{0.299}}&0.621&0.732&0.742&0.877&0.388&0.473&0.337&0.599&0.841&0.547&0.730&0.580&0.621	 \cr
    \qquad	&MTime	& 2039&
    2166&7442&4204&6577&8091&3865&4016&6132&6501&3190 
    &\textcolor[rgb]{1,0,0}{\textbf{945}}&3798&\textcolor[rgb]{0,0,1}{\textbf{1675}}&1765 \cr

   \hline
  \multirow{4}{*}{    \tabincell{c}   {$0.5\%$ }  } &MPSNR
 &  \textcolor[rgb]{0,0,0}{\textit{30.12}}   & \textcolor[rgb]{1,0,0}{\textbf{31.20}}&25.12&23.21&22.88&21.64&28.83&28.22&\textcolor[rgb]{0,0,1}{\textbf{30.15}}&25.67&23.09&26.17&23.32&24.90&24.24 \cr

   \qquad	&MSSIM	& 0.802 & \textcolor[rgb]{1,0,0}{\textbf{0.857}}&0.665&0.585&0.736&0.716&0.811& \textcolor[rgb]{0,0,1}{\textbf{0.819}}&0.816&0.772&0.682&0.717&0.713&0.556&0.536 \cr
   \qquad	&MRSE	 & \textcolor[rgb]{0,0,1}{\textbf{0.291}} &  \textcolor[rgb]{1,0,0}{\textbf{0.261}}&0.502&0.632&0.655&0.757&0.338&0.361&\textcolor[rgb]{0,0,1}{\textbf{0.291}}&0.481&0.637&0.452&0.629&0.516&0.559 \cr
    \qquad	&MTime	& 2227 & 2408&7415&4352&6538&8128&3873&4025&6080&6670&3577 
    &\textcolor[rgb]{1,0,0}{\textbf{904}}&3832&\textcolor[rgb]{0,0,1}{\textbf{1686}}&1760\cr

\hline
  \multirow{4}{*}{    \tabincell{c}   {$1\%$ }  } &MPSNR& 31.87
 & \textcolor[rgb]{1,0,0}{\textbf{32.06}}&28.75&25.51&25.21&23.66&30.72&30.66&\textcolor[rgb]{0,0,1}{\textbf{31.94}}&29.24&25.77&28.09&25.50&27.35&26.38 \cr
   \qquad	&MSSIM	&0.834 & \textcolor[rgb]{0,0,1}{\textbf{0.858}}&0.759&0.638&0.787&0.764&0.843&\textcolor[rgb]{1,0,0}{\textbf{0.867}}&0.847&0.838&0.745&0.773&0.766&0.640&0.617 \cr
   \qquad	&MRSE	 &  \textcolor[rgb]{0,0,1}{\textbf{0.241}} &\textcolor[rgb]{1,0,0}{\textbf{0.239}}&0.335&0.486&0.502&0.601&0.272&0.276&\textcolor[rgb]{1,0,0}{\textbf{0.239}}&0.322&0.472&0.366&0.491&0.391&0.439 \cr
    \qquad	&MTime	&  2370 &2730&7498&4574&7544&8177&4080 
    &4026&6027&8224&3500 
    &\textcolor[rgb]{1,0,0}{\textbf{948}}&3875&\textcolor[rgb]{0,0,1}{\textbf{1703}}&1769\cr

   \hline
    \hline

      \Xhline{1pt}
    \end{tabular}
    \end{threeparttable}
\end{table*}


\begin{table*} [!htbp]
\renewcommand{\arraystretch}{0.85}
\setlength\tabcolsep{1.7pt}
  \centering
  \caption{
  Quantitative evaluation PSNR, SSIM, RSE and  CPU Time (Second) of 
  various LRTC  methods  on MRSIs.}
  \vspace{-0.15cm}
  \label{table-MRSIMRSI1}
  \scriptsize
  \begin{threeparttable}
     \begin{tabular}{cccc cccc cccc  ccc c c}
    \Xhline{1pt}
   \tabincell{c}   {SR}&
   \tabincell{c} {Evaluation\\ Metric}&
 \tabincell{c}{R1-FCTN \\ -GNTC} &
 \tabincell{c}{R2-FCTN \\ -GNTC} & \tabincell{c}{FCTNFR \\ \cite{zheng2022tensor}}& \tabincell{c}{ FCTN-TC\\  \cite{zheng2021fully}}  &
\tabincell{c}{ FCTN \\ -NNM \cite{liu2024fully} }& %
\tabincell{c}{ TRNN \\  \cite{yu2019tensor} }& %
\tabincell{c}{TCTV-TC  \\ \cite{wang2023guaranteed}}&
\tabincell{c}{MTTD   \\ \cite{feng2023multiplex} }& %
\tabincell{c}{EMLCP \\ -LRTC\cite{zhang2023tensor}}&
 \tabincell{c}{WSTNN \\ \cite{zheng2020tensor44} }&
 \tabincell{c}{METNN \\  \cite{liu2024revisiting} }& %
 \tabincell{c} {HTNN \\ \cite{qin2022low} }& %
  \tabincell{c} {OTNN \\ \cite{qiuyn2025}}& %
 \tabincell{c} {GTNN \\ -HOC  \cite{wang2024low2222}}&
 \tabincell{c} {t-$\epsilon$-LogDet \\ \cite{ yang2022355}}


    \cr

    \Xhline{1pt}
\hline    \hline
   \multirow{4}{*}{    \tabincell{c}   {$0.5\%$ }  } &
   MPSNR& \textcolor[rgb]{1,0,0}{\textbf{21.88}}   & \textcolor[rgb]{0,0,1}{\textbf{21.84}}&17.43&17.04&20.63&17.25&21.53&20.95&18.98&20.82&15.77&19.83&19.00&18.32&16.32\cr

   \qquad	&MSSIM	& \textcolor[rgb]{1,0,0}{\textbf{0.537}}   &\textcolor[rgb]{1,0,0}{\textbf{0.537}}&0.366&0.368&0.433&0.391&\textcolor[rgb]{0,0,1}{\textbf{0.536}}&0.466&0.477&0.470&0.356&0.413&0.409&0.364&0.325\cr

   \qquad	&MRSE	& \textcolor[rgb]{1,0,0}{\textbf{0.301}}   &\textcolor[rgb]{0,0,1}{\textbf{0.302}}&0.533&0.556&0.345&0.551&0.316&0.327&0.397&0.349&0.640&0.364&0.462&0.408&0.545  \cr
    \qquad	&MTime	&  \textcolor[rgb]{1,0,0}{\textbf{2611}} 
    & 3026&14543&10043&11986&13633&9197&9106&19071&20064&9833&\textcolor[rgb]{0,0,1}{\textbf{2852}}&5751&3184&3343  \cr

   \hline
   \multirow{4}{*}{    \tabincell{c}   {$1\%$ }  } &MPSNR
  &  \textcolor[rgb]{1,0,0}{\textbf{23.91}}   &\textcolor[rgb]{0,0,1}{\textbf{23.70}}&20.93&19.37&21.54&19.99&23.28&23.02&19.36&22.42&18.41&21.25&20.60&19.75&18.30 \cr

   \qquad	&MSSIM	 & \textcolor[rgb]{1,0,0}{\textbf{0.658}}     &\textcolor[rgb]{0,0,1}{\textbf{0.650}}&0.522&0.462&0.472&0.444&0.639&0.580&0.536&0.563&0.429&0.497&0.462&0.468&0.425\cr
   \qquad	&MRSE&\textcolor[rgb]{1,0,0}{\textbf{0.234}}   &\textcolor[rgb]{0,0,1}{\textbf{0.242}}&0.350&0.416&0.308&0.385&0.256&0.262&0.376&0.287&0.471&0.316&0.367&0.351&0.431\cr
    \qquad	&MTime	& \textcolor[rgb]{0,0,1}{\textbf{2788}} 
     &3182&15259&12002&12145&13536&9232&8855&19136&20130&9862&\textcolor[rgb]{1,0,0}{\textbf{2739}}&5793&3194&3195  \cr

\hline
   \multirow{4}{*}{    \tabincell{c}   {$3\%$ }  } &MPSNR &
 {26.26}    & \textcolor[rgb]{0,0,1}{\textbf{26.27}}&25.88&23.72&23.64&22.35& \textcolor[rgb]{1,0,0}{\textbf{26.32}}&25.52&23.60&24.89&24.03&23.17&23.51&22.11&21.17 \cr
   \qquad	&MSSIM	&  \textcolor[rgb]{0,0,1}{\textbf{0.791}}   &
  {0.788}&0.770&0.718&0.619&0.573&\textcolor[rgb]{1,0,0}{\textbf{0.794}}&0.748&0.745&0.716&0.616&0.647&0.646&0.650&0.608 \cr
\qquad	&MRSE&  \textcolor[rgb]{0,0,1}{\textbf{0.185}}    &
\textcolor[rgb]{0,0,1}{\textbf{0.185}}&0.194&0.248&0.243&0.283&\textcolor[rgb]{1,0,0}{\textbf{0.183}}&0.202&0.240&0.202&0.229&0.254&0.259&0.275&0.308 \cr
 \qquad	&MTime&
 3659 & 
4047
&22794&18985&12600&16285&10099&8342&17810&18757&12831&\textcolor[rgb]{1,0,0}{\textbf{2584}}&6924&3466&\textcolor[rgb]{0,0,1}{\textbf{3433}} \cr

 \hline
   \multirow{4}{*}{    \tabincell{c}   {$5\%$ }  } &
   MPSNR& \textcolor[rgb]{0,0,1}{\textbf{27.09}}    &
   27.06&26.79&25.79&25.02&23.66& \textcolor[rgb]{1,0,0}{\textbf{27.87}}&26.85&25.01&26.34&25.49&24.44&24.98&23.75&22.92 \cr
   \qquad	&MSSIM	& \textcolor[rgb]{0,0,1}{\textbf{0.831}}    &
   0.827&0.807&0.786&0.722&0.673& \textcolor[rgb]{1,0,0}{\textbf{0.850}}&0.814&0.795&0.788&0.726&0.727&0.736&0.730&0.706 \cr
\qquad	&MRSE& \textcolor[rgb]{0,0,1}{\textbf{0.171}} &
0.172&0.174&0.199&0.209&0.243& \textcolor[rgb]{1,0,0}{\textbf{0.157}}&0.177&0.210&0.177&0.196&0.222&0.220&0.236&0.259 \cr
 \qquad	&MTime& 4195 
  &4637&23355&19016&12603&16316&11639&7416&17985&15626&12828& \textcolor[rgb]{1,0,0}{\textbf{2463}}&6501&3504& \textcolor[rgb]{0,0,1}{\textbf{3380}}  \cr

   \hline
    \hline

      \Xhline{1pt}
    \end{tabular}
    \end{threeparttable}
\end{table*}

%
\begin{figure*}[!htbp]
\renewcommand{\arraystretch}{0.0}
\setlength\tabcolsep{1pt}
\centering
\begin{tabular}{c c  cc}
\centering
\includegraphics[width=1.7in, height=1.46514in]{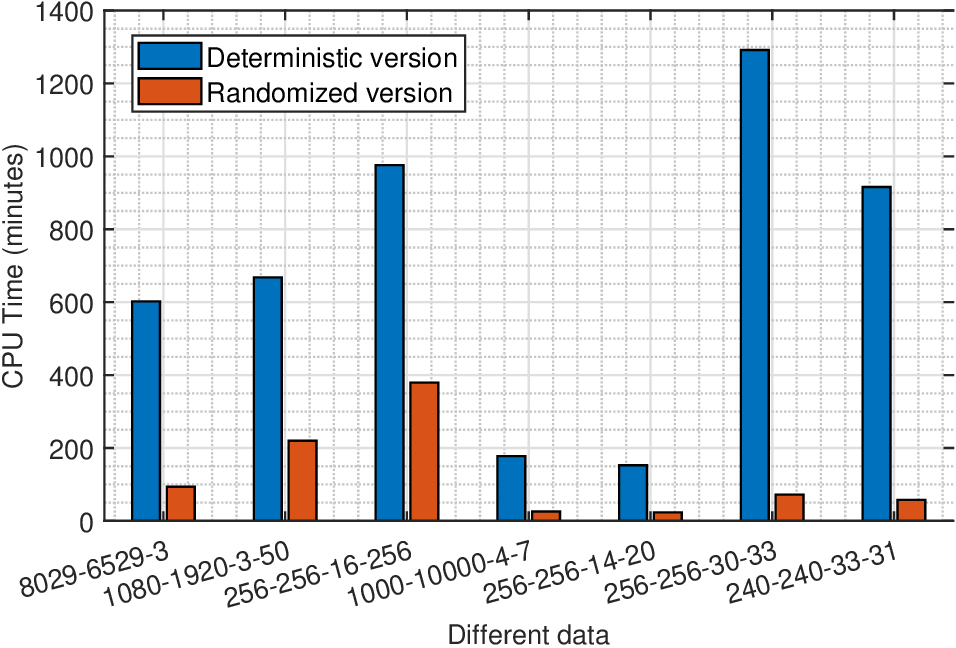}&
\includegraphics[width=1.7in, height=1.46514in]{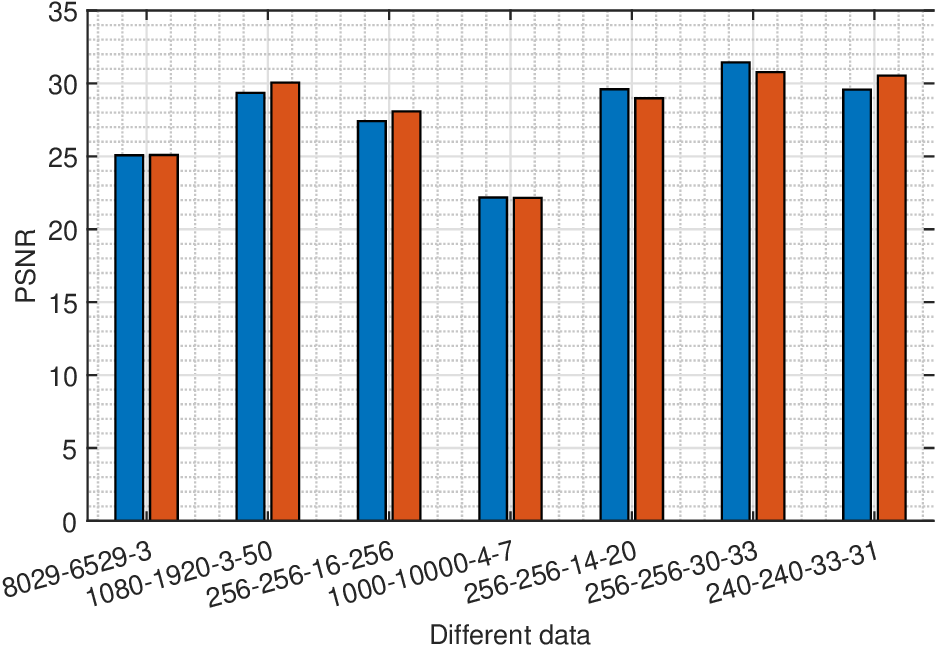} &
\includegraphics[width=1.7in, height=1.46514in]{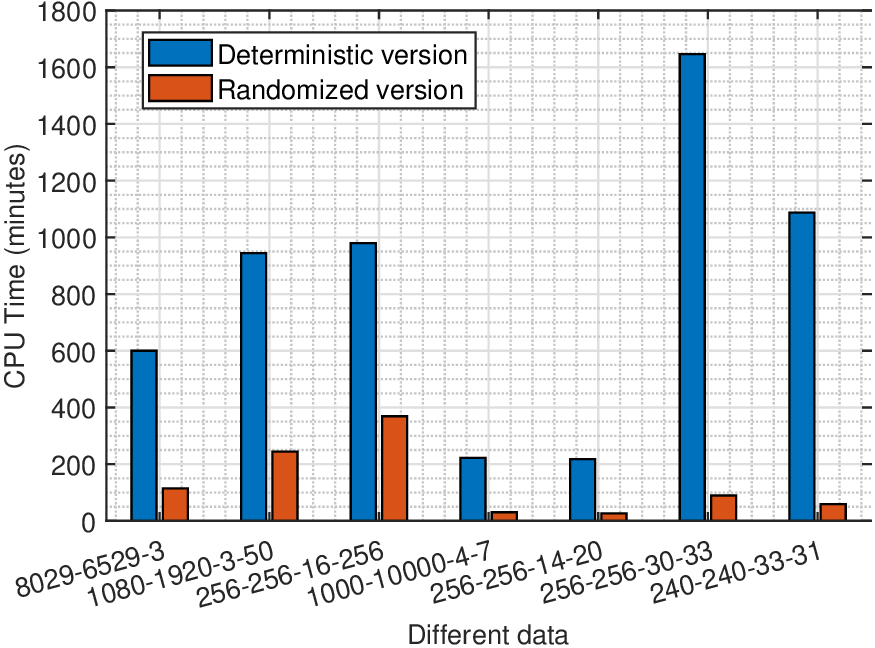}&
\includegraphics[width=1.7in, height=1.46514in]{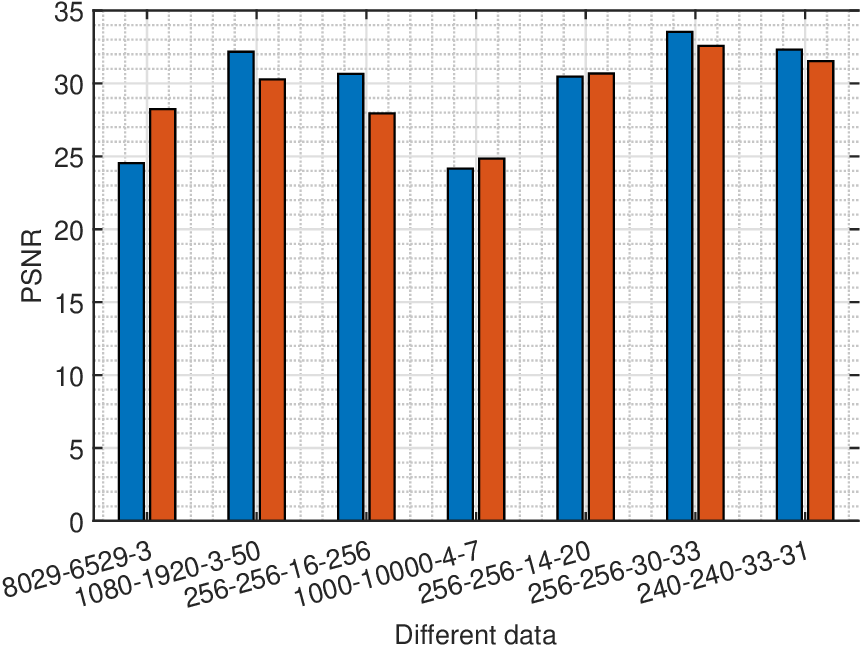}
%
\\
\cmidrule(rl){1-2} \cmidrule(rl){3-4} 
\\
\multicolumn{2}{c}{\text{SR}$=0.5\%$}   & \multicolumn{2}{c}{\text{SR}$=1\%$} 
\\
\end{tabular}
\caption{The   tensor data type
  (\textit{from left to right are   color image, color video,  multi-temporal hyperspectral images,
magnetic resonance image, face dataset, hyperspectral video, respectively})
versus  CPU Time  and  PSNR in LRTC task.
The sampling rates 
are set to be {\text{SR}$=0.5\%$} and {\text{SR}$=1\%$},
respectively.
While maintaining comparable accuracy,
%
 the randomized version is on average 9X faster than the deterministic one,
and in some individual cases, it achieves up to 20X speedup.
 }
\label{INTROtimevsPsnr}
\end{figure*}

\begin{table*}[tp]
\renewcommand{\arraystretch}{0.95}
\setlength\tabcolsep{3.5pt}
  \centering
  \caption{Quantitative evaluation PSNR, SSIM, RSE, CPU Time (Second) of various RTC  methods
  on MRSIs.
  }
  \vspace{-0.15cm}
  \label{table-mrsi-rhtc}
  \scriptsize
  \begin{threeparttable}
     \begin{tabular}{cccc cccc cccc cccc cc  c}
    \Xhline{1pt}
   \tabincell{c}   {SR 
    $\textbf{\&}$ 
   NR}&
   \tabincell{c} {Evaluation\\ Metric}&
\tabincell{c}{TRNN  \\    \cite{huang2020robust}}&
\tabincell{c}{TTNN  \\    \cite{song2020robust}}&
\tabincell{c} {TSPK \\    \cite{lou2019robust}}&
   \tabincell{c}{TTLRR  \\ \cite{yang2024robust22}}& 
\tabincell{c}{LNOP \\  \cite{chen2020robust}}&
 \tabincell{c}{NRTRM\\  \cite{qiu2021nonlocal}}&
\tabincell{c}{HWTNN  \\    \cite{qin2021robust}}&
 \tabincell{c} {HWTSN  \\  \cite{qin2023nonconvex}}&
 \tabincell{c} {R-HWTSN 
  \\  \cite{qin2023nonconvex}}&
  \tabincell{c} {TCTV-RTC \\ \cite{wang2023guaranteed}  }
 & \tabincell{c} {{FCTN} \\ {-GNRTC}}&
\tabincell{c}{R1-FCTN \\ -GNRTC} & \tabincell{c}{R2-FCTN \\ -GNRTC}

    \cr
    \Xhline{1pt}
\hline
     \hline

   \multirow{4}{*}{    \tabincell{c}   {SR=0.2\\ $\textbf{\&}$ \\NR=1/3}  } &MPSNR
  &   
  23.5808&25.0459&25.1370&25.3187&26.1520&25.1798&26.2033&26.5120&26.1580&28.5136&28.4088&\textcolor[rgb]{1,0,0}{\textbf{28.8936}}&\textcolor[rgb]{0,0,1}{\textbf{28.8736}}
  \cr
   \qquad	&MSSIM	& 
   0.5218&0.6127&0.6151&0.5464&0.6181&0.6340&0.6075&0.6370&0.6177&\textcolor[rgb]{1,0,0}{\textbf{0.7446}}&\textcolor[rgb]{0,0,1}{\textbf{0.7366}}&0.7365&0.7293
   \cr
   \qquad	&MRSE	&
   0.2446&0.2315&0.2289&0.2189&0.2041&0.2307&0.1945&0.2000&0.1983&0.1643&0.1510&\textcolor[rgb]{1,0,0}{\textbf{0.1386}}&\textcolor[rgb]{0,0,1}{\textbf{0.1392}}
   \cr
    \qquad	&MTime	&  
    3993.67&\textcolor[rgb]{0,0,1}{\textbf{1524.45}}&{3046.91}&3826.11&2484.09&2099.35&2057.09&2684.24&\textcolor[rgb]{1,0,0}{\textbf{1497.06}}&4111.31&15397.15&3150.63&3532.49
    \cr

   \hline

  \multirow{4}{*}{    \tabincell{c}   {SR=0.2  \\ $\textbf{\&}$ \\NR=0.5}  } &MPSNR
  &
  21.6410&23.3963&23.5539&23.2420&24.1864&23.8436&24.2885&23.7792&23.6564&26.4475&26.4772&\textcolor[rgb]{0,0,1}{\textbf{27.1675}}&\textcolor[rgb]{1,0,0}{\textbf{27.2392}}
  \cr
   \qquad	&MSSIM	& 
   0.4911&0.4813&0.5010&0.4247&0.5323&0.5515&0.4940&0.5066&0.5090&\textcolor[rgb]{1,0,0}{\textbf{0.6737}}&0.6595&\textcolor[rgb]{0,0,1}{\textbf{0.6704}}&0.6618
   \cr
   \qquad	&MRSE	&
    0.3122&0.2677&0.2643&0.2586&0.2523&0.2608&0.2336&0.2594&0.2619&0.2094&0.1871&\textcolor[rgb]{0,0,1}{\textbf{0.1663}}&\textcolor[rgb]{1,0,0}{\textbf{0.1660}}
    \cr
    \qquad	&MTime	& 
    4019.17&\textcolor[rgb]{0,0,1}{\textbf{1569.65}}&3027.08&3806.64&2601.71&2088.82&2048.02&2673.53&\textcolor[rgb]{1,0,0}{\textbf{1508.72}}&3986.73&14095.62&3154.07&3337.86
    \cr

     \hline

     \multirow{4}{*}{    \tabincell{c}   {SR=0.1 \\ $\textbf{\&}$ \\NR=1/3}  } &MPSNR
  &   

  20.8668&23.5098&23.5289&23.9632&23.8890&23.6531&23.9005&24.0602&23.9741&26.0960&26.1907&\textcolor[rgb]{1,0,0}{\textbf{26.5446}}&\textcolor[rgb]{0,0,1}{\textbf{26.5185}}

   \cr
   \qquad	&MSSIM	&
   0.4904&0.5281&0.5283&0.4840&0.4909&0.5482&0.4931&0.5064&0.5002&\textcolor[rgb]{1,0,0}{\textbf{0.6553}}&\textcolor[rgb]{0,0,1}{\textbf{0.6518}}&0.6494&0.6333

   \cr
   \qquad	&MRSE	& 
   0.3387&0.2711&0.2687&0.2398&0.2518&0.2684&0.2518&0.2473&0.2476&0.2171&0.1999&\textcolor[rgb]{1,0,0}{\textbf{0.1810}}&\textcolor[rgb]{0,0,1}{\textbf{0.1823}}

   \cr
    \qquad	&MTime	&   
    3942.03&\textcolor[rgb]{0,0,1}{\textbf{1719.65}}&3176.77&3550.68&2213.47&2113.38&1994.28&2654.98&\textcolor[rgb]{1,0,0}{\textbf{1194.43}}&4108.24&14158.85&2485.03&2567.18
    \cr

     \hline
     \multirow{4}{*}{    \tabincell{c}   {SR=0.1  \\ $\textbf{\&}$ \\NR=0.5}  } &MPSNR
  &  
  19.7638&22.1445&22.0155&21.3846&21.8048&22.5362&22.3005&22.1249&22.0964&24.3981&24.2819&\textcolor[rgb]{0,0,1}{\textbf{24.8372}}&\textcolor[rgb]{1,0,0}{\textbf{24.8560}}

  \cr
   \qquad	&MSSIM	& 
   0.4724&0.4208&0.4306&0.3155&0.4141&0.4767&0.4042&0.3994&0.4069&0.5804&\textcolor[rgb]{0,0,1}{\textbf{0.5830}}&\textcolor[rgb]{1,0,0}{\textbf{0.5915}}&0.5674

   \cr
   \qquad	&MRSE	&
   0.3802&0.3121&0.3176&0.2884&0.3380&0.3051&0.3003&0.2963&0.2990&0.2644&0.2576&\textcolor[rgb]{1,0,0}{\textbf{0.2256}}&\textcolor[rgb]{0,0,1}{\textbf{0.2265}}

    \cr
    \qquad	&MTime	&
    3967.29&\textcolor[rgb]{0,0,1}{\textbf{1731.32}}&3207.05&3482.18&2268.15&2118.69&1982.68&2636.47&\textcolor[rgb]{1,0,0}{\textbf{1203.59}}&4010.51&\textcolor[rgb]{0,0,0}{14412.76}&2495.99&2587.71
    \cr

   \hline
     \hline

      \Xhline{1pt}
    \end{tabular}
    \end{threeparttable}
\end{table*}

\begin{figure*}[!htbp]
\renewcommand{\arraystretch}{0.25}
\setlength\tabcolsep{3pt}
\centering
\begin{tabular}{c|c|c }
\centering

\includegraphics[width=2.3in, height=1.7912in]{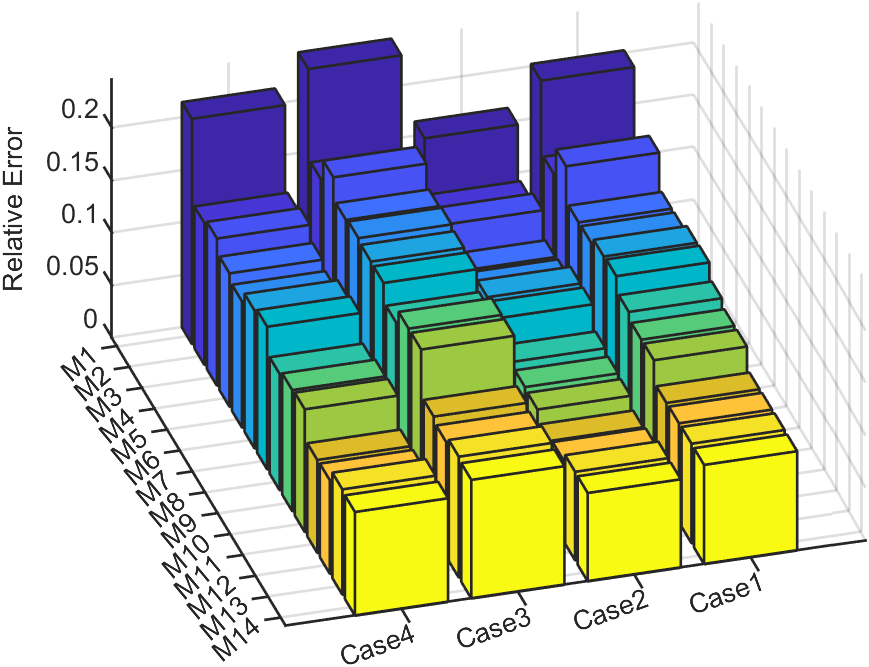}&
\includegraphics[width=2.3in, height=1.7912in]{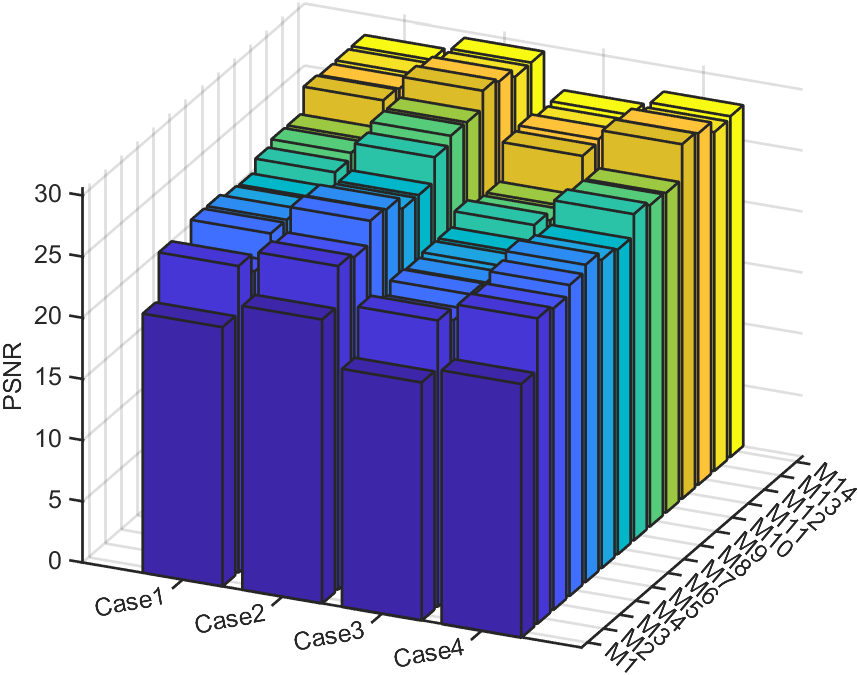}&
\includegraphics[width=2.3in, height=1.7912in]{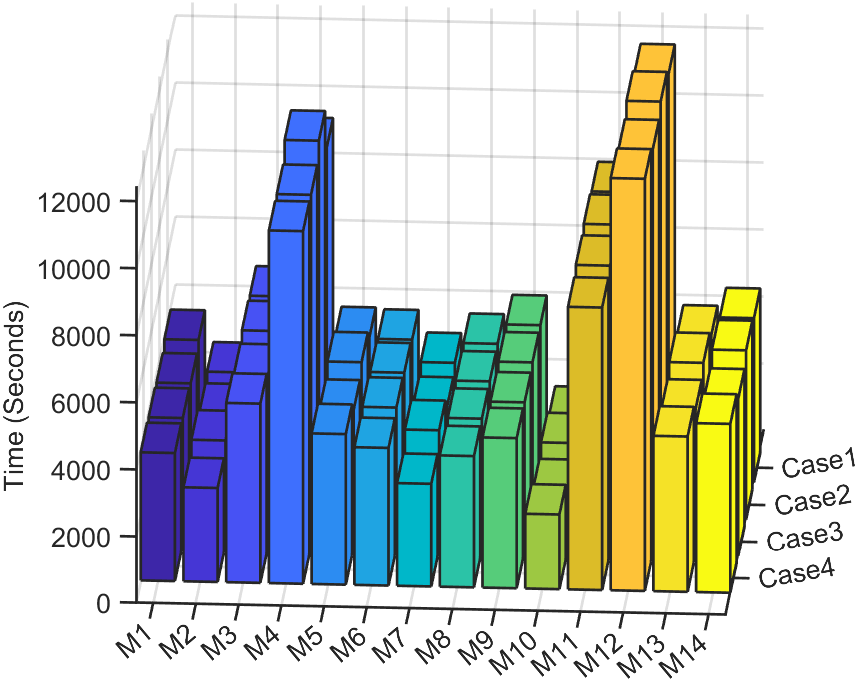}

\end{tabular}
\vspace{-0.15cm}
\caption{
Quantitative evaluation MPSNR, MSSIM, MRSE and CPU Time (Second) of various RTC methods on four color videos.
\textbf{X label:}
(M1) TRNN,  
(M2) UTNN,
  (M3) TSPK,    (M4) TTLRR, (M5) LNOP, (M6) NRTRM, (M7) BCNRTC,
(M8) HWTNN,  (M9) HWTSN, (M10)  R-HWTSN, (M11) TCTV-RTC, (M12) FCTN-GNRTC,  (M13) R1-FCTN-GNRTC,  (M14) R2-FCTN-GNRTC.
\textbf{Y label:} (Case 1) $ SR= 5\%,  SNR= 3dB$, (Case 2) $SR= 10 \%,  SNR= 3dB$,  (Case 3) $SR= 5\%,  SNR= 2dB$,  (Case 4) $SR= 10 \%,  SNR= 2dB$.
}
\label{rhtc_cv}
\end{figure*}

\begin{figure*}[!htbp]
\renewcommand{\arraystretch}{0.5}
\setlength\tabcolsep{0.43pt}
\centering
\begin{tabular}{ccc  ccc cc }
\\
 \toprule

(a)   &
  (b)  & (c) &
(d) & (e)
 &(f)& (g)&(h)
 \\
\includegraphics[ width=0.875in, height=0.951362in]{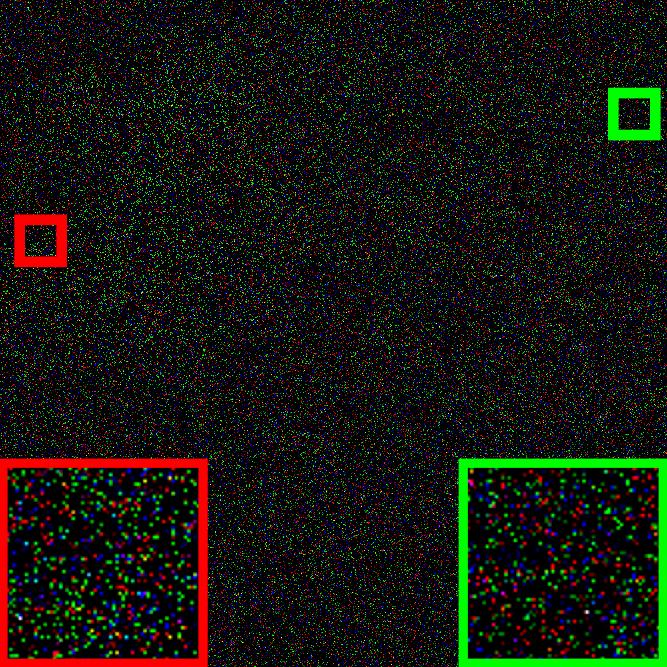} %
&
\includegraphics[ width=0.875in, height=0.951362in]{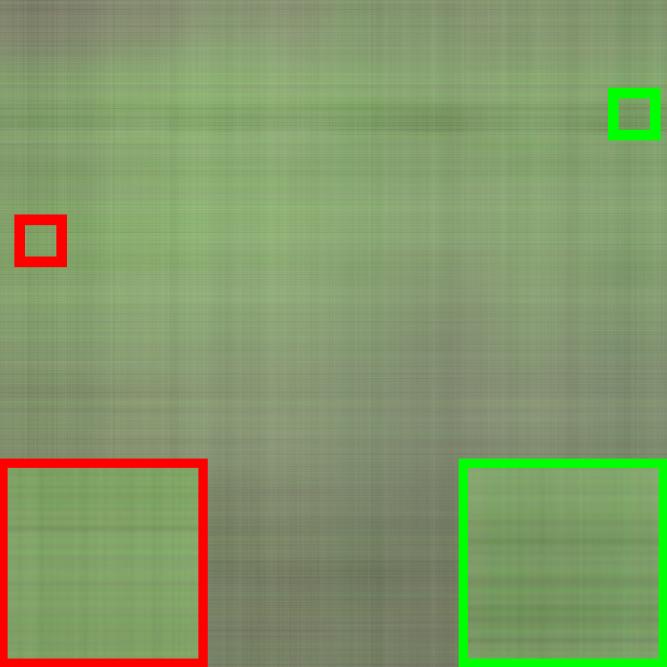}
&
\includegraphics[ width=0.875in, height=0.951362in]{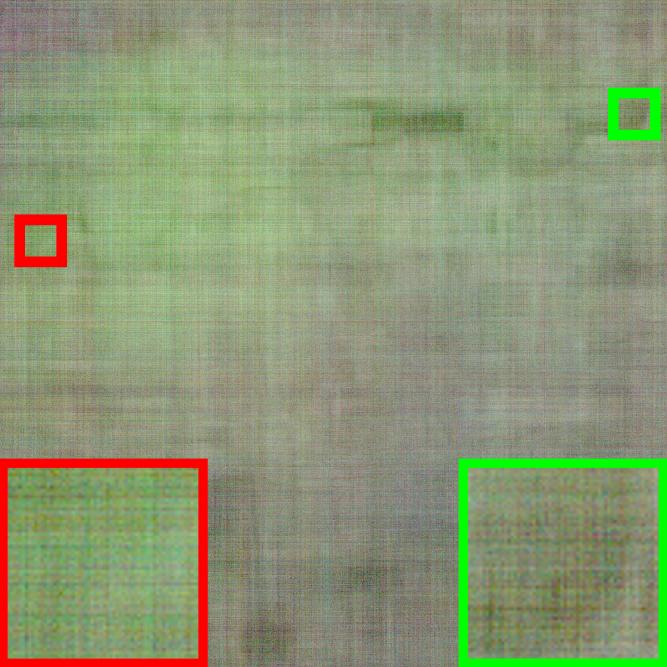}
&
\includegraphics[width=0.875in, height=0.951362in]{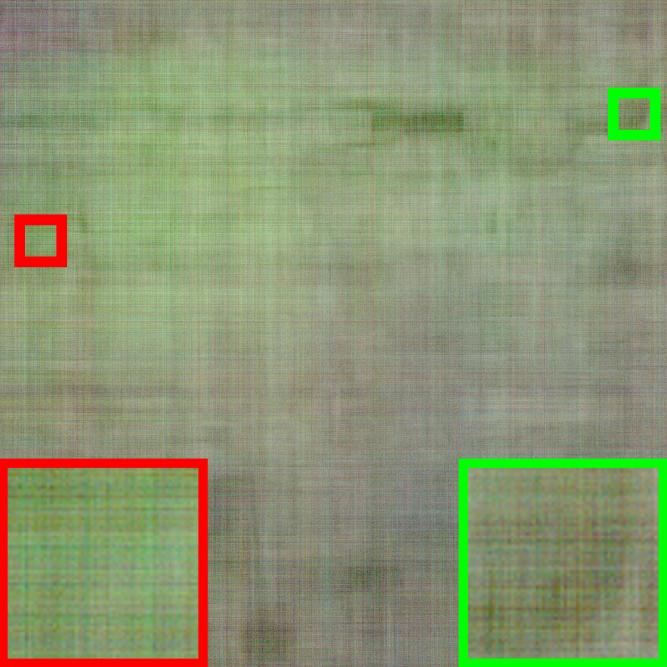}
&
\includegraphics[width=0.875in, height=0.951362in]{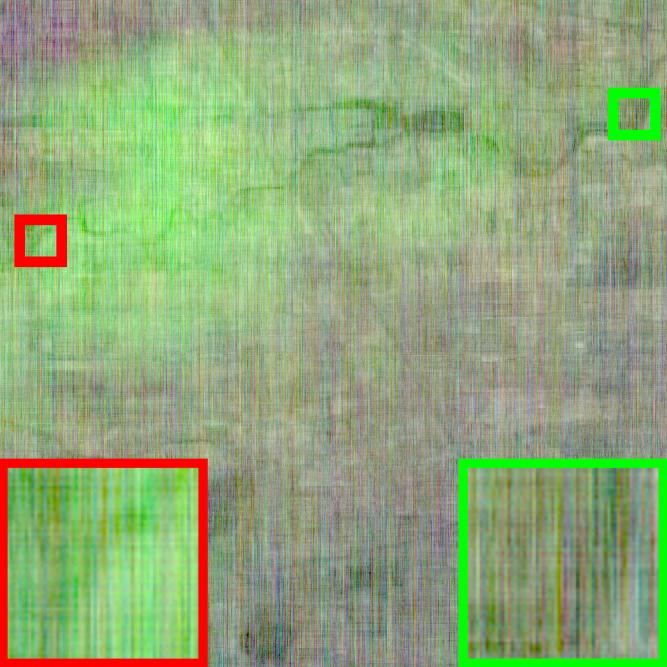}
&
\includegraphics[ width=0.875in, height=0.951362in]{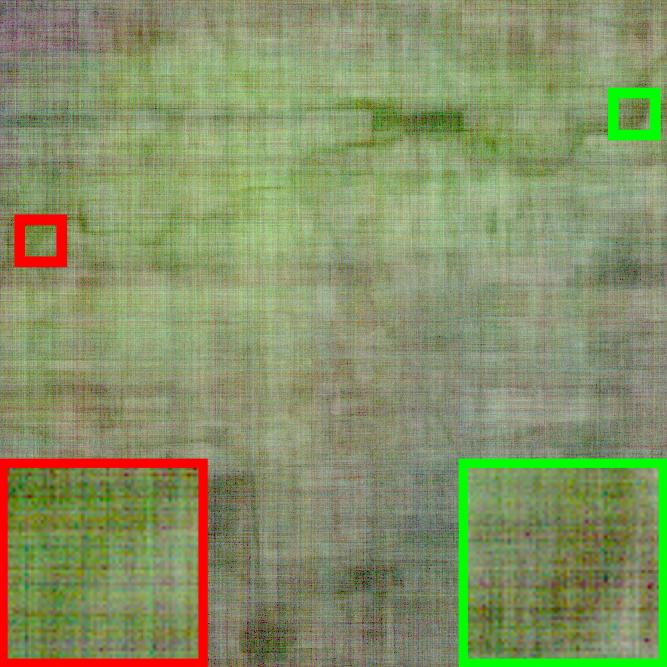}
&
\includegraphics[ width=0.875in, height=0.951362in]{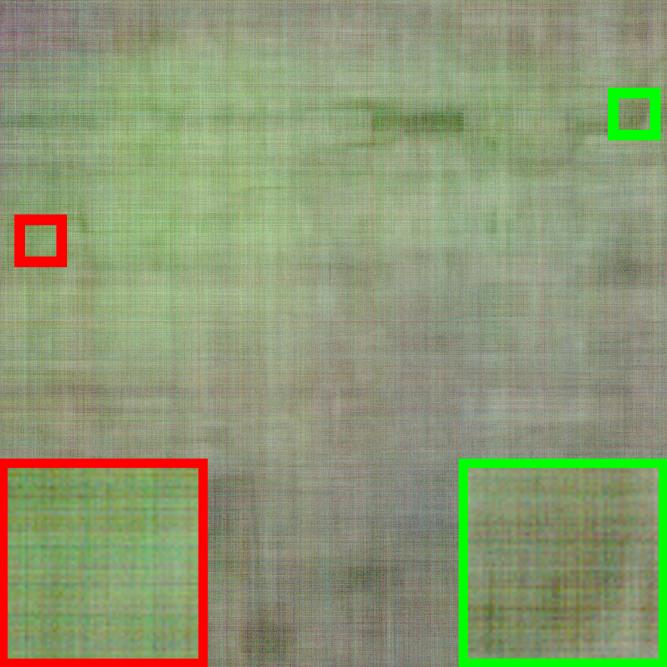}
&
\includegraphics[ width=0.875in, height=0.951362in]{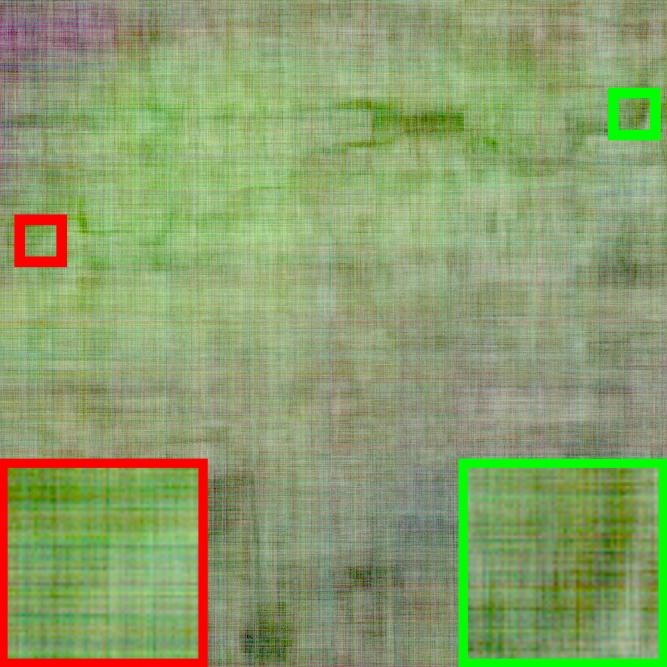}
\\

 (i)
 &  (j)
  &
 (k) &(l)&(m) & (n)& (o) &\\
\includegraphics[  width=0.875in, height=0.951362in]{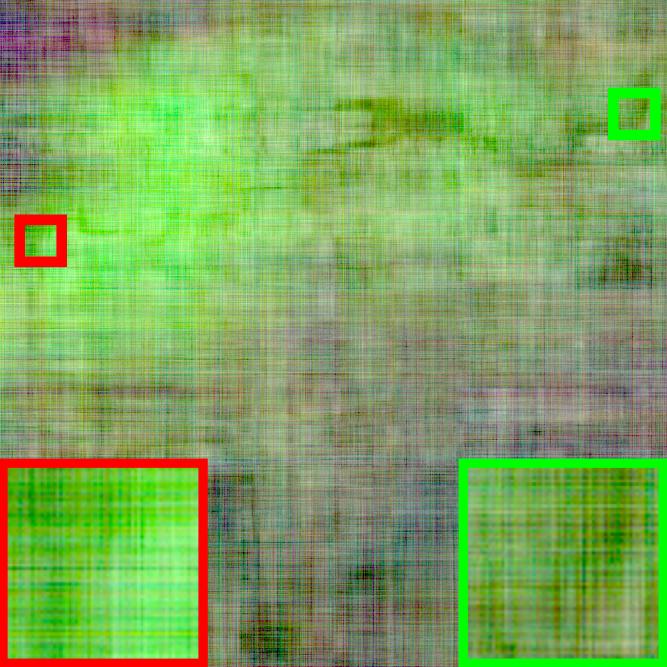}
&
\includegraphics[ width=0.875in, height=0.951362in]{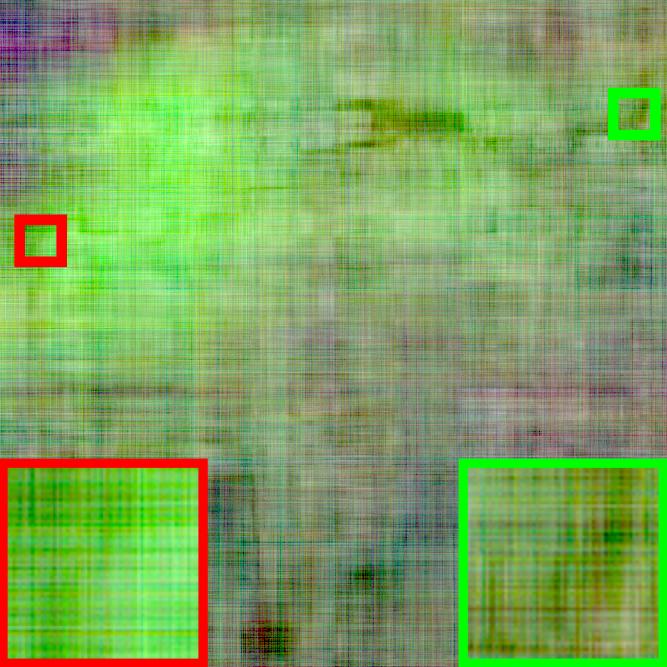}
&
\includegraphics[  width=0.875in, height=0.951362in]{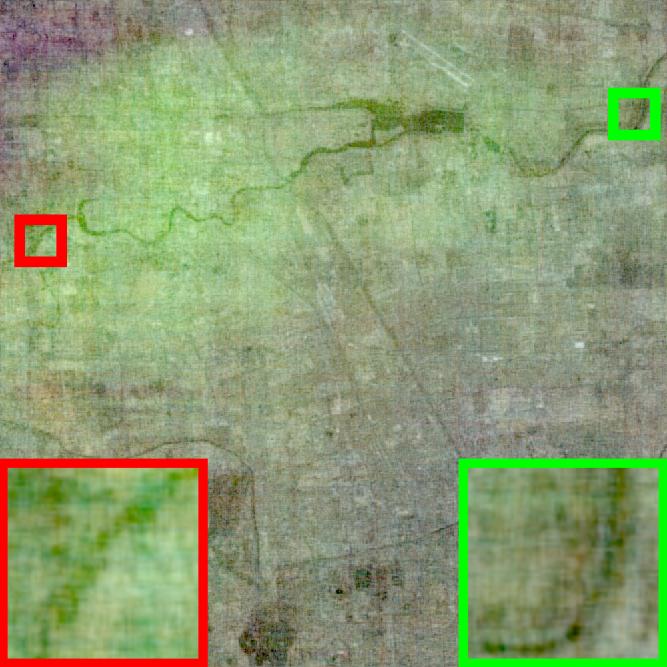}
&
\includegraphics[  width=0.875in, height=0.951362in]{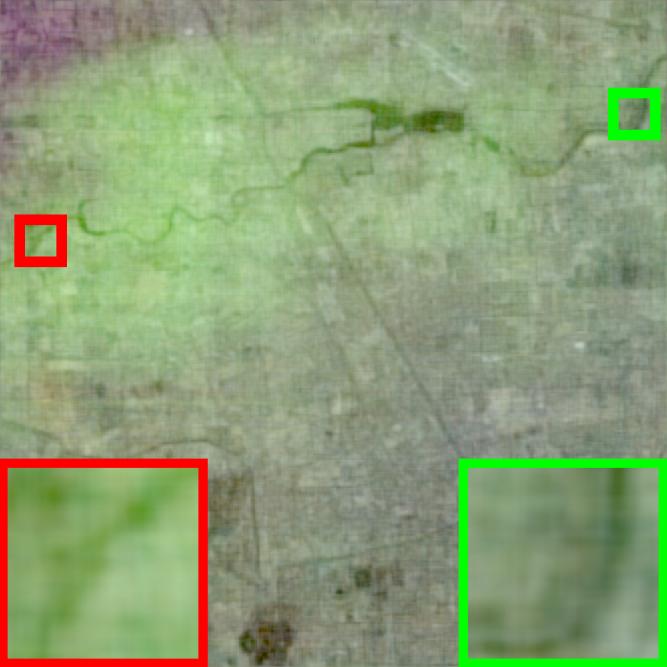}
&
\includegraphics[ width=0.875in, height=0.951362in]{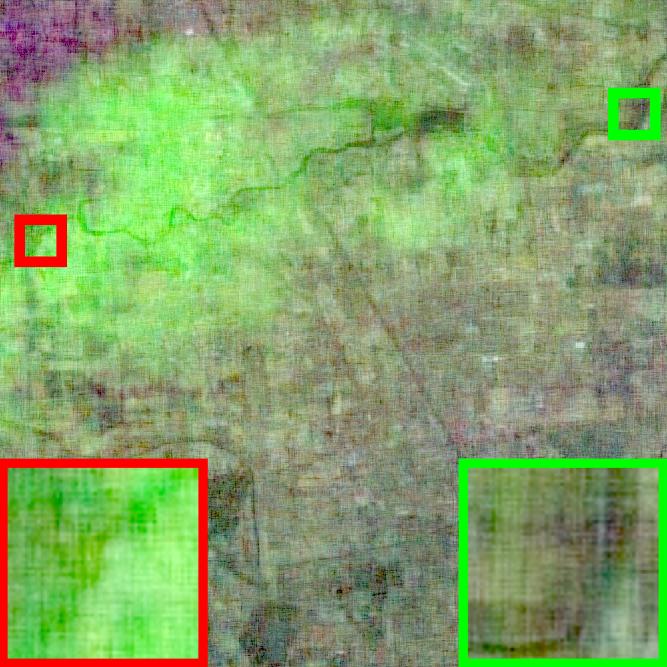}
&
\includegraphics[  width=0.875in, height=0.951362in]{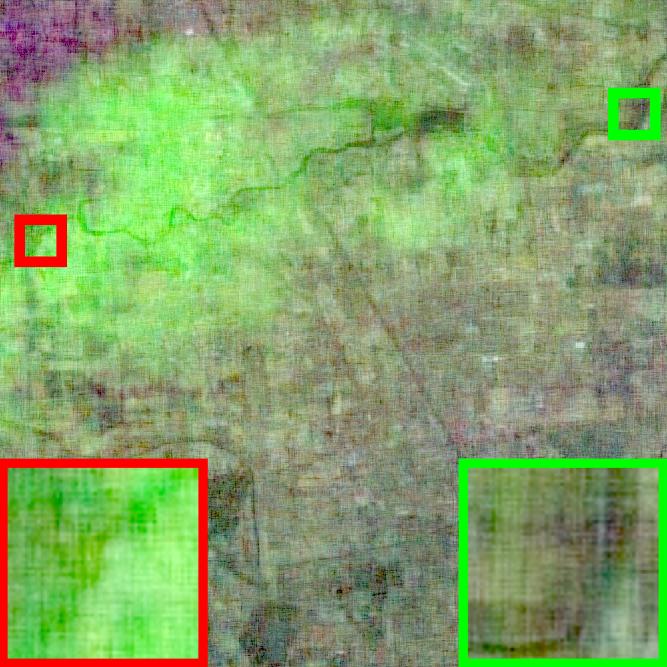}
&
\includegraphics[  width=0.875in, height=0.951362in]{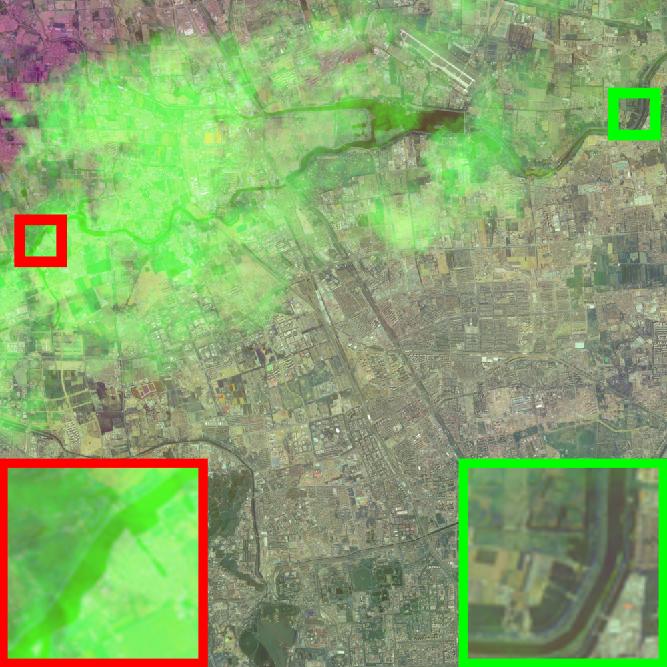} &

\\
 \toprule
\end{tabular}
\caption{
Visual comparison of various RTC methods for  MRSIs inpainting under 
  $(SR, NR)= (0.1, 0.5)$.
From left to right: (a) Observed, (b) TRNN,
(c)  TTNN, (d) TSPK, (e) TTLRR, (f) LNOP, (g) NRTRM, (h) HWTNN,  (i) HWTSN,  (j) R-HWTSN, 
 (k)  TCTV-RTC, (l) FCTN-GNRTC, (m) R1-FCTN-GNRTC,  (n) R2-FCTN-GNRTC, (o) Ground-truth.}
\label{fig_MRSI2}
\end{figure*}

\begin{figure*}[!htbp]
\renewcommand{\arraystretch}{0.5}
\setlength\tabcolsep{0.43pt}
\centering
\begin{tabular}{ccc  ccc cc }
\\
 \toprule

(a)   &
  (b)  & (c) &
(d) & (e)
 &(f)& (g)&(h)
 \\
\includegraphics[ width=0.875in, height=0.95875in]{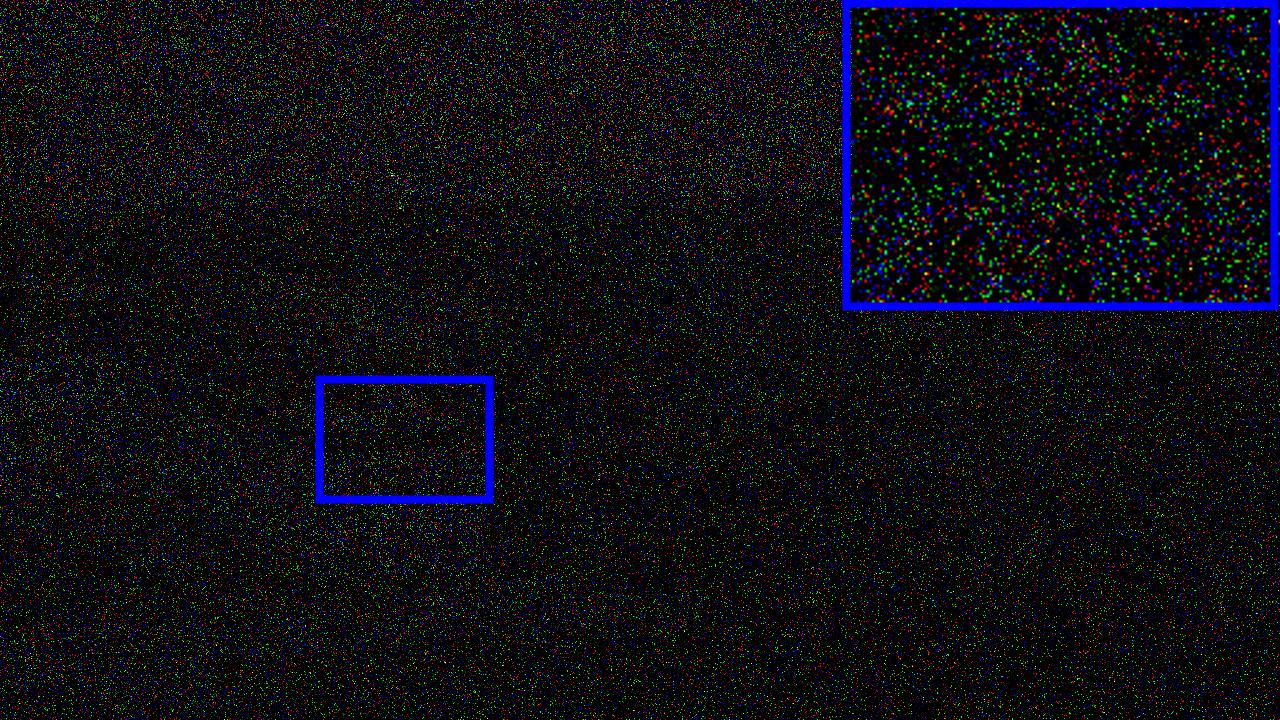} %
&
\includegraphics[ width=0.875in, height=0.95875in]{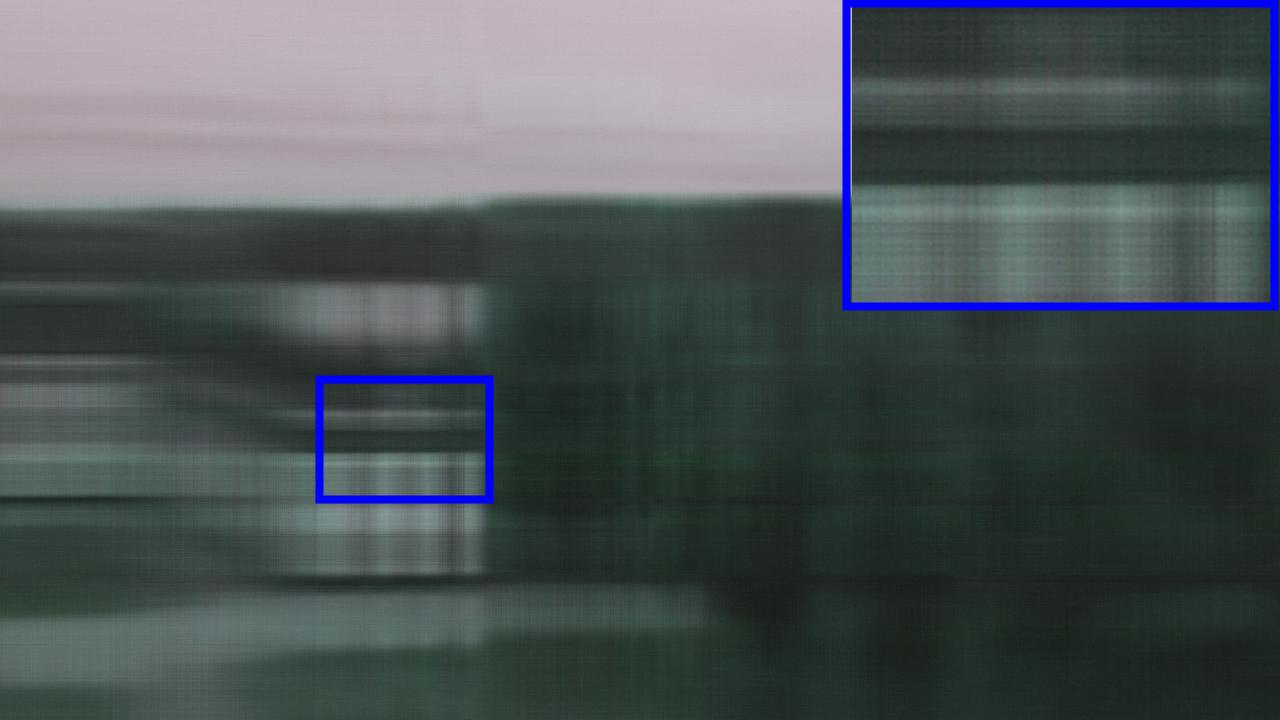}
&
\includegraphics[ width=0.875in, height=0.95875in]{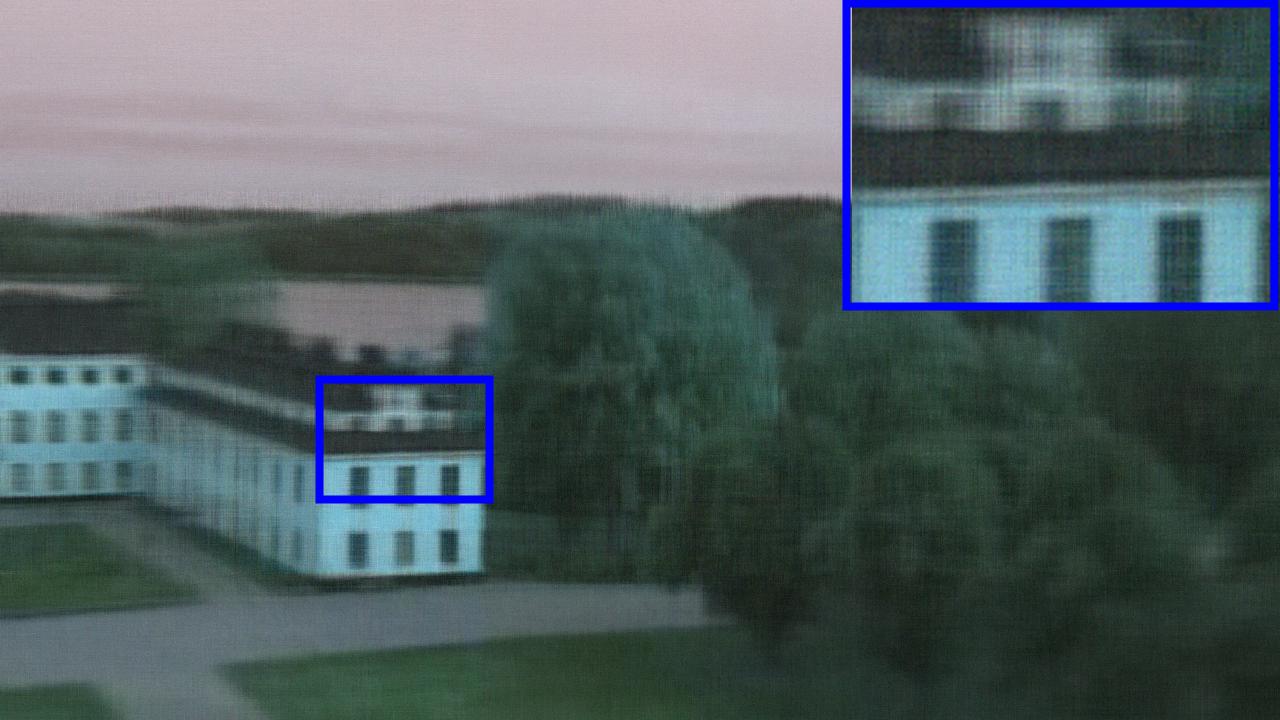}
&
\includegraphics[width=0.875in, height=0.95875in]{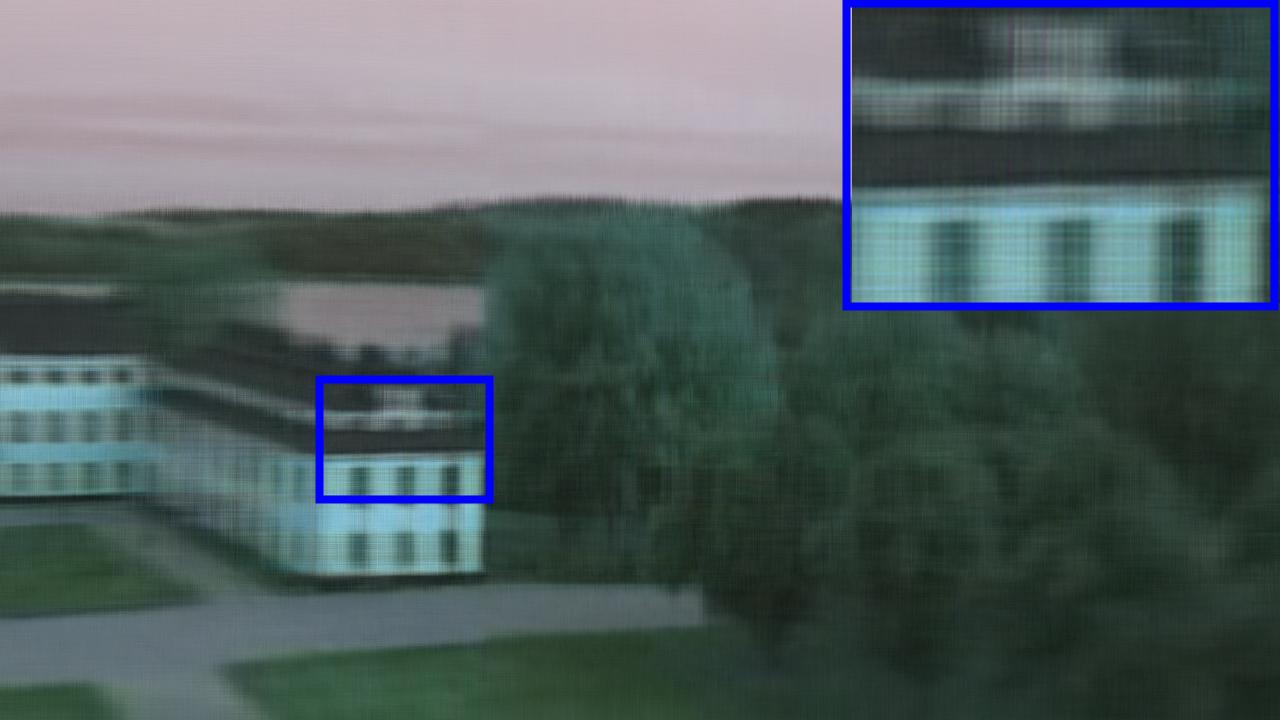}
&
\includegraphics[width=0.875in, height=0.95875in]{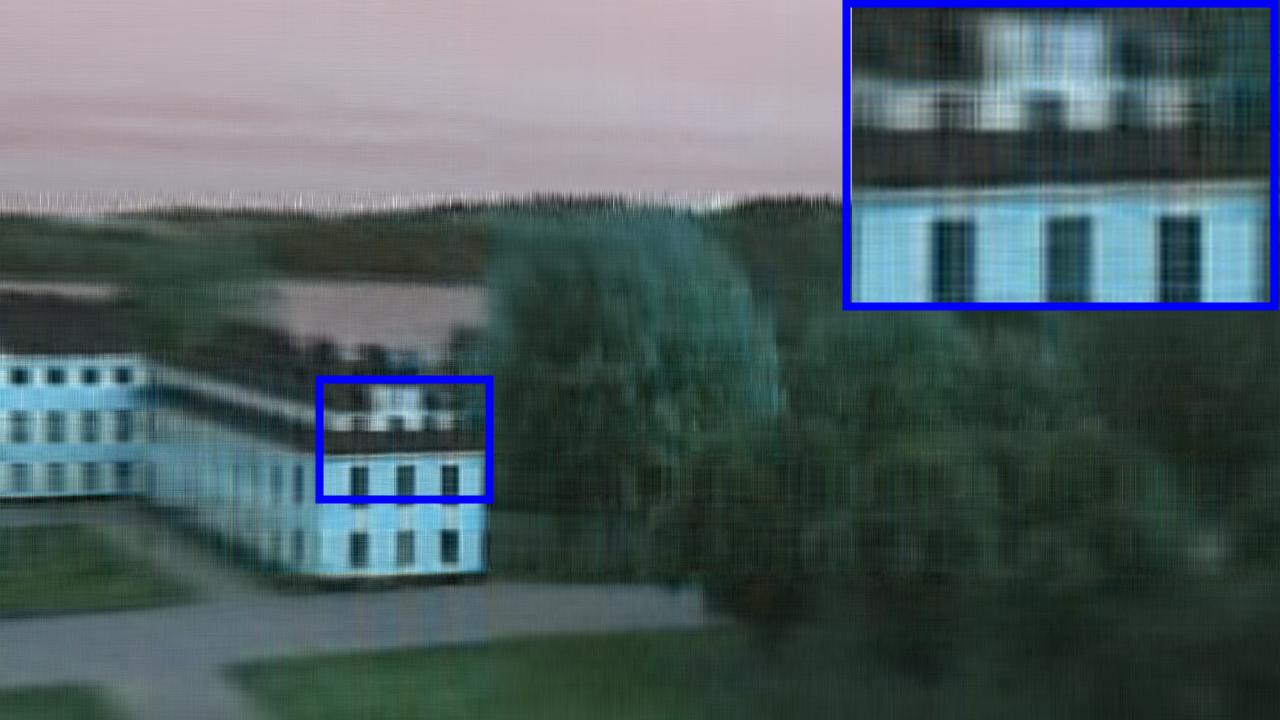}
&
\includegraphics[ width=0.875in, height=0.95875in]{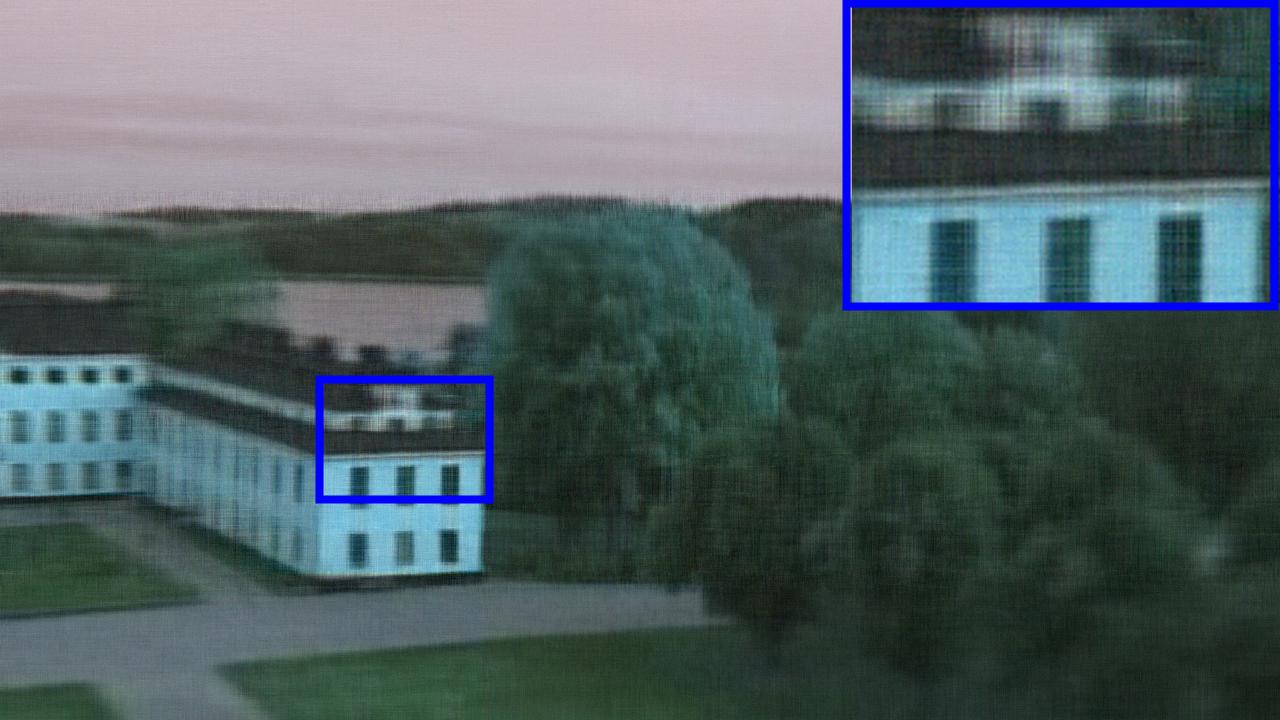}
&
\includegraphics[ width=0.875in, height=0.95875in]{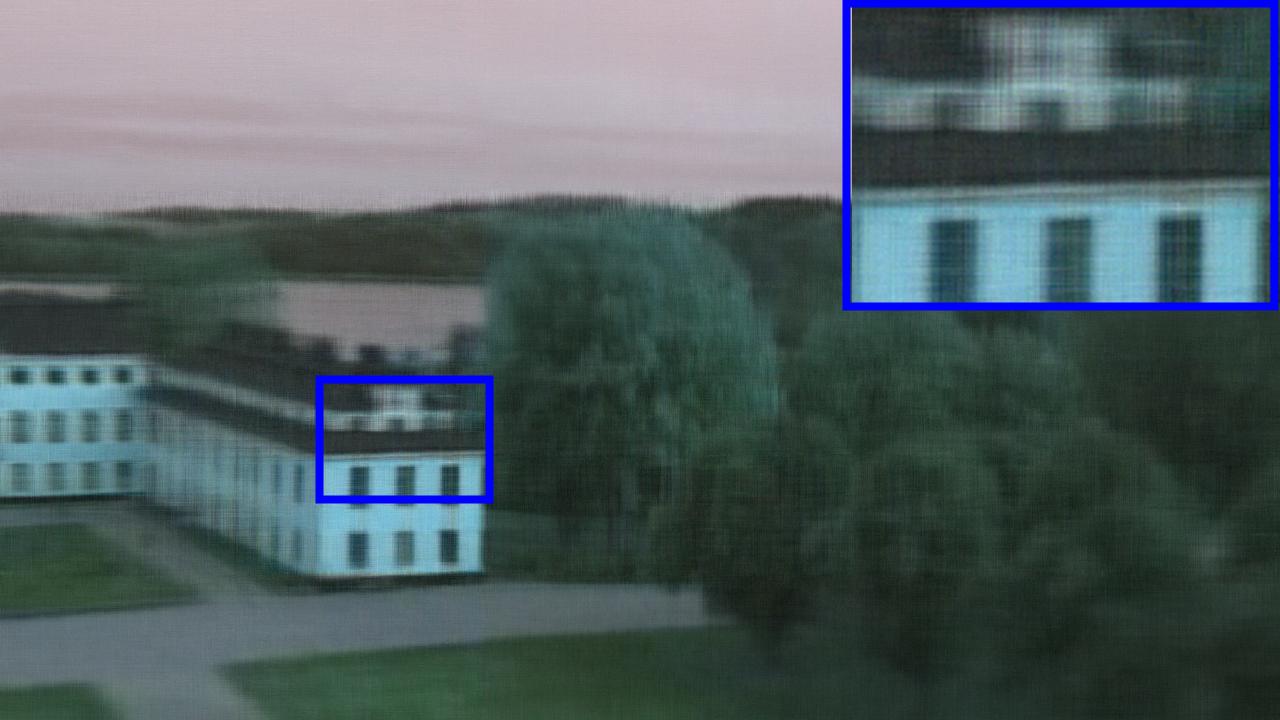}
&
\includegraphics[ width=0.875in, height=0.95875in]{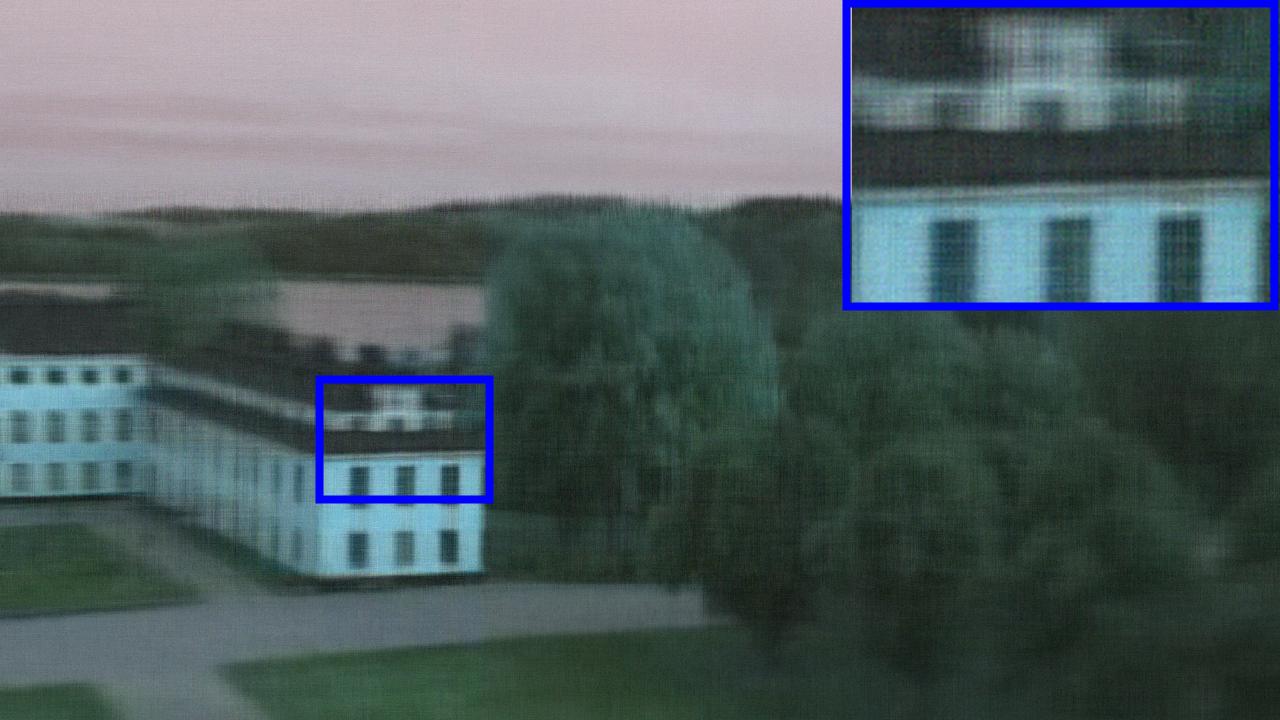}
\\

 (i)
 &  (j)
  &
 (k) &(l)&(m) & (n)& (o) &(p)\\
\includegraphics[  width=0.875in, height=0.95875in]{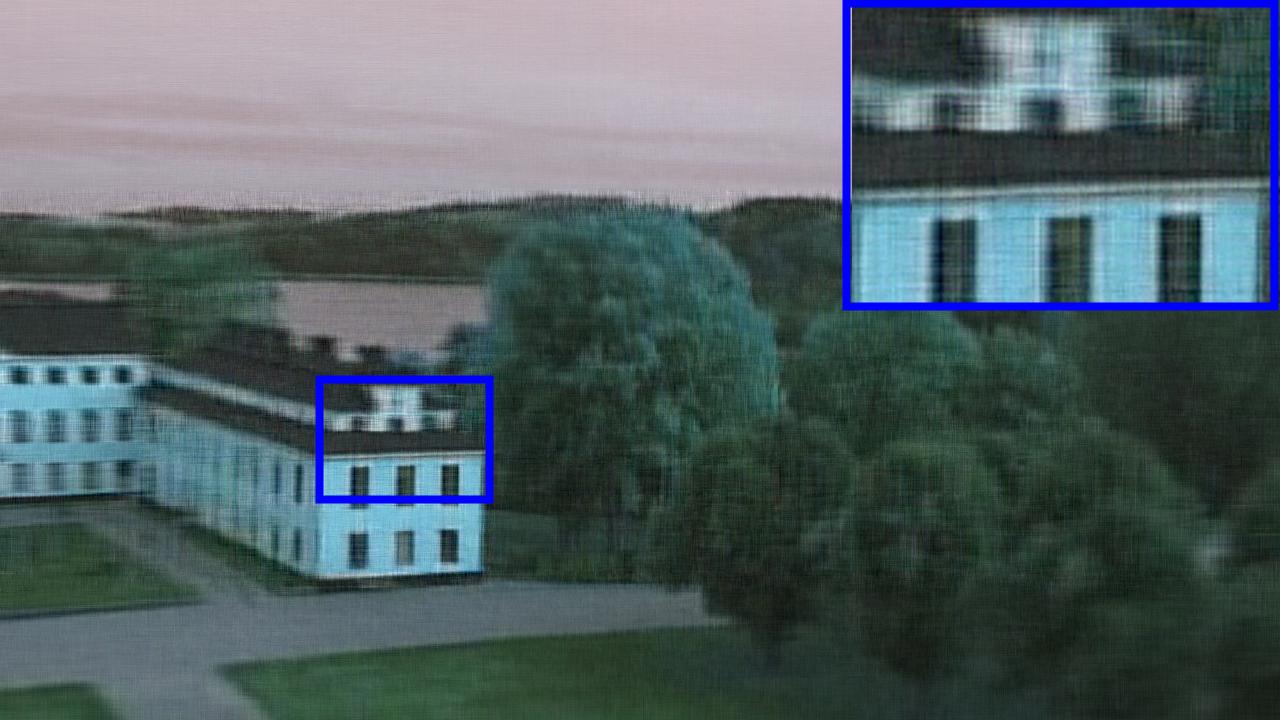}
&
\includegraphics[ width=0.875in, height=0.95875in]{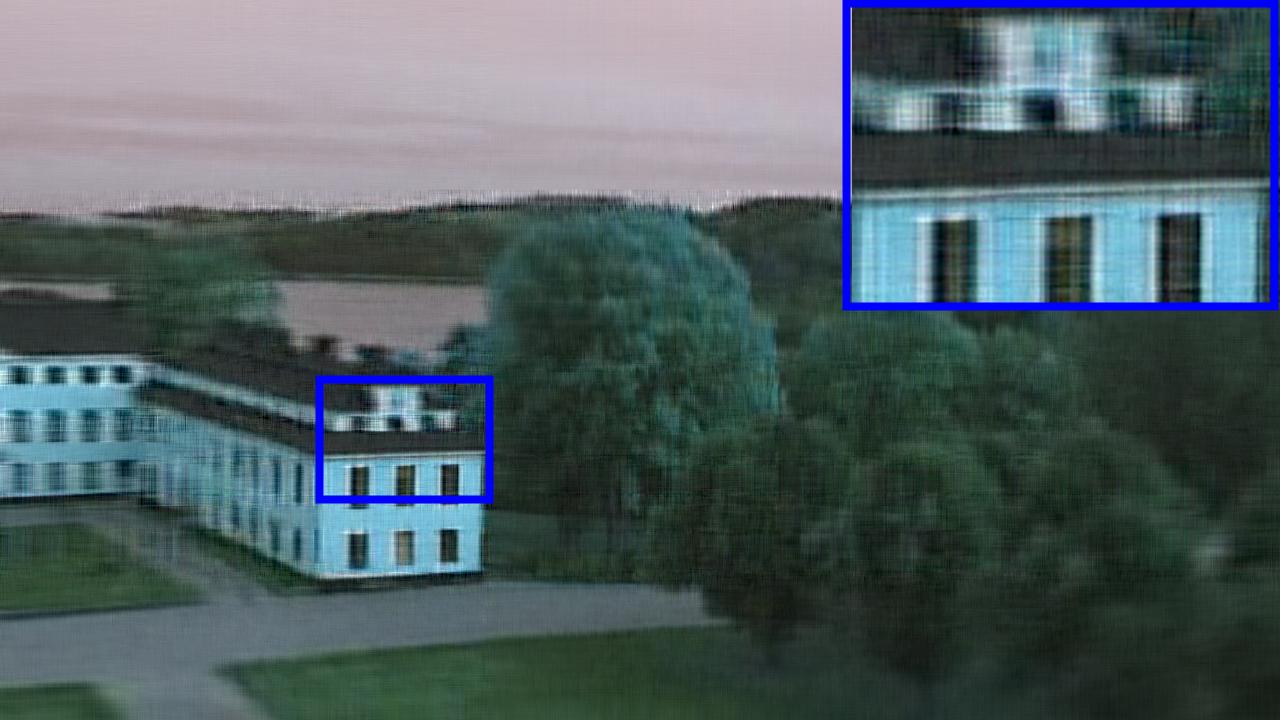}
&
\includegraphics[  width=0.875in, height=0.95875in]{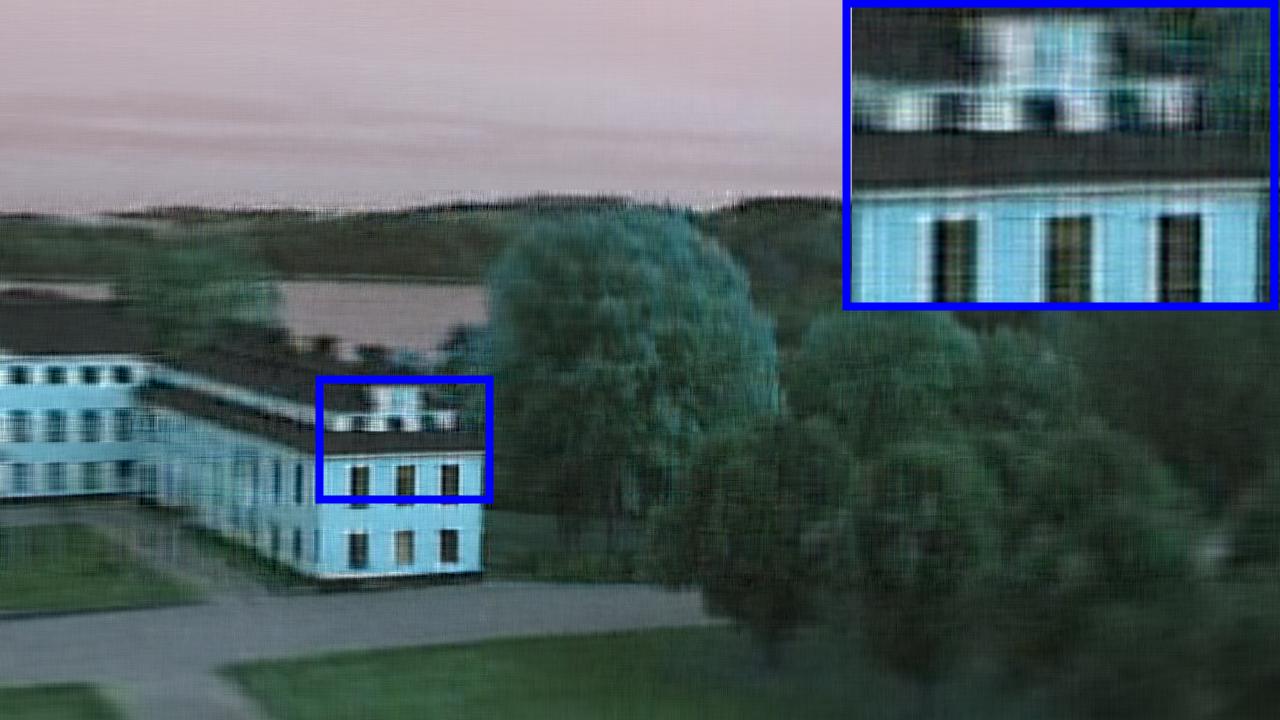}
&
\includegraphics[  width=0.875in, height=0.95875in]{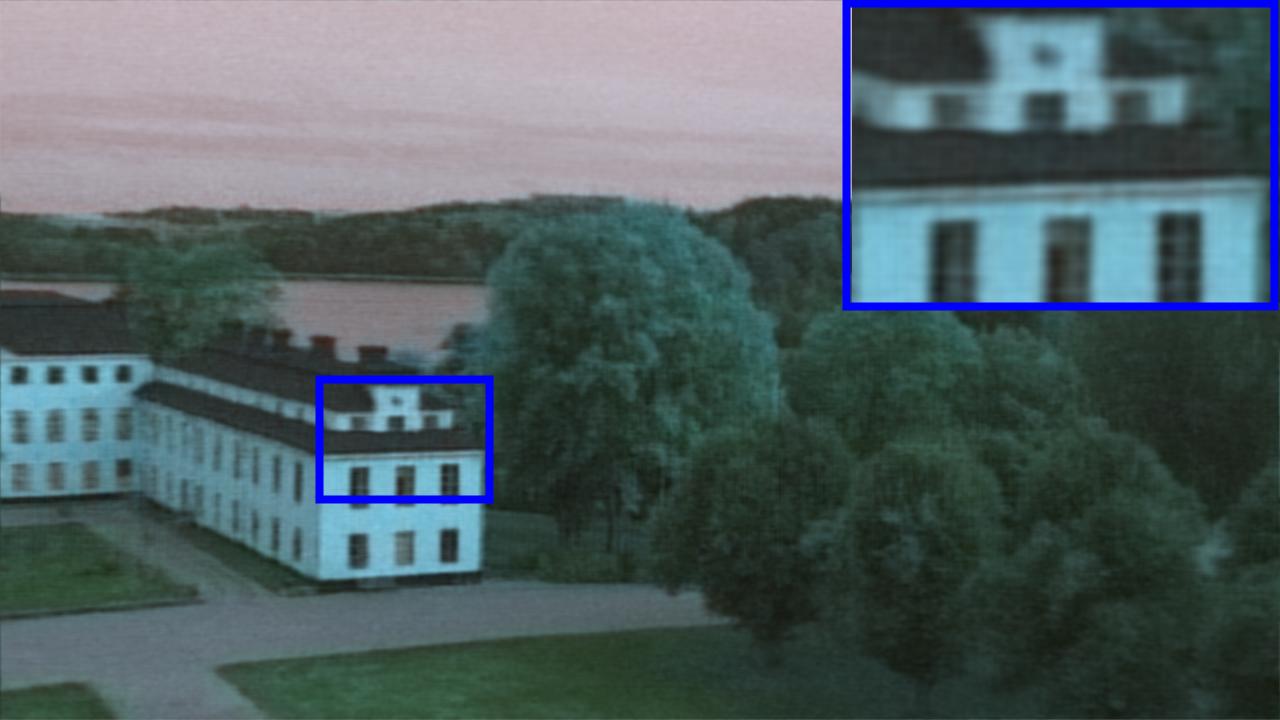}
&
\includegraphics[ width=0.875in, height=0.95875in]{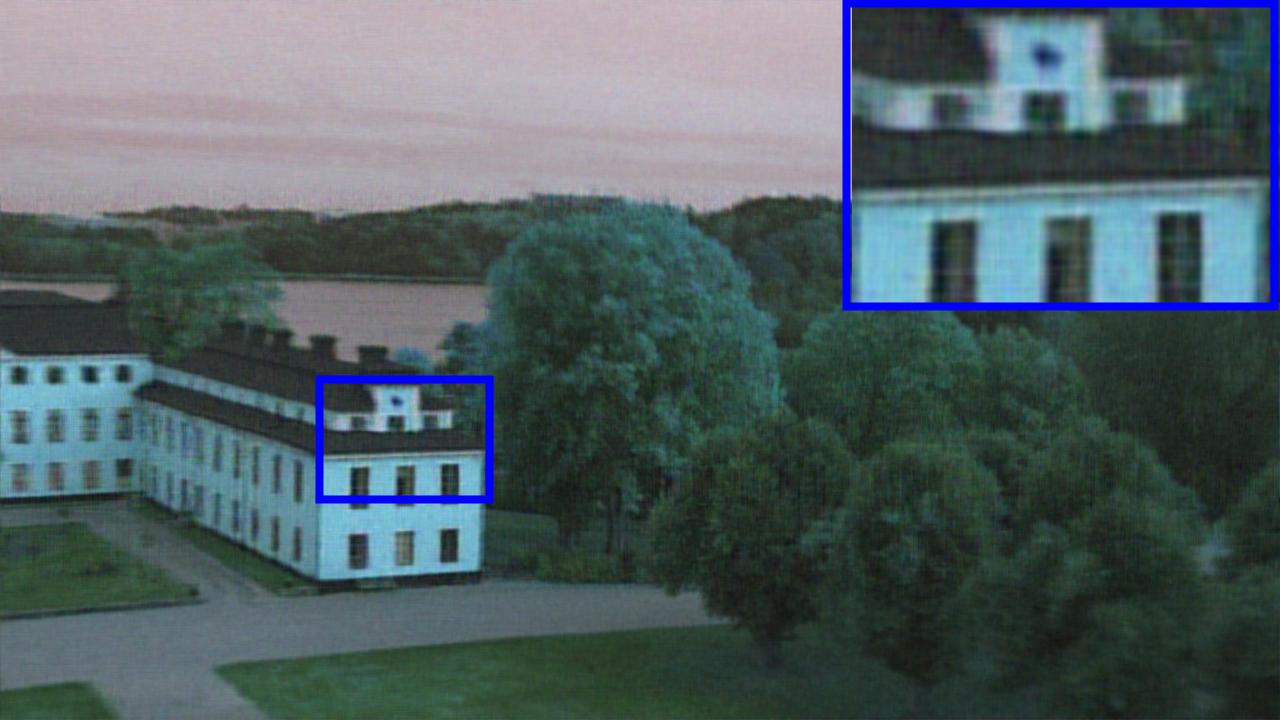}
&
\includegraphics[  width=0.875in, height=0.95875in]{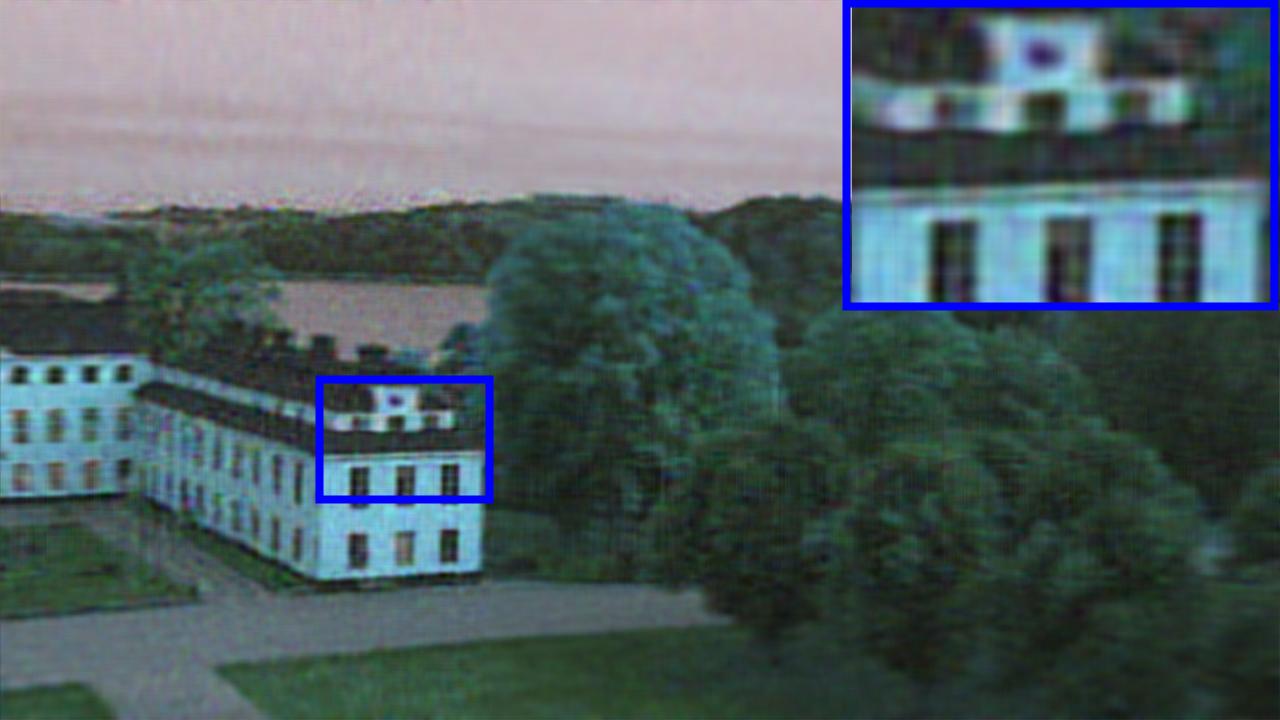}
&
\includegraphics[  width=0.875in, height=0.95875in]{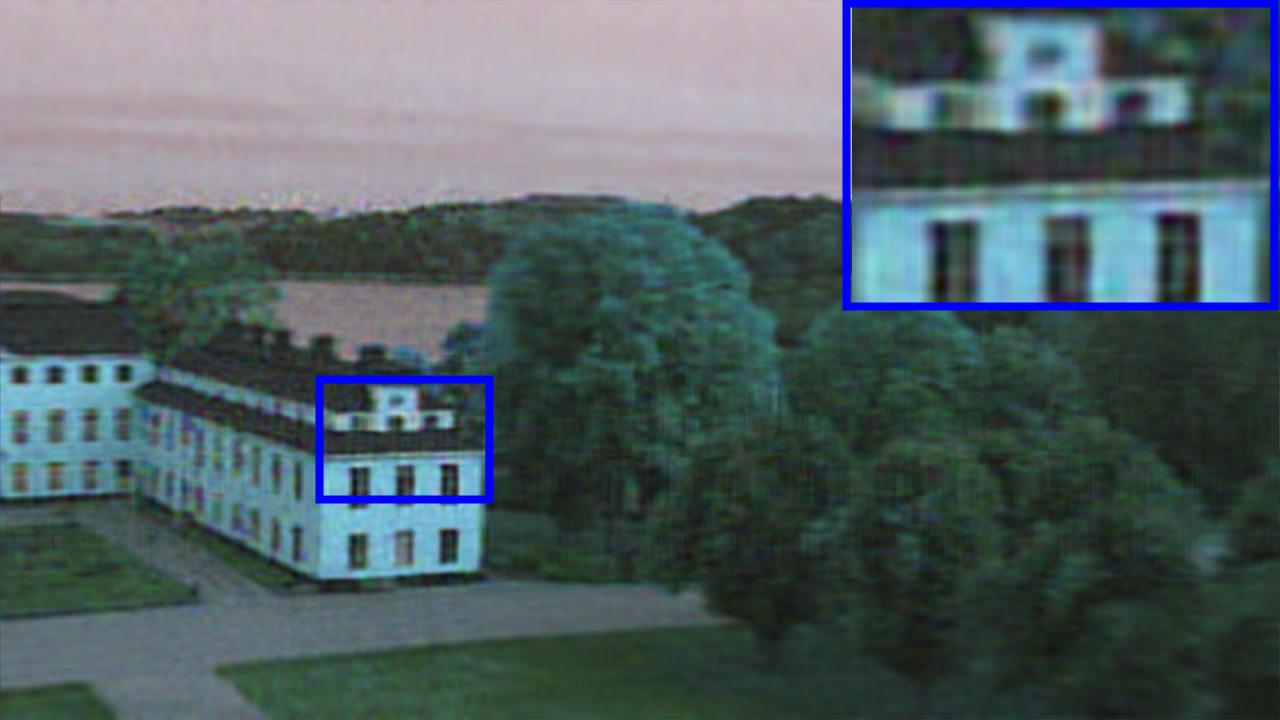} &

\includegraphics[  width=0.875in, height=0.95875in]{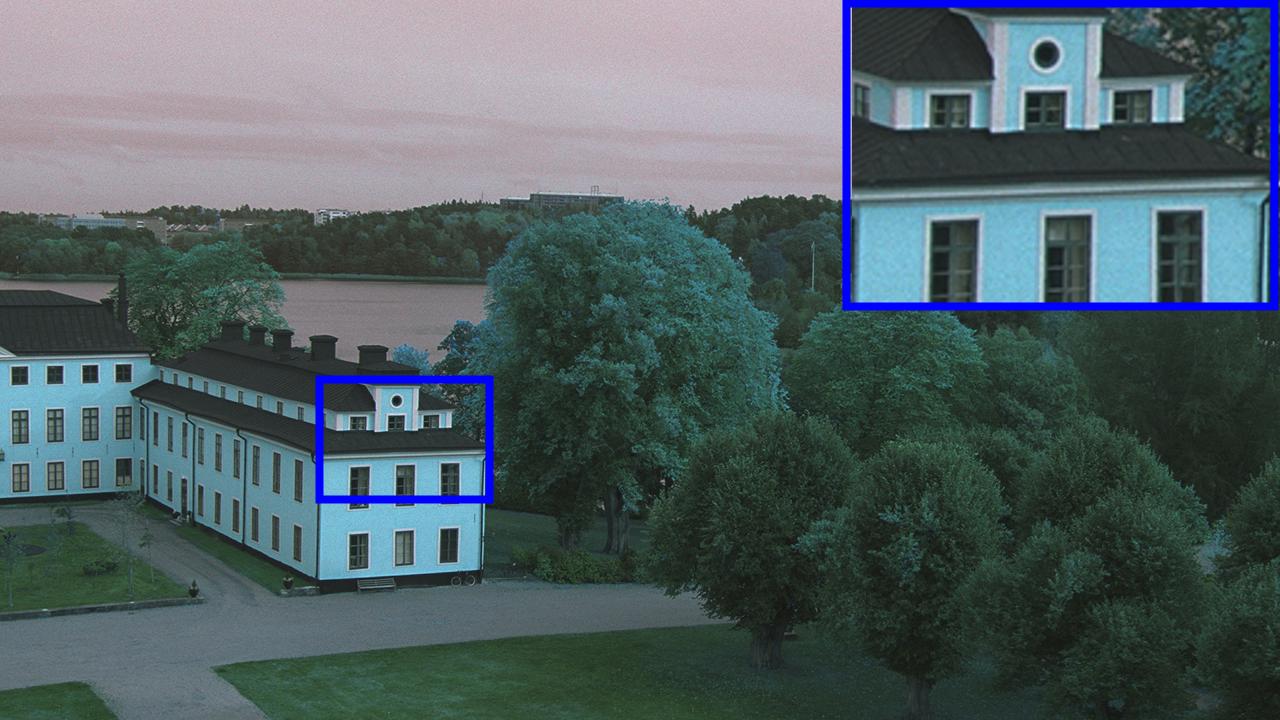}
\\
 \toprule
\end{tabular}
\caption{
Visual comparison of various RTC methods for  CVs restoration under 
  $SR=0.1, SNR= 2dB$.
From left to right: (a) Observed, (b) TRNN,
(c)  TTNN, (d) TSPK, (e) TTLRR, (f) LNOP, (g) NRTRM, (h) BCNRTC,   (i) HWTNN,   (j) HWTSN,  
 (k)  R-HWTSN, (l) TCTV-RTC,   (m) FCTN-GNRTC,   (n) R1-FCTN-GNRTC,  (o) R2-FCTN-GNRTC,  (p) Ground-truth.}
\label{fig_CVCV2} 
\end{figure*}

\begin{figure*}[!htbp]
\renewcommand{\arraystretch}{0.5}
\setlength\tabcolsep{1pt}
\centering
\begin{tabular}{c|c|c |c} 
\centering

\includegraphics[width=1.758in, height=1.25in]{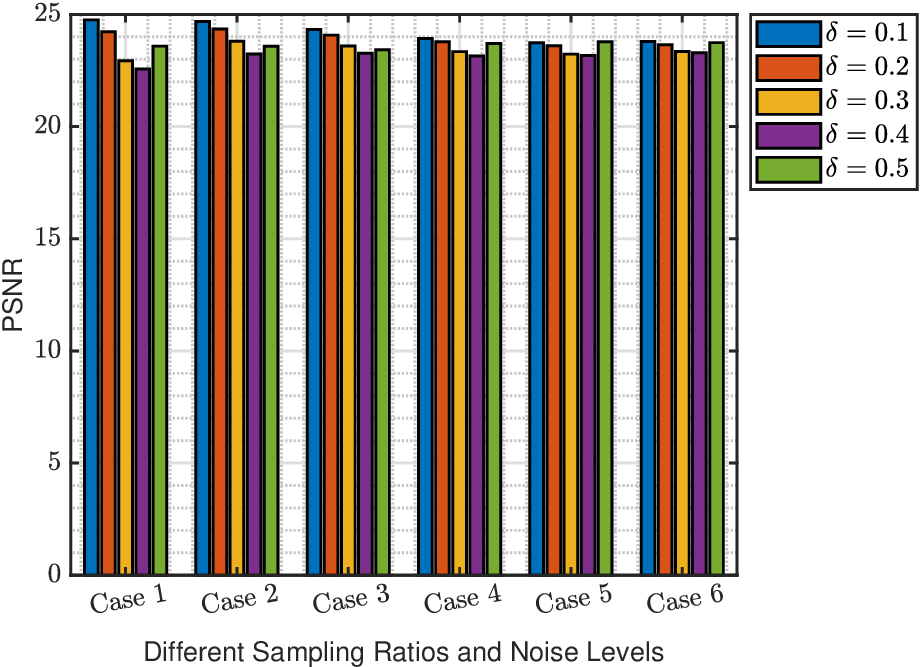}&
\includegraphics[width=1.758in, height=1.25in]{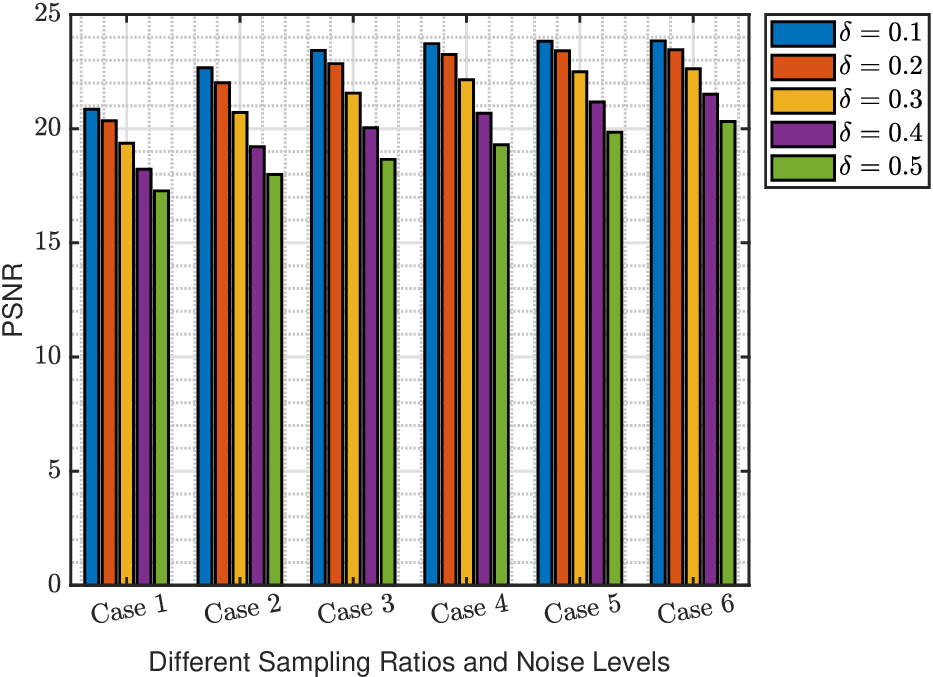}&
\includegraphics[width=1.758in, height=1.25in]{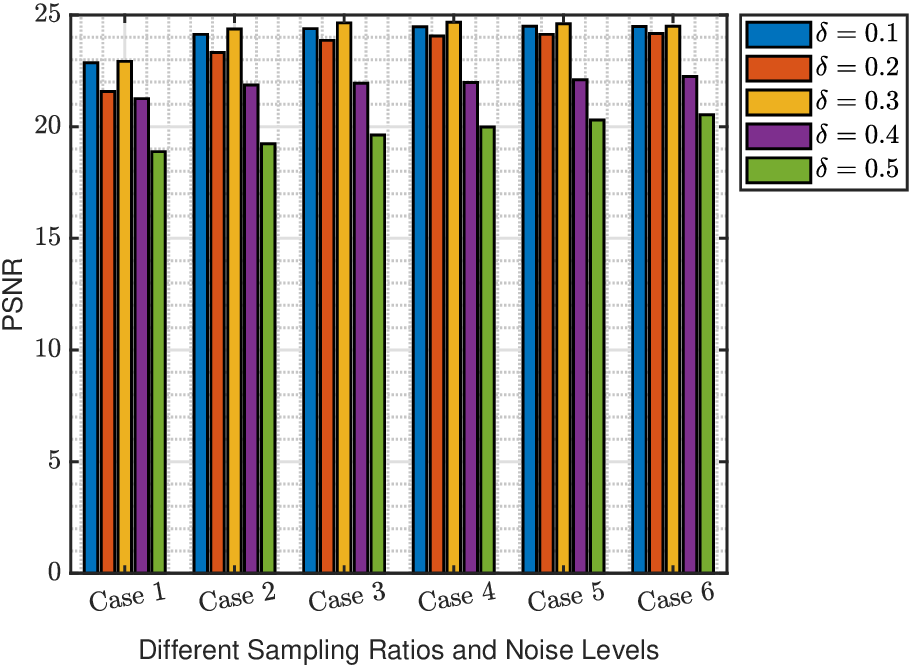}&
\includegraphics[width=1.758in, height=1.25in]{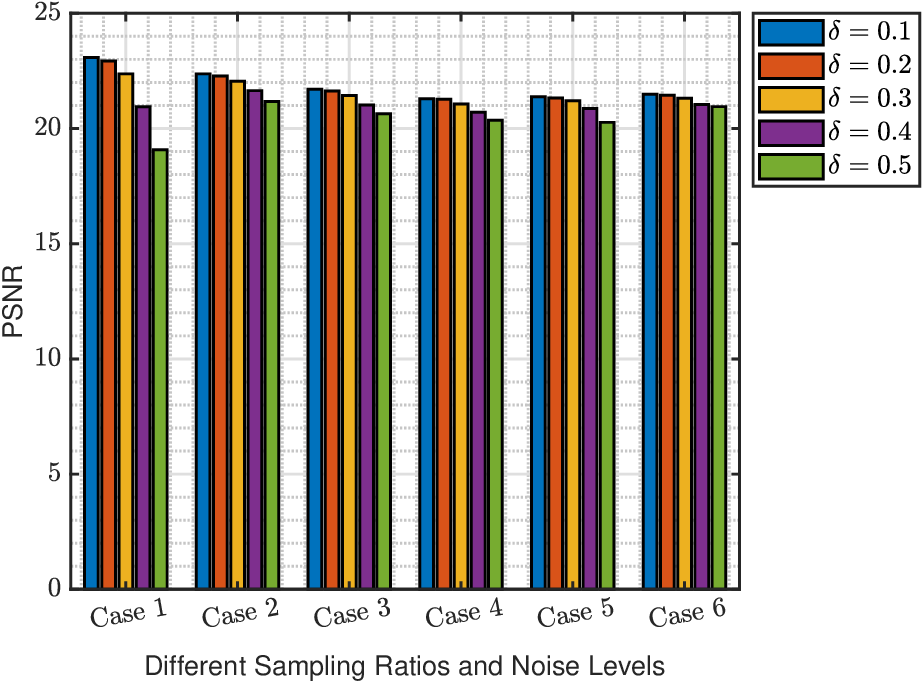}
\\
 %
\includegraphics[width=1.758in, height=1.25in]{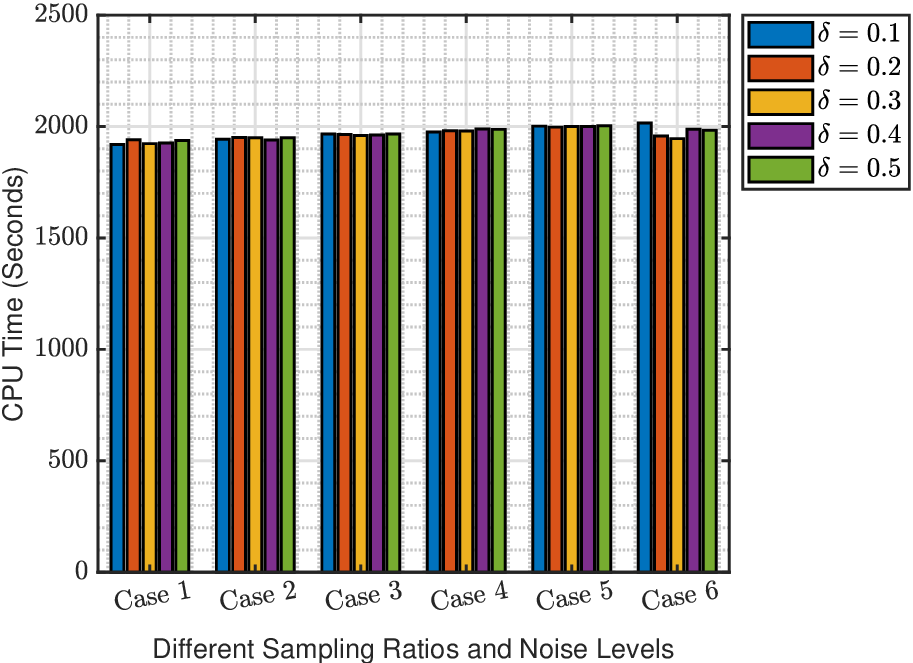}&
\includegraphics[width=1.758in, height=1.25in]{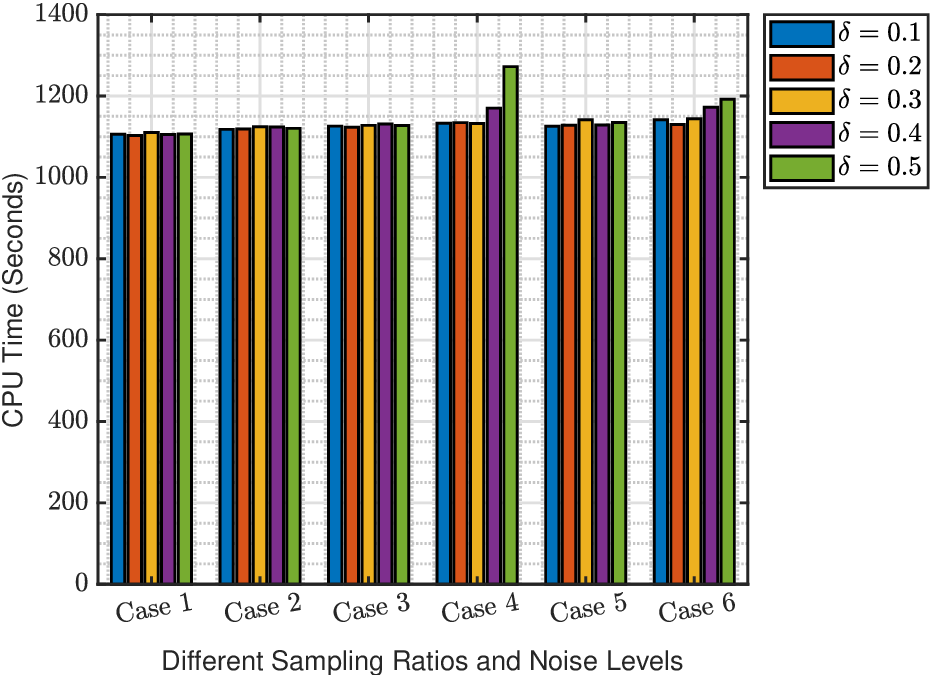}&
\includegraphics[width=1.758in, height=1.25in]{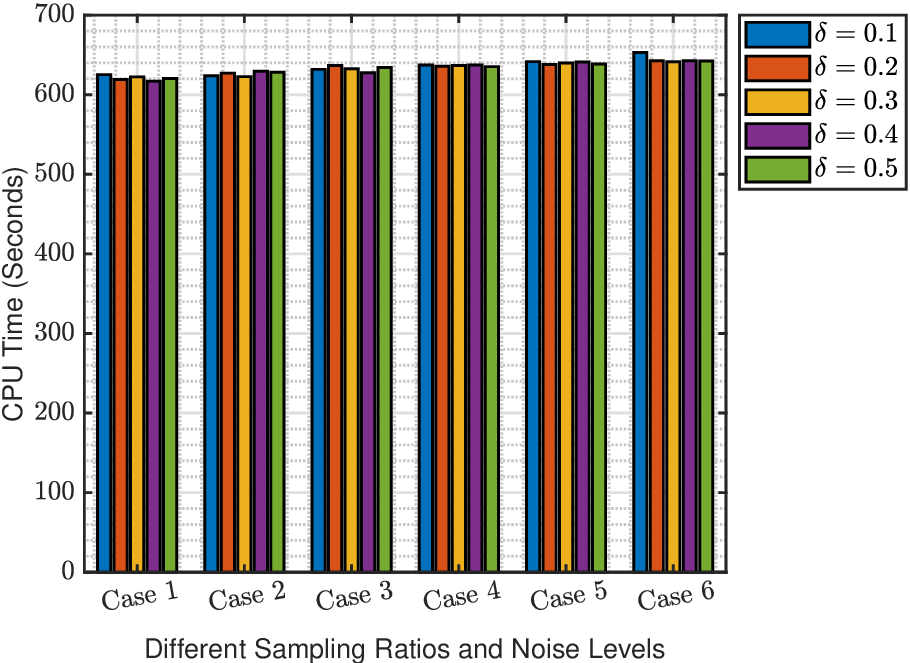}&
\includegraphics[width=1.758in, height=1.25in]{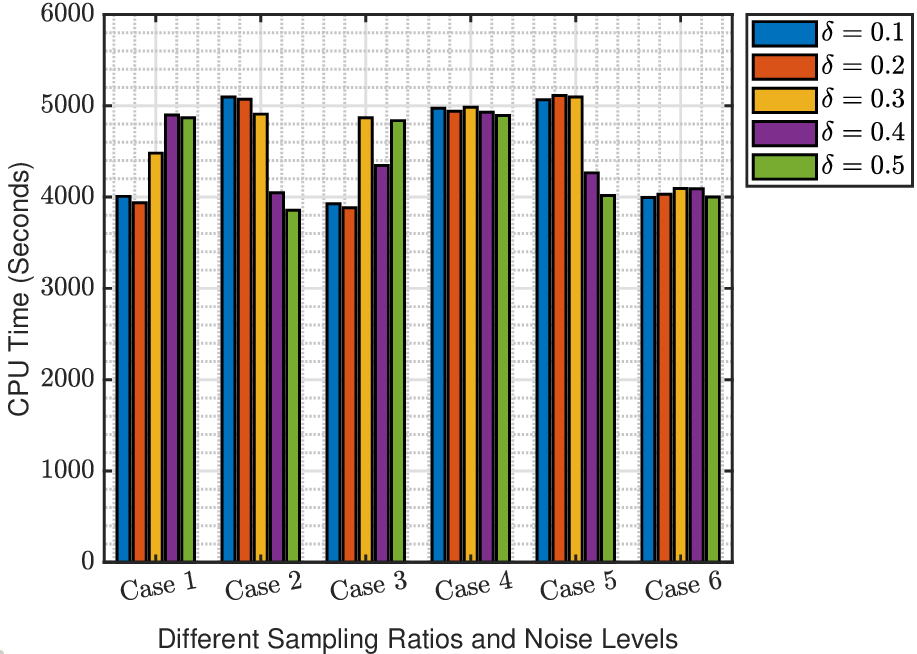}
\\
\small{(a)  CVs} & \small{(b)  Faces} & \small{(c)  MRSIs} &\small{(d)  HVs}
\end{tabular}
\caption{
Recovery performance 
of the proposed  
randomized FCTN-GNQRTC 
on 
various 
datasets
under different quantization resolutions $\delta$, sampling ratios $SR$,  sparse noise levels  $NR$,  and
Gaussian noise levels $\sigma$.
From left to right are
$SR=0.1, NR=0.1, \sigma=0.1$ (Case 1), $SR=0.2, NR=0.1, \sigma=0.1$ (Case 2),
  $SR=0.3, NR=0.1, \sigma=0.1$ (Case 3),   $SR=0.4, NR=0.1, \sigma=0.1$ (Case 4),
  $SR=0.5, NR=0.1, \sigma=0.1$ (Case 5) and  $SR=0.6, NR=0.1, \sigma=0.1$ (Case 6),
   respectively.
}
\label{quan-tensor}
\end{figure*}

\subsection{\textbf{Experiment 1: 
   Tensor Completion}}
\subsubsection{\textbf{Comparison Methods}} %
 We compare the proposed tensor completion method (i.e,  FCTN-GNTC) and its two  accelerated randomized  versions
with  several  state-of-the-art LRTC approaches:
  TRNNM \cite{yu2019tensor},   FCTN-NNM \cite{liu2024fully},  FCTN-TC \cite{zheng2021fully},  FCTNFR \cite{zheng2022tensor},
  HTNN \cite{ qin2022low},  METNN \cite{liu2024revisiting},
   WSTNN \cite{zheng2020tensor44}, EMLCP-LRTC \cite{zhang2023tensor}, OTNN \cite{qiuyn2025},   
   MTTD \cite{feng2023multiplex},     TCTV-TC \cite{wang2023guaranteed},
   %
    GTNN-HOC \cite{wang2024low2222}, and t-$\epsilon$-LogDet \cite{ yang2022355}.
  In our randomized  versions, one is named  R1-FCTN-GNTC, integrating the 
fixed-rank randomized compression strategy,
while the other is called  R2-FCTN-GNTC,  incorporating the 
fixed-accuracy randomized compression strategy.
To ensure fairness, we configure $\Phi(\cdot)=  \psi(\cdot)= \ell_1$ in the proposed methods.

\subsubsection{\textbf{Results and Analysis}}
The  quantitative  metrics (MPSNR, MSSIM, MRSE, MTime)
obtained by the proposed and competitive LRTC methods on different types of tensor datasets are presented in Tables \ref{table-facedata1},
\ref{table-MSIMSI}, \ref{table-MRSIMRSI1}.
From these comprehensive 
evaluations, we arrive at the following conclusions: 1) Especially in low-sampling scenarios, compared with previous 
TR/FCTN-based LRTC methods (e.g., FCTNFR, FCTN-TC,
FCTN-NNM,
TRNN),
the proposed
randomized   algorithms 
 with jointed \textbf{L}+\textbf{S} priors
achieve enhanced restoration 
accuracy
 while incurring relatively low computational costs;
2)
When compared with competitive algorithms (e.g., TCTV-TC,  MTTD,  EMLCP-LRTC)
derived from other tensor frameworks, the proposed method can deliver 
 relatively lower MRSE values,  lower CPU runtime
in most cases, and higher MPSNR and MSSIM values;
3)
Although some methods (e.g., HTNN,  GTNN-HOC,  t-$\epsilon$-LogDet) that rely solely on low-rank priors can achieve shorter computation times than the proposed method,
they perform significantly worse in terms of various accuracy metrics, namely MPSNR, MSSIM, and MRSE.

The corresponding visual comparisons of locally enlarged regions are exemplified in Figures \ref{fig_facehv},
\ref{fig_mri}.
More visual examples are available in the supplementary materials.
 From the perspectives of clarity, color richness, texture, and edge information, the proposed method achieves
the best visual restoration effect.
 Especially under extremely low sampling rates, the proposed
method is still capable of restoring the general appearance of visual data,  
while other comparative algorithms largely fail.
Figure \ref{INTROtimevsPsnr} illustrates the differences in computational efficiency and recovery accuracy between the randomized 
and deterministic versions of the proposed method
across multiple tensor datasets.
We can observe that
while maintaining comparable accuracy (measured by PSNR values), the randomized version is
on average $9$X faster than the deterministic one, and in some individual cases, it achieves up to $20$X speedup.
As the randomized  algorithms induced by 
the  fixed-precision  scheme and the fixed-rank strategy achieve similar results, we  only present the outcomes of the former.

\subsection{\textbf{Experiment 2:  
Robust  Tensor Completion}}

\subsubsection{\textbf{Comparison Methods}} 

 We 
   compare the proposed method (i.e., FCTN-GNRTC) and its two  accelerated randomized  versions
with several  RTC approaches:
TRNN  \cite{huang2020robust},
UTNN  \cite{song2020robust},   TSPK \cite{lou2019robust},
TTLRR \cite{yang2024robust22}, 
LNOP \cite{chen2020robust}, NRTRM \cite{qiu2021nonlocal}, BCNRTC \cite{zhao2022robust},
HWTNN  \cite{qin2021robust}, HWTSN \cite{qin2023nonconvex},   R-HWTSN \cite{qin2023nonconvex},
and TCTV-RTC \cite{wang2023guaranteed}.
In our proposed randomized algorithms, the versions coupled with fixed-rank and fixed-accuracy randomized compression  strategies are
named R1-FCTN-GNRTC and R2-FCTN-GNRTC, respectively.
To uphold the principle of fairness, we set both $\Phi(\cdot)$ and $ \psi(\cdot) $  to be equivalent to the basic
$\ell_1$-norm  in the proposed methods.


\subsubsection{\textbf{Results and Analysis}}

Table \ref{table-mrsi-rhtc}  
provides a comprehensive comparison of the proposed algorithm against several competitive benchmarks,
presenting quantitative results (MPSNR, MSSIM, MRSE, CPU Time) on several MRSI datasets.
Figure \ref{rhtc_cv} showcases the corresponding performance outcomes obtained from CV datasets.
Accordingly, 
a comprehensive  comparison 
of the visual quality 
on two categories of robust completion  tasks is illustrated in Figures \ref{fig_MRSI2}, \ref{fig_CVCV2}.
From these results, we find that
in comparison with existing approaches 
that are only modeled by
the global low-rankness prior,
the proposed method achieves a $2 \thicksim 3$ dB gain in MPSNR.
The proposed
R1-FCTN-GNRTC and R2-FCTN-GNRTC
algorithms reach recovery accuracy similar to the competitive TCTV-RTC algorithm,
and at the same time cut down computational time significantly.
In the proposed method, the randomized  versions are approximately $2.5$  times faster than the deterministic version for  CVs recovery task, and about $5 \thicksim 7$  times faster for MRSIs  restoration task.

%
%

\subsection{\textbf{Experiment 3: 
 Quantized  Tensor Recovery}}

In this subsection, we firstly focus on investigating the impacts of various 
quantization resolutions 
upon the performance of the proposed quantized 
algorithms
under different sampling rates and noise levels.
The relevant experimental results are presented in Figures \ref{quan-tensor}, where
$\delta=0.1, 0.2, 0.3, 0.4, 0.5$,
$SR=0.1,0.2,0.3,0.4,0.5,0.6$, $NR=0.1$, $\sigma=0.1$.
The results for more sampling and  noise corruption scenarios 
are available in the supplementary materials.
From these results,
it can be observed that increasing the quantization step size $\delta$ leads to a gradual decline in recovery performance
for the proposed quantized method.

\section{\textbf{Conclusions and  Future Work}}\label{conclusion}
In this article, by investigating  
efficient  gradient-domain regularization and   randomized acceleration strategies,
we propose a novel scale-aware tensor modeling and computation 
framework 
for large-scale high-dimensional visual data recovery.
Specifically, we put forward reliable and effective tensor models by 
an innovative
gradient-domain regularization  scheme based on FCTN decomposition,
where the observation transitions from non-quantized scenarios to quantized scenarios.
To alleviate the computational bottlenecks encountered in processing large-scale data, fast randomized  compression and representation strategies are developed
as the core acceleration modules in algorithm design.
From a theoretical perspective, 
the theoretical error bounds and convergence properties of the proposed algorithms are analyzed.
Extensive experimental results on various multi-dimensional 
tensor datasets
have demonstrated the effectiveness and superiority of the proposed method compared to current competing approaches.
In future work, 
by integrating tensor methods/theories, quantization, randomized  sketching
with deep-learning techniques,
 we plan to research faster and more efficient approaches for large-scale high-dimensional data approximation and recovery.


\ifCLASSOPTIONcaptionsoff
  \newpage
\fi

\ifCLASSOPTIONcaptionsoff
  \newpage
\fi

\bibliographystyle{IEEEtran}
\bibliography{reference_newnew//rhtc}

\end{document}